\documentclass[journal]{IEEEtran}
\usepackage[T1]{fontenc}
\usepackage[latin9]{inputenc}
\usepackage{color}
\usepackage{array,authblk,amssymb}
\usepackage{multirow}
\usepackage{fixltx2e}
\usepackage{graphicx}
\usepackage{cite}
\usepackage{amsfonts,xcolor,colortbl} 
\usepackage{amsmath}

\usepackage{amsthm,epstopdf }
\theoremstyle{definition}
\newtheorem{definition}{Definition}

\makeatletter
\def\endthebibliography{%
	\def\@noitemerr{\@latex@warning{Empty `thebibliography' environment}}%
	\endlist
}
\makeatother

\providecommand{\tabularnewline}{\\}

\makeatother

\begin{document}

\title{Transfer Learning for Wireless Networks:\\ A Comprehensive Survey}

\author{\IEEEauthorblockN{Cong T. Nguyen\IEEEauthorrefmark{1}, Nguyen Van Huynh\IEEEauthorrefmark{1}, Nam~H.~Chu\IEEEauthorrefmark{1}, Yuris Mulya Saputra\IEEEauthorrefmark{1}, Dinh Thai Hoang\IEEEauthorrefmark{1}, Diep~N.~Nguyen\IEEEauthorrefmark{1}, Quoc-Viet Pham\IEEEauthorrefmark{2}, Dusit Niyato\IEEEauthorrefmark{3}, Eryk Dutkiewicz\IEEEauthorrefmark{1} and Won-Joo Hwang\IEEEauthorrefmark{4}}
	
\IEEEauthorblockA{\IEEEauthorrefmark{1}School of Electrical and Data Engineering, University of Technology Sydney, Sydney, NSW 2007, Australia}
\IEEEauthorblockA{\IEEEauthorrefmark{2}Research Institute of Computer, Information and Communication, Pusan National University, Busan 46241, Korea}
\IEEEauthorblockA{\IEEEauthorrefmark{3}School of Computer Science and Engineering, Nanyang Technological University, Singapore 639798, Singapore}
\IEEEauthorblockA{\IEEEauthorrefmark{4}Department of Biomedical Convergence Engineering, Pusan National University, Yangsan 50612, Korea }
}

\maketitle
\begin{abstract}
With outstanding features, Machine Learning (ML) has become the backbone of numerous applications in wireless networks. However, the conventional ML approaches face many challenges in practical implementation, such as the lack of labeled data, the constantly changing wireless environments, the long training process, and the limited capacity of wireless devices. These challenges, if not addressed, can impede the effectiveness and applicability of ML in wireless networks. To address these problems, Transfer Learning (TL) has recently emerged to be a promising solution. The core idea of TL is to leverage and synthesize distilled knowledge from similar tasks as well as from valuable experiences accumulated from the past to facilitate the learning of new problems. Doing so, TL techniques can reduce the dependence on labeled data, improve the learning speed, and enhance the ML methods' robustness to different wireless environments. This article aims to provide a comprehensive survey on applications of TL in wireless networks. Particularly, we first provide an overview of TL including formal definitions, classification, and various types of TL techniques. We then discuss diverse TL approaches proposed to address emerging issues in wireless networks. The issues include spectrum management, localization, signal recognition, security, human activity recognition and caching, which are all important to next-generation networks such as 5G and beyond. Finally, we highlight important challenges, open issues, and future research directions of TL in future wireless networks.
\end{abstract}


\section{Introduction}
The last decade has witnessed the rapid growth of Machine Learning (ML) applications in wireless networks thanks to its agility and efficacy, especially in dealing with uncertainty and dynamics in large-scale problems~\cite{sun_application_2019,bkassiny_survey_2012}. A few recent reports by Samsung~\cite{Samsung_2020} and Ericsson~\cite{Ericsson_2019} showed that ML has been developing as the backbone of 5G networks, and it will be the main driver of mobile technology and the 6G in the near future. However, some recent studies revealed that conventional ML solutions have shortcomings, especially when they are applied to solve emerging problems in wireless networks, due to special characteristics of wireless communications, such as high mobility, dynamic environments, diverse connections, and interference~\cite{kato_ten_2010,chen_artificial_2019}. 
In particular, conventional ML techniques are usually trained for a specific scenario. However, the wireless environments are subject to significant variations which make the new scenarios different from the previous ones (e.g., mobility of mobile users as well as changes in mobile users' data demands). Consequently, this can significantly impact the ML techniques' performances and hinder their applicability. Moreover, the performances of ML techniques mainly rely on the availability of training data, but acquiring a sufficient amount of data might be costly and time-consuming. Even if the training data are sufficient, ML techniques, such as deep learning, usually require a long training time, which makes them impractical for many latency-sensitive applications. Apart from the training time issues, many wireless devices, e.g., IoT devices, are constrained by their limited computing capacity, and thus they are unable to train high-complexity ML models. Moreover, centralized ML systems usually require a huge amount of data to be collected at central servers for processing and training. In addition to communication overhead, sending raw data to the servers may also threaten network users' privacy because sensitive information, e.g., healthcare, is sent over wireless networks or public networks.

\begin{figure}[!t]
	\centering
	\includegraphics[scale=0.27]{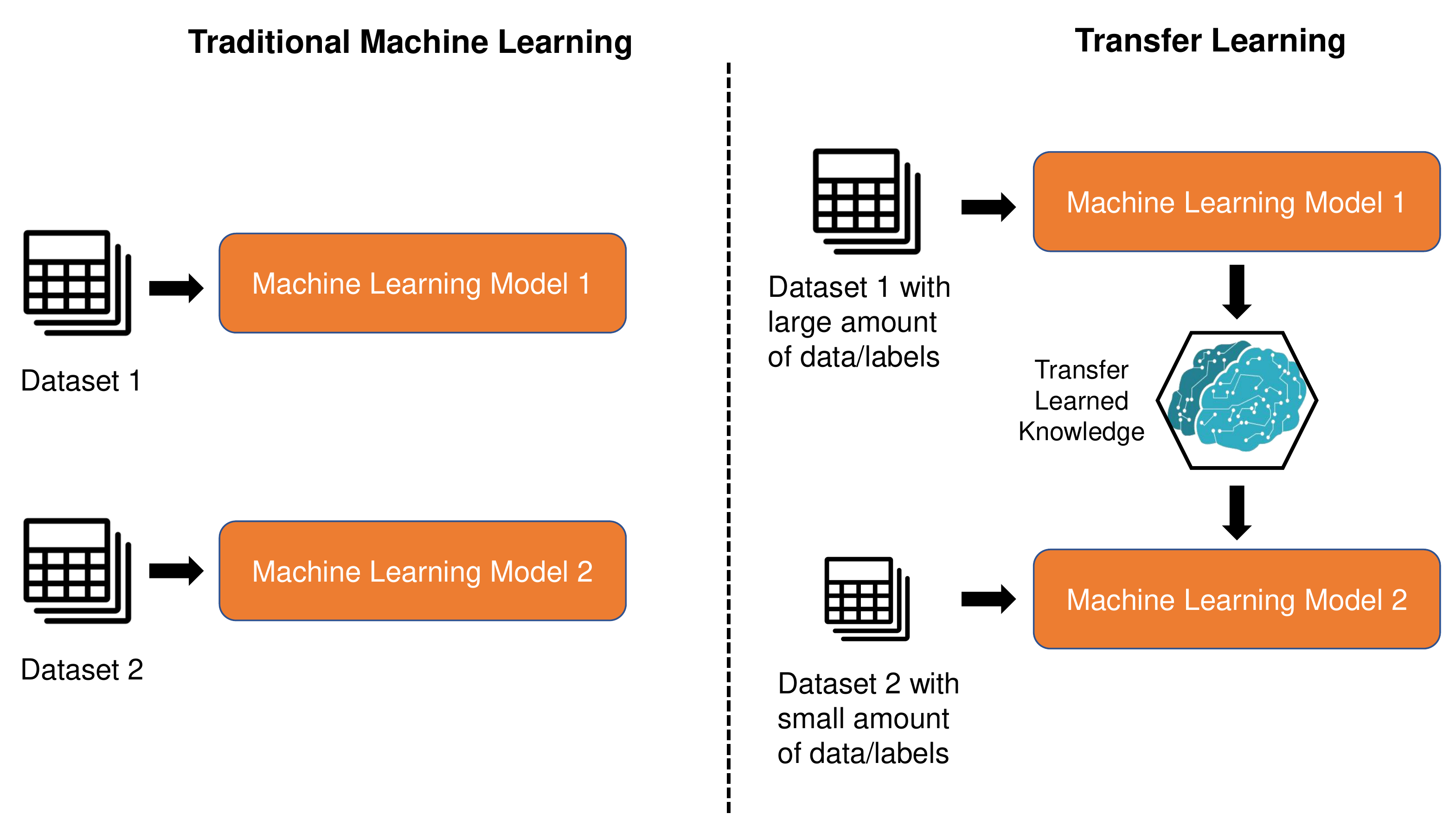}
	\caption{Traditional ML vs. TL.}
	\label{Fig.compare_ml_tl}
\end{figure}

\begin{figure*}[!t]
	\centering
	\includegraphics[width=2\columnwidth]{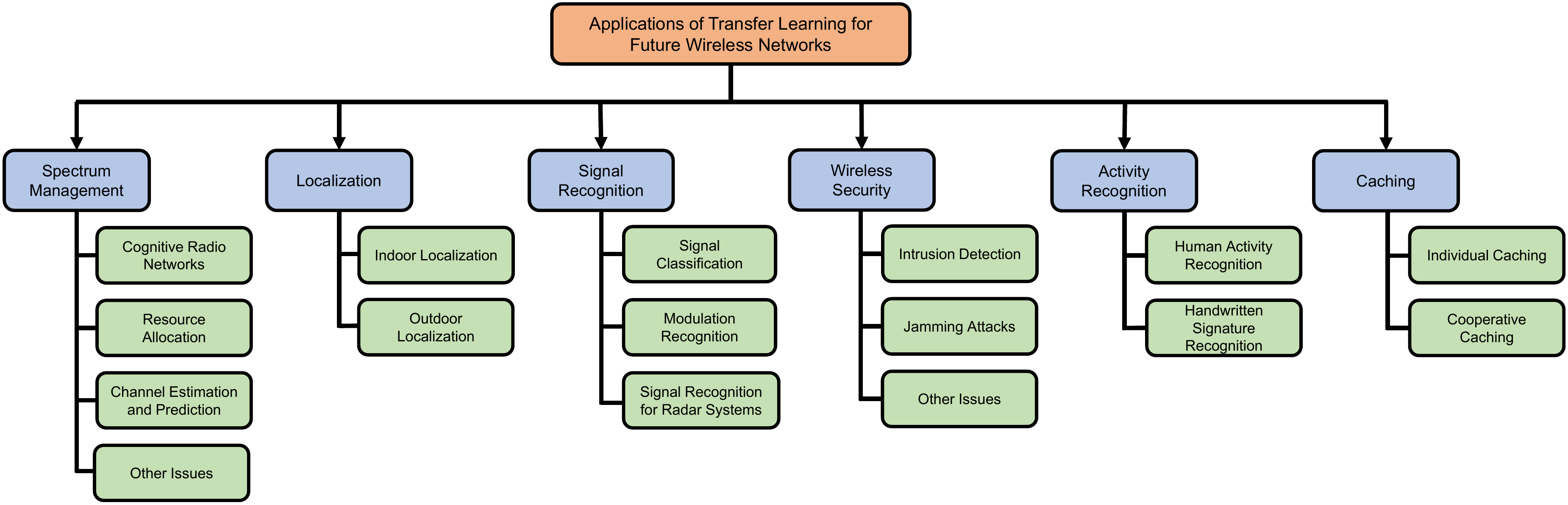}
	\caption{Taxonomy of the applications of TL for wireless networks.}
	\vspace{-0.5em}
	\label{fig:intro}
\end{figure*}

To address these challenges, Transfer Learning (TL) has recently emerged as a highly effective solution. Unlike conventional ML techniques that are trained to solve a specific problem, TL leverages valuable knowledge from similar tasks and previous experiences to significantly enhance the learning performance of conventional ML techniques, as illustrated in Fig.~\ref{Fig.compare_ml_tl}. As a result, TL possesses numerous advantages over traditional ML approaches~\cite{zhuang_comprehensive_2020,cook_transfer_2013} which can be summarized as follows:
\begin{itemize}
	\item \textit{Enhance quality and quantity of training data:} One of the most challenging tasks for conventional ML approaches is finding sufficient and high-quality data for the training process. TL can easily overcome this problem by selecting and transferring knowledge from similar domains with a large amount of high-quality data. As a result, TL is considered to be a promising solution for ML-based wireless networks in the near future. 
	\item \textit{Speed-up learning processes:} Instead of learning from scratch like conventional ML approaches, the training process in TL can be significantly sped up thanks to valuable knowledge shared from other similar domains and/or learned in the past. As a result, this can remarkably improve the learning rate, which is especially crucial for the development of ultra-low latency applications for future wireless networks. 
	\item \textit{Reduce computing demands:} Conventional ML approaches usually require a huge amount of computing resources for the training processes. However, with TL, most of the data were trained by other source domains before the trained models are transferred to the target domain, thereby significantly reducing computing demands for the training process at the target domain. This is particularly useful for wireless devices (e.g., smartphones and edge devices) as they usually have hardware constraints.
	\item \textit{Mitigate communication overhead:} For TL approaches, instead of sending the raw data with large sizes, only knowledge, e.g., weights of trained models, needs to be transferred. As a result, the communication overhead can be substantially reduced for wireless networks. 
	\item \textit{Protect data privacy:} In TL, instead of learning from raw data from other domains, ones only need to learn from their trained models (expressed through weights), and thus data privacy can be protected. This feature of TL is very helpful for privacy-sensitive wireless applications such as healthcare and military communication networks. 
\end{itemize}

With the aforementioned outstanding advantages, TL has been leveraged to address various challenges that traditional ML techniques are facing in numerous wireless network applications. This article aims to provide an in-depth and comprehensive survey on TL applications in wireless networks. Particularly, we first provide an in-depth tutorial of TL, including formal definitions and some fundamental knowledge of various types of TL techniques. Then, a comprehensive survey on the applications of TL in many areas of wireless networks is presented. For these TL applications, we provide detailed reviews and analyses on how TL is leveraged to overcome the challenges that the underlying ML techniques are facing. Finally, we discuss the current challenges and open issues of implementing TL in wireless networks and introduce some promising research directions for applications of TL in the future.

There are a few surveys on the applications of ML in wireless networks, such as~\cite{bkassiny_survey_2012,kumar_machine_2019,jagannath_machine_2019,klaine_survey_2017,alsheikh_machine_2014}, with different focuses. However, these papers emphasize conventional ML approaches, e.g., supervised learning, unsupervised learning, and reinforcement learning. Alternatively, recent surveys focus on applications of advanced ML techniques in wireless networks such as~\cite{zhang_deep_2019a,lim_federated_2020}, and~\cite{luong_applications_2019}. Particularly,~\cite{zhang_deep_2019a,lim_federated_2020}, and~\cite{luong_applications_2019} focus on applications of Deep Learning (DL), Federated Learning, and Deep Reinforcement Learning (DRL) in wireless networks, respectively. In addition, there are a couple of survey papers focusing on TL techniques, e.g.,~\cite{zhuang_comprehensive_2020}, and its applications in activity recognition only, e.g.,~\cite{cook_transfer_2013}. Nevertheless, from all aforementioned papers, and to the best of our knowledge, there is still no comprehensive survey on the applications of TL in wireless networks. Given the rapid development of wireless communication technologies along with increasing requirements for high Quality-of-Service (QoS), interoperability, robustness, as well as privacy and security from mobile users, TL is in urgent need to address the current limitations of ML-based wireless networks, especially in beyond 5G and 6G networks. As a result, this paper is expected to fill the gap and significantly contribute to the future development of intelligent wireless networks.

The rest of this paper is organized as follows. Section~\ref{sec:tutorial} provides a tutorial on TL including formal definitions, classification, and various types of TL techniques. Then, TL-based applications in various areas of wireless networking (as summarized in Fig.~\ref{fig:intro}) such as spectrum management, localization, signal recognition, security, human activity recognition, and caching, are presented from Section~\ref{sec:spectrum} to Section~\ref{sec:caching}. Section~\ref{sec:challenge} discusses challenges, open issues, and future research directions of TL. Finally, Section~\ref{sec:Summary} concludes the paper.


\section{Transfer Learning: An Overview}\label{sec:tutorial}
In this section, we first present the fundamental knowledge of TL and its advantages in learning processes. Then, the formal definition of TL is presented. Afterward, important TL strategies are discussed in detail. Finally, we present the current advances in TL.
\subsection{Overview of Transfer Learning}
Recently, ML has been successfully applied to various areas and received great attention from both industry and academia. The main goal of ML is to extract and learn data structures to estimate future outcomes. However, existing ML approaches usually assume that the unlabeled data and labeled data share the same data distribution and feature space~\cite{pan_a_2010}. In many scenarios, collecting the training data with the same data distribution and feature space as in the testing data is costly and intractable~\cite{weiss_a_2016}. In this case, the prediction and classification performance is poor as what has been learned from the training data is less useful for the testing data. Another problem of conventional ML techniques is the lack of training data as collecting data in some cases is complex or even impossible~\cite{tan_a_2018}.

To address these problems, TL is introduced to be an effective solution. Different from traditional ML, the key fundamental of TL is transferring knowledge gained from one source task to improve the learning of a related target task as illustrated in Fig.~\ref{Fig.compare_ml_tl}. TL is motivated by the fact that based on knowledge learned before, we can solve new problems more efficient and faster. For example, people who have learned to play violin may learn to play guitar faster than others with no previous experience in playing musical instruments~\cite{weiss_a_2016, zhuang_comprehensive_2020}.

TL has a long history. The concept of TL may initially be introduced in educational psychology~\cite{zhuang_comprehensive_2020}. In the field of ML, TL was reviewed in NIPS-95's workshop on ``Learning to Learn''. Since then, under different terms like ``learning to learn'', ``life-long learning'', and ``knowledge transfer'', TL has been studied and applied in various aspects of our daily life. It is expected that ``TL will be the next driver of ML commercial success after supervised learning''~\cite{ng_nuts_2016}.
\subsection{Fundamentals of Transfer Learning}
We present the formal definition of TL. As mentioned, in TL, learned knowledge will be transferred from the source domain to the target domain to improve the learning process of the target task. Thus, in the following, we first present the definition of a ``domain'' and a ``task''.

\begin{definition}
	\label{def.domain}
	\textit{``Domain: A domain $\mathcal{D}$ is defined by two parts: (i) a feature space $\mathcal{X}$ and (ii) a marginal probability distribution $\mathcal{P}(X)$ in which $X = \{x_1, \ldots, x_n\} \in \mathcal{X}$ where $n$ is the number of feature vectors in $X$. As such, $\mathcal{D}= \{\mathcal{X}, \mathcal{P}(X)\}$.''}
\end{definition}

\begin{definition}
	\textit{``Task: Given domain $\mathcal{D}$, a task $\mathcal{T}$ is defined by two parts: (i) a label space $\mathcal{L}$ and (ii) a predictive function $f(\cdot)$. The predictive function (or decision function) is learned from the feature vector and label space pairs $\{x_i, l_i\}$, with $x_i \in X$ and $l_i\in \mathcal{L}$. In other words, a task is defined by $\mathcal{T}=\{\mathcal{L}, f(\cdot)\}$.''}
\end{definition}

In general, the predictive function represents the prediction of the corresponding label $f(x_i)$ given instance $x_i$~\cite{pan_a_2010},~\cite{zhuang_comprehensive_2020}. In this case, the predictive function can be defined as $f(x_i)= \{P(l_k|x_i)|l_k \in \mathcal{L}, k = 1, \ldots, |\mathcal{L}|\}.$ For example, in the signal classification application, $\mathcal{L}$ is the set of all labels which are ``0'' or ``1''. The predictive function can predict the probability of bit ``0'' or bit ``1'' based on instance $x$ in the feature space. Note that $x$ can contain all the features in the feature space or just critical features (which depend on the considered problem) in the feature space.

It is worth noting that a domain contains various labeled or unlabeled data which are denoted as the domain data. In the context of TL, we denote $D_\mathrm{S}$ as the source domain data, i.e., $D_\mathrm{S} = \{(x_{\mathrm{S}_1}, l_{\mathrm{S}_1}), \ldots, (x_{\mathrm{S}_i}, l_{\mathrm{S}_i}), \ldots, (x_{\mathrm{S}_{n_\mathrm{S}}}, l_{\mathrm{S}_{n_\mathrm{S}}})\}$ with $x_{\mathrm{S}_i} \in \mathcal{X}_\mathrm{S}$ represents the $i$-th data instance of $D_\mathrm{S}$ and $l_{\mathrm{S}_i} \in \mathcal{L}_\mathrm{S}$ denotes the corresponding label of $x_{\mathrm{S}_i}$. Similarly, the target domain data can be defined as $D_\mathrm{T} = \{(x_{\mathrm{T}_1}, l_{\mathrm{T}_1}), \ldots, (x_{\mathrm{T}_i}, l_{\mathrm{T}_i}), \ldots, (x_{\mathrm{T}_{n_\mathrm{T}}}, l_{\mathrm{T}_{n_\mathrm{T}}})\}$, in which $x_{\mathrm{T}_i} \in \mathcal{X}_\mathrm{T}$ is the $i$-th data instance of $D_\mathrm{T}$ and $l_{\mathrm{T}_i} \in \mathcal{L}_\mathrm{T}$ is the corresponding label of $x_{\mathrm{T}_i}$. Practically, the number of instances at the target domain is often much lower than that of the source domain, i.e., $0 \leq n_\mathrm{T} \ll n_\mathrm{S}$. With TL, the learning process of the target domain can be significantly improved by leveraging the knowledge transferred from the source domain. We now formally define TL in Definition~\ref{def.transfer_learning}.

\begin{definition}
	\label{def.transfer_learning}
	\textit{``Transfer Learning: Given a source domain $\mathcal{D_\mathrm{S}}$ with a corresponding source task $\mathcal{T}_\mathrm{S}$ and a target domain $\mathcal{D}_\mathrm{T}$ with a corresponding target task $\mathcal{T}_\mathrm{T}$, the goal of TL is to learn the target predictive function $f_\mathrm{T}(.)$ by leveraging the knowledge gained from $\mathcal{D}_\mathrm{S}$ and $\mathcal{T}_\mathrm{S}$, where $\mathcal{D}_\mathrm{S} \neq \mathcal{D}_\mathrm{T}$ or $\mathcal{T}_\mathrm{S} \neq \mathcal{T}_\mathrm{T}$\footnote{It is worth noting that when the source domain and target domain are the same (i.e., $\mathcal{D}_\mathrm{S} = \mathcal{D}_\mathrm{T}$) and the source task and the target task are the same (i.e., $\mathcal{T}_\mathrm{S} = \mathcal{T}_\mathrm{T}$), TL becomes traditional ML~\cite{pan_a_2010}.}.''}
\end{definition}

From Definition~\ref{def.domain}, the target domain is a pair $\mathcal{D}_\mathrm{T}= \{\mathcal{X}_\mathrm{T}, \mathcal{P}(X_\mathrm{T})\}$ and the source domain is a pair $\mathcal{D}_\mathrm{S}= \{\mathcal{X}_\mathrm{S}, \mathcal{P}(X_\mathrm{S})\}$. Thus, $\mathcal{D}_\mathrm{S} \neq \mathcal{D}_\mathrm{T}$ means that $\mathcal{X}_\mathrm{S} \neq \mathcal{X}_\mathrm{T}$ and/or $\mathcal{P}(X_\mathrm{S}) \neq \mathcal{P}(X_\mathrm{T})$. When $\mathcal{X}_\mathrm{S} \neq \mathcal{X}_\mathrm{T}$, the learning problem is defined as heterogeneous TL. Differently, when  $\mathcal{X}_\mathrm{S} = \mathcal{X}_\mathrm{T}$, the learning problem is defined as homogeneous TL~\cite{weiss_a_2016}. $\mathcal{P}(X_\mathrm{S}) \neq \mathcal{P}(X_\mathrm{T})$ when the marginal distributions of the target domain and the source domain are different. It is proven that knowledge gained from a given source domain will not achieve the optimal prediction on a target domain if their marginal distributions $\mathcal{P}$ are not the same~\cite{pan_a_2010, weiss_a_2016}.

Another scenario of TL is $\mathcal{T}_\mathrm{S} \neq \mathcal{T}_\mathrm{T}$. As mentioned,  $\mathcal{T}_\mathrm{S}=\{\mathcal{L}_\mathrm{S}, f(.)\}$ = $\{\mathcal{L}_\mathrm{S}, \mathcal{P}(L_\mathrm{S}|X_\mathrm{S})\}$ and $\mathcal{T}_\mathrm{T}=\{\mathcal{L}_\mathrm{T}, f(.)\}$ = $\{\mathcal{L}_\mathrm{T}, \mathcal{P}(L_\mathrm{T}|X_\mathrm{T})\}$. Thus, $\mathcal{T}_\mathrm{S} \neq \mathcal{T}_\mathrm{T}$ represents the cases when $\mathcal{L}_\mathrm{S} \neq \mathcal{L}_\mathrm{T}$ and/or $\mathcal{P}(L_\mathrm{S}|X_\mathrm{S}) \neq \mathcal{P}(L_\mathrm{T}|X_\mathrm{T})$. When $\mathcal{P}(L_\mathrm{S}|X_\mathrm{S}) \neq \mathcal{P}(L_\mathrm{T}|X_\mathrm{T})$, the conditional probability distributions of the source and the target domains are not the same. The case in which $\mathcal{L}_\mathrm{S} \neq \mathcal{L}_\mathrm{T}$ represents that the label spaces of the source domain and the target domain are different. Another case that reduces the prediction performance of TL is when $\mathcal{P}(L_\mathrm{S}) \neq \mathcal{P}(L_\mathrm{T})$, which happens when the labeled samples of the source and target domains are unbalanced~\cite{weiss_a_2016}.

\subsection{Transfer Learning Techniques}

The key fundamental of TL is utilizing knowledge learned from the source domain to improve the learning of the target domain. To guarantee a good transfer performance, the following three main issues need to be taken into account.

\begin{itemize}
	\item \textit{What to transfer:} Deciding what to be transferred is the most critical step in TL. To address this issue, one needs to decide which part of learned knowledge will be transferred to improve the learning process of the target domain. This stems from the fact that not all learned knowledge from the source domain can be useful for the target domain in many scenarios. For example, some knowledge can be common in both the source and the target domains, whereas some knowledge is specific to the source domain but not the target domain.
	
	\item \textit{When to transfer:} Transferring knowledge is not always helpful in speeding up the learning process of the target domain. Thus, one needs to know when the knowledge should not be transferred. For instance, if the target domain does not have anything in common with the source domain, i.e., they are not related, TL may not improve the learning process and even make the learning process less effective (i.e., \textit{negative transfer}).
	
	\item \textit{How to transfer:} Once the what and when questions have been addressed, one can proceed to transfer learned knowledge to the target domain. This process requires different techniques and designs to maximize the transferring utilization at the target domain.
\end{itemize}

Through these three questions, several categorization criteria of TL are proposed. For example, TL can be classified into (i) feature-based, (ii) parameter-based, (iii) relational-based, and (iv) instance-based TL~\cite{zhuang_comprehensive_2020}. In particular, feature-based techniques can be used to construct a new feature representation based on the original features. They can be further categorized into (a) symmetric feature-based and (b) asymmetric-based TL. The symmetric method aims to transform the target and source features to a new feature representation, whereas the asymmetric one aims to transform the source features to the form of the target features. Differently, parameter-based techniques aim to transfer hyper-parameters and/or information of the source model. Relational-based techniques transfer knowledge by learning the common relationships between the target domain and the source domain~\cite{zhuang_comprehensive_2020}. Finally, instance-based techniques aim to reweight the data samples in the source domain to reduce the differences between the source and the target marginal distributions.

In this survey, we categorize TL based on the problem perspective. In particular, there are three types of TL: (i) inductive TL, (ii) transductive TL, and (iii) unsupervised TL. In the inductive TL, the target domain is the same as the source domain, but their tasks are different. In this case, the inductive biases of the source domain will be transferred to the target domain and utilized to improve the learning performance of the target task. In the inductive TL, there are two subcategories: (a) self-taught learning and (b) multi-task learning. In self-taught learning, the labeled data are not available at the source domain. Alternatively, in multi-task learning, the labeled data are available at the source domain, and both the target and source domains learn their tasks simultaneously. In transductive TL, the source task and the target task are similar, but their domains are different. Finally, in unsupervised learning, the labeled data are not available at both the source domain and the target domain. These settings of TL are summarized in Table~\ref{table_categorization}.

\begin{table*}
	\caption{TL Categorization}
	\label{table_categorization}
	\begin{centering}
		\begin{tabular}{|>{\raggedright\arraybackslash}m{2cm}|>{\raggedright\arraybackslash}m{4cm}|>{\raggedright\arraybackslash}m{2.5cm}|>{\raggedright\arraybackslash}m{2cm}|>{\raggedright\arraybackslash}m{2cm}|>{\raggedright\arraybackslash}m{2.5cm}|}
			\hline 
			\multicolumn{1}{|>{\centering\arraybackslash}m{2cm}|}{\cellcolor{gray!35}\textbf{TL Strategy}} & \multicolumn{1}{>{\centering\arraybackslash}m{4.0cm}|}{\cellcolor{gray!35}\textbf{Related Research Areas}} & \multicolumn{1}{>{\centering\arraybackslash}m{2.5cm}|}{\cellcolor{gray!35}\textbf{Source and Target Domains}} & \multicolumn{1}{>{\centering\arraybackslash}m{2.0cm}|}{\cellcolor{gray!35}\textbf{Source Domain Labels}} & \multicolumn{1}{>{\centering\arraybackslash}m{2.0cm}|}{\cellcolor{gray!35}\textbf{Target Domain Labels}} & \multicolumn{1}{>{\centering\arraybackslash}m{2.5cm}|}{\cellcolor{gray!35}\textbf{Source and Target Tasks}}\tabularnewline
			\hline 
			\hline
			\parbox[t]{2mm}{\multirow{3}{*}{Inductive TL}} 
			& Multi-task Learning & The same & Available	& Available & Different but related	\tabularnewline 	\cline{2-6} 
			& Self-taught Learning & The same & Unavailable & Available & Different but related		\tabularnewline 	\cline{2-6}
			\hline 
			\parbox[t]{2mm}{\multirow{1}{*}{Transductive TL}} 
			& Domain Adaptation, Sample Selection Bias and Covariate Shift & Different but related & Available & Unavailable & The same	\tabularnewline \cline{2-6} 
			\hline
			\parbox[t]{2mm}{\multirow{1}{*}{Unsupervised TL}} 
			&  & Different but related & Unavailable & Unavailable & Different but related	\tabularnewline \cline{2-6} 
			\hline 
		\end{tabular}
		\par\end{centering}
\end{table*}
\subsubsection{Inductive Transfer Learning}
In general, inductive TL can be defined as follows~\cite{pan_a_2010}:

\begin{definition}
	\textit{``Inductive TL: Given a source domain $\mathcal{D_\mathrm{S}}$ with a corresponding source task $\mathcal{T}_\mathrm{S}$ and a target domain $\mathcal{D}_\mathrm{T}$ with a corresponding target task $\mathcal{T}_\mathrm{T}$, inductive TL aims to enhance the learning of the target predictive function $f_\mathrm{T}(.)$ of target domain $\mathcal{D}_\mathrm{T}$ based on the knowledge gained in $\mathcal{D}_\mathrm{S}$ and $\mathcal{T}_\mathrm{S}$, in which the source task and the target task are different, i.e., $\mathcal{T}_\mathrm{S} \neq \mathcal{T}_\mathrm{T}$.''}
\end{definition}

It is worth noting that labeled data should be available at the target domain to guarantee good performance of the target predictive function. This is because the source and the target tasks are not the same. As such, learning only from the labeled data of the source domain may not provide good performance for the target domain. Inductive TL can be further divided into: (i) self-taught learning and (ii) multi-task learning based on the availability of labeled data in the source domain. In particular, when there is no labeled data in the source domain, inductive TL can be categorized as self-taught learning. The key idea of self-taught learning is to obtain a ``higher-level feature representation on the inputs'' based on the unlabeled data of the source domain~\cite{raina_self_2007}. After that, data in the target domain can be analyzed and classified in this representation. In other words, the unlabeled data of the source domain are used to reduce the dimension of the feature space, and the labeled data in the target domain are used for classification in the new feature space. In the case that both the source domain and the target domain have labeled data and they learn their tasks simultaneously, inductive TL can be categorized as multi-task learning~\cite{pan_a_2010}.

In the inductive TL, knowledge can be transferred to the target domain in the forms of instance, feature representation, parameters, and relational-knowledge. In particular, the instance-transfer approach aims to re-weight a part of labeled data in the source domain in order to reduce the differences in the source and target marginal distributions, and thus improving the learning process of the target domain~\cite{dai_boosting_2007},~\cite{dai_transferring_2007}. Differently, the feature-representation approach constructs a new feature representation and then transform both the source and target features to this new representation to mitigate domain divergence and errors~\cite{pan_a_2010}. Note that depending on the availability of labeled data at the source domain, several learning methods can be adopted to construct the new feature representation. For example, if the source domain has many labeled data, supervised learning techniques can be adopted~\cite{lee_learning_2007},~\cite{jebara_multi_2004}. In contrast, unsupervised learning can be used when the labeled data are not available at the source domain~\cite{wang_manifold_2008}. The parameter transfer approach is used if the models of related tasks have the same parameters or prior distribution of hyper-parameters~\cite{pan_a_2010}. Finally, the relational-knowledge transfer approach is applied if the data are not independent and identically distributed (i.i.d), such as networked and social network data~\cite{mihalkova_mapping_2007}.


\subsubsection{Transductive Transfer Learning}
Transductive TL can be defined as follows~\cite{pan_a_2010}:
\begin{definition}
	\textit{``Transductive TL: Given a source domain $\mathcal{D_\mathrm{S}}$ with a corresponding source task $\mathcal{T}_\mathrm{S}$ and a target domain $\mathcal{D}_\mathrm{T}$ with a corresponding target task $\mathcal{T}_\mathrm{T}$, transductive TL aims to enhance the learning of the target predictive function $f_\mathrm{T}(.)$ of target domain $\mathcal{D}_\mathrm{T}$ based on the knowledge gained in $\mathcal{D}_\mathrm{S}$ and $\mathcal{T}_\mathrm{S}$, in which the source domain and the target domain are different, i.e., $\mathcal{D}_\mathrm{S} \neq \mathcal{D}_\mathrm{T}$ and the source task and the target task are the same, i.e., $\mathcal{T}_\mathrm{S} = \mathcal{T}_\mathrm{T}$. Furthermore, unlabeled data must be available at the target domain when training.''}
\end{definition}

Transductive TL was first introduced in 2007~\cite{arnold_comparative_2007} to address the problem of TL in which labeled data are not available at the target domain. To ensure good learning performance, transductive TL requires all unlabeled data to be available when training. However, the authors in~\cite{pan_a_2010} point out that this requirement can be relaxed, and only a part of the unlabeled data are required at the training time. In transductive TL, there are two scenarios corresponding to the settings of feature spaces and marginal distributions as follows:
\begin{itemize}
	\item The feature spaces of the target domain and the source domain are different, i.e., $\mathcal{X}_\mathrm{S} \neq \mathcal{X}_\mathrm{T}$.
	\item The feature spaces of the target domain and the source domain are the same, i.e., $\mathcal{X}_\mathrm{S} = \mathcal{X}_\mathrm{T}$. However, their marginal distributions are not the same, i.e., $\mathcal{P}(X_\mathrm{S}) \neq \mathcal{P}(X_\mathrm{T})$.
\end{itemize}
These two settings are similar to domain adaptation and sample selection bias~\cite{pan_a_2010}. Domain adaptation aims to transfer knowledge learned from one or more source domains to enhance the learning performance of a target domain. Meanwhile, sample selection bias is used to mitigate the differences in data distributions of the source and target domains~\cite{zadrozny_learning_2004}. Similar to inductive TL, in transductive TL, the learned knowledge can be transferred to the target domain through feature representation-transfer, relational-knowledge-transfer, parameter-transfer, and instance-transfer approaches~\cite{pan_a_2010},~\cite{sangineto_we_2014}.


\subsubsection{Unsupervised Transfer Learning}
Unsupervised TL can be defined as follows~\cite{pan_a_2010}:
\begin{definition}
	\label{def:unsupervised}
	\textit{``Unsupervised TL: Given a source domain $\mathcal{D_\mathrm{S}}$ with a corresponding source task $\mathcal{T}_\mathrm{S}$ and a target domain $\mathcal{D}_\mathrm{T}$ with a corresponding target task $\mathcal{T}_\mathrm{T}$, unsupervised TL aims to enhance the learning of the target predictive function $f_\mathrm{T}(.)$ of target domain $\mathcal{D}_\mathrm{T}$ based on the knowledge gained in $\mathcal{D}_\mathrm{S}$ and $\mathcal{T}_\mathrm{S}$, in which the source task and the target task are different, i.e., $\mathcal{T}_\mathrm{S} \neq \mathcal{T}_\mathrm{T}$ and the labeled data $\mathcal{L}_\mathrm{S}$ and $\mathcal{L}_\mathrm{T}$ are not observable.''}
\end{definition}

From Definition~\ref{def:unsupervised}, it can be observed that unsupervised TL is adopted in the case that labeled data are not available at both the target and the source domains. Moreover, the target task is not the same as the source task. Compared to other TL approaches, unsupervised TL has little research work. However, recently, unsupervised TL has been getting tremendous attention from both industry and academia. The reason is that, in practice, there are many scenarios in which the labeled data are not readily available or impossible to obtain. In~\cite{dai_self_2008}, the authors propose a self-taught clustering method to improve the learning process of unsupervised TL. The key idea of this method is clustering the source data and the target data to obtain the common features such that the unlabeled data of the source domain can benefit the learning of the target domain. In~\cite{huang_unsupervised_2020}, the authors propose a self-supervised remedy approach to improve the learning performance of unsupervised TL by learning a discriminative space of the unlabeled data at the target domain.

\subsection{Deep Transfer Learning }
\begin{figure}[t]
	\centering
	\includegraphics[scale=0.7]{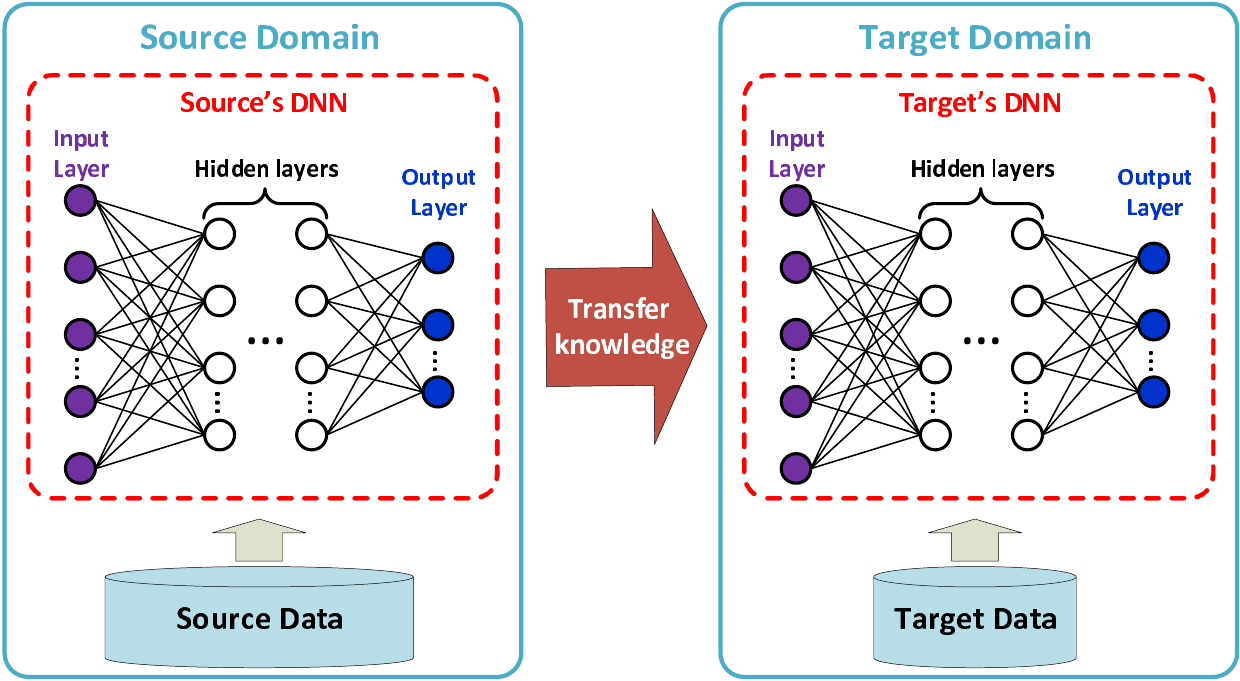}
	\caption{An illustration of DTL.}
	\label{Fig.Intro-DTL}
\end{figure}

DL is a sub-domain of ML. Inspired by the human brain structure, DL uses a multi-layered architecture called Deep Neural Network (DNN) consisting of an input layer, an output layer, and multiple hidden layers between them, as illustrated in Fig.~\ref{Fig.Intro-DTL}. DNN also intimates the human brain behaviors. Specifically, human learns how to perform a task, e.g., riding a bicycle, by practicing. The knowledge of this task is ``coded'' in the brain, and when the task is mastered, the brain will never forget this knowledge. Similarly, after being created, a DNN is trained to perform a specific task such as classification, clustering, or regression. During the learning process, the DNN updates its knowledge, represented by its parameters, according to the training data. 
After the learning phase, the DNN uses its knowledge to execute the trained task. Indeed, a trained DL model, including its architecture and parameters, can be considered as knowledge obtained from training data. Therefore, DL can be considered to be a powerful tool for TL to transfer knowledge between different domains, namely Deep Transfer Learning (DTL), as illustrated in Fig. \ref{Fig.Intro-DTL}. There are currently three typical strategies used in DTL, including Off-the-shelf Pre-trained Models, Pre-trained Models as Feature Extractors, and Fine-Tuning Pre-trained Models.
In the following, we review different DTL strategies and discuss their advantages/disadvantages in different applications.  
\subsubsection{Deep Transfer Learning Strategies}
\paragraph{Off-the-shelf pre-trained models}
Training DL models for complex tasks are data-hungry and time-consuming. For example, it may require a server equipped with 16 TPUv3 chips about three days to train the Bidirectional Encoder Representations from Transformers (BERT) model, a famous model developed by Google for Natural Language Processing (NLP). Note that a single TPUv3 chip is approximately 200 times faster than a Xeon Platinum 8163, one of the most powerful chips from Intel in 2017 when they are used to train ResNet50 \cite{wang_benchmarking_2019}. Thus, such a long training process is currently one of the main obstacles hindering DL advancement. In this context, instead of training a DNN from scratch, a pre-trained model, i.e., a model trained from a neighbor domain, can be directly leveraged for the target task. For example, suppose that one wants to perform an image recognition task based on the ImageNet dataset, the pre-trained models such as VGG, GoogLeNet, or ResNet50 can be used directly with reasonable accuracy.

The first fundamental requirement for directly using pre-trained models is their availabilities (i.e., whether these models exist or not). Fortunately, many pre-trained state-of-the-art models are shared freely in the community. These models span from various domains, e.g., computer vision and NLP. Most of the famous pre-trained models can be easily obtained from the Model Zoo section of the Caffe library \cite{model_zoo}. Alternatively, Keras, a popular DL library for Python, can provide another handy way to access these models \cite{model_keras}.

Although this approach can be straightforwardly implemented, the main disadvantage of this strategy is that the source and target domains, e.g., image datasets, have to be the same, and the source and target tasks, e.g., classifications, need to be similar. In practice, there is hardly a situation that pre-trained models can be directly used to perform target tasks because the target domain and the target task may be similar but not identical to the source domain and the source task, respectively. In addition, if the target and source tasks and domains are quite different. For example, if the source task is to classify different types of cars while the target task is to recognize whether an image displays a car or not, a pre-trained model may not be able to achieve a good result. Therefore, the off-the-shelf pre-trained model is less popular than other strategies.

\paragraph{Pre-trained models as feature extractors}
For traditional ML algorithms, such as Support Vector Machine (SVM), Decision Tree, and Logistic Regression, raw data cannot be used as input directly. It needs to be preprocessed to extract features. 
For example, to determine whether an image depicts a human face or not, we must manually identify the human face's unique features, such as shape, nose, and ear. Thanks to DL, the feature extraction step is entirely unnecessary since DNN can automatically learn these features. Leveraging this DL ability, a pre-trained model can be used as a feature extractor in the target domain data. In particular, target data will be passed to the pre-trained model to obtain a new feature representation before being used. As a result, this new representation embedded with the knowledge from the source domain can improve the learning process.

\begin{figure}[t]
	\centering
	\includegraphics[scale=1]{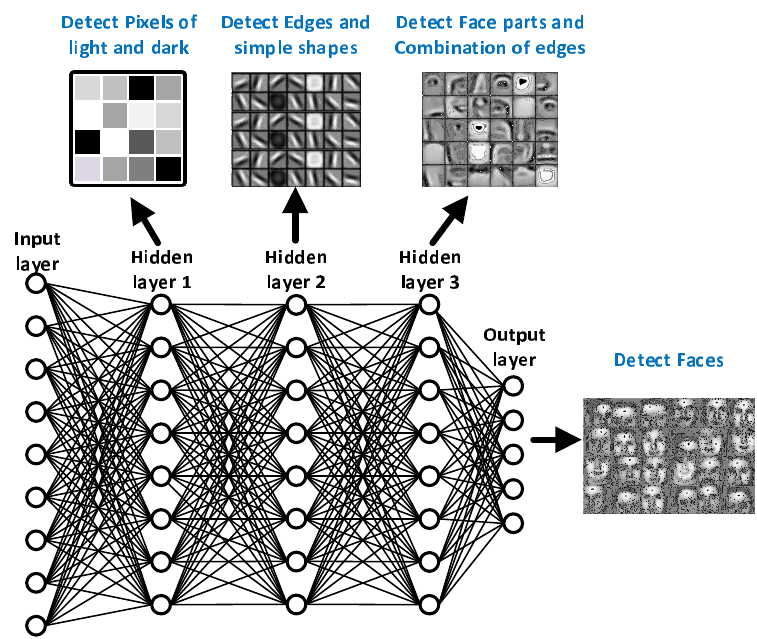}
	\caption{Hierarchical feature representations \cite{lecun_deep_2015}.}
	\label{Fig.Feature-learning}
\end{figure}

DL can also be considered to be a Learning Hierarchical Representation \cite{lecun_deep_2015} because it has the ability to learn hierarchical feature representations from a dataset. 
Specifically, an individual layer of DNN learns different features, from general, i.e., low-level, to more specific, i.e., high-level, as going deeper in DNN. 
It was proven that features learned by DL are more transferable, which means that they are easier to be reused across similar domains \cite{glorot_domain_2011,donahue_decaf_2014, yosinski_how_2014}.  
These studies also reveal that the higher-level features largely depend on a domain and a specific task, while lower-level features are more general. 
For example, as illustrated in Fig. \ref{Fig.Feature-learning}, when performing a face recognition, the layers closer to the input layer of DNN learn low-level features, e.g., pixels of light and dark. 
The middle layers learn more complex abstract features that combine these low-level features, e.g., edges and simple shapes. Finally, the layers closer to the output learn the high-level features related to specific tasks, e.g., nose, eye, and chin.

\begin{figure}[t]
	\begin{center}
		$\begin{array}{c}
			\includegraphics[scale=0.7]{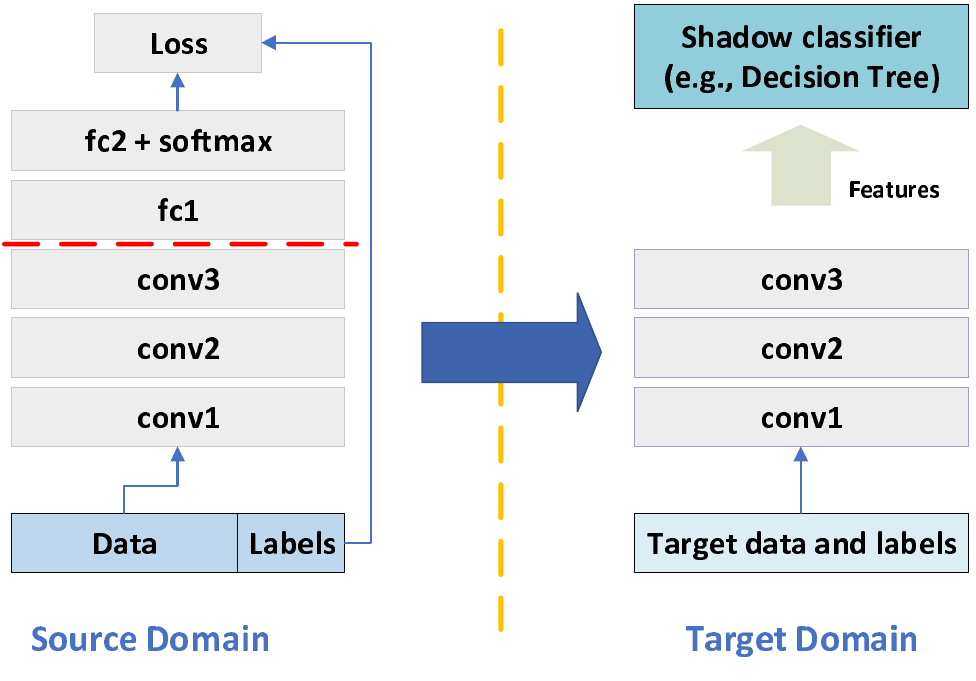} \\
			\fontsize{8}{0}\text{a) Standalone feature extractor.}\\
			\includegraphics[scale=0.7]{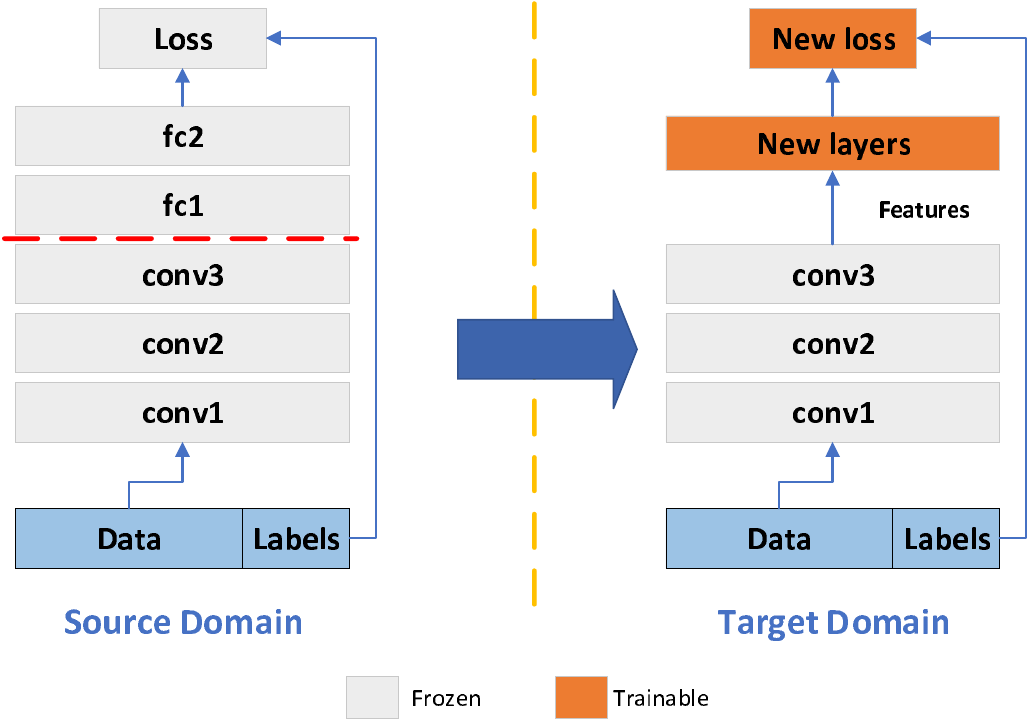} \\
			\fontsize{8}{0}\text{b) Integrated feature extractor.}
		\end{array}$			
		\caption{Pre-trained models as feature extractors.}
		\label{Fig.Feature-extractor}
	\end{center}
\end{figure}

The understanding of the DNN's ability allows us to choose appropriate levels of features from an existing pre-trained model. 
In particular, when the source and target tasks are quite different, it would be appropriate to use features obtained from the first few layers. 
Alternatively, if they are similar using the output of the deeper layers is more suitable. 
A pre-trained model could be used as a standalone or an integrated feature extractor.  
Fig.~\ref{Fig.Feature-extractor} shows examples of using a pre-trained model as a feature extractor. 
In the pre-train model, the first three layers are convolution (conv), following by two fully connected (fc) layers. Fig.~\ref{Fig.Feature-extractor}(a) is an example of a pre-trained model as a standalone feature extractor. 
The target data are fed into the first four layers of a pre-trained model to extract features, and then the output is used to train a shallow classifier, e.g., SVM. In the integration approach, illustrated in Fig.~\ref{Fig.Feature-extractor}(b), layers from a pre-trained model are integrated into a new DNN, but their parameters are not updated during the training process to preserve knowledge from the source model.

\paragraph{Fine-tuning pre-trained models}
Generally, the way of leveraging pre-trained models in this approach is similar to the second strategy, i.e., Pre-trained Models as Feature Extractors. 
However, instead of freezing all parameters of pre-trained models, certain parts or the whole pre-trained source model can continue to be trained with target data to improve the effectiveness of transfer knowledge further. 
The fine-tuning of pre-trained models can be carried out in two different ways as follows:
\begin{itemize}
	\item\textit{Weight Initialization:} The target model parameters are first initialized by the pre-trained model's parameters. 
	Then, the new model is trained with target domain data. Weight initialization is often applied when the target domain has plenty of labeled data.
	\item \textit{Selective Fine-tuning:} Recall that the lower layers in a DNN can learn general features (domain-independent), whereas the higher layers can learn more specific features (domain-dependent). 
	Based on this knowledge, we can determine how many layers of a pre-trained model should be tuned. 
	In particular, if the target data are small and the DNN has a massive number of parameters, more layers should be frozen to prevent an overfitting problem. 
	On the other hand, if target data are large and the number of parameters is small, more layers should be trained with new data. 
\end{itemize}

\begin{figure}[t]
	\begin{center}
		$\begin{array}{c}
			\includegraphics[scale=0.7]{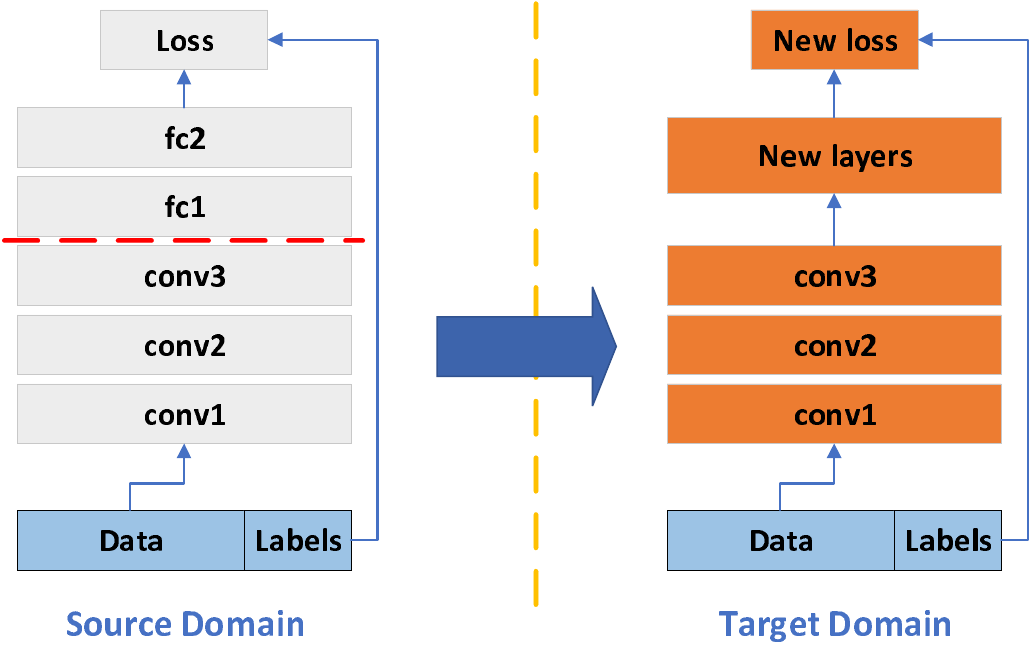} \\
			\fontsize{8}{0}\text{a) Weight initialization.}\\
			\includegraphics[scale=0.7]{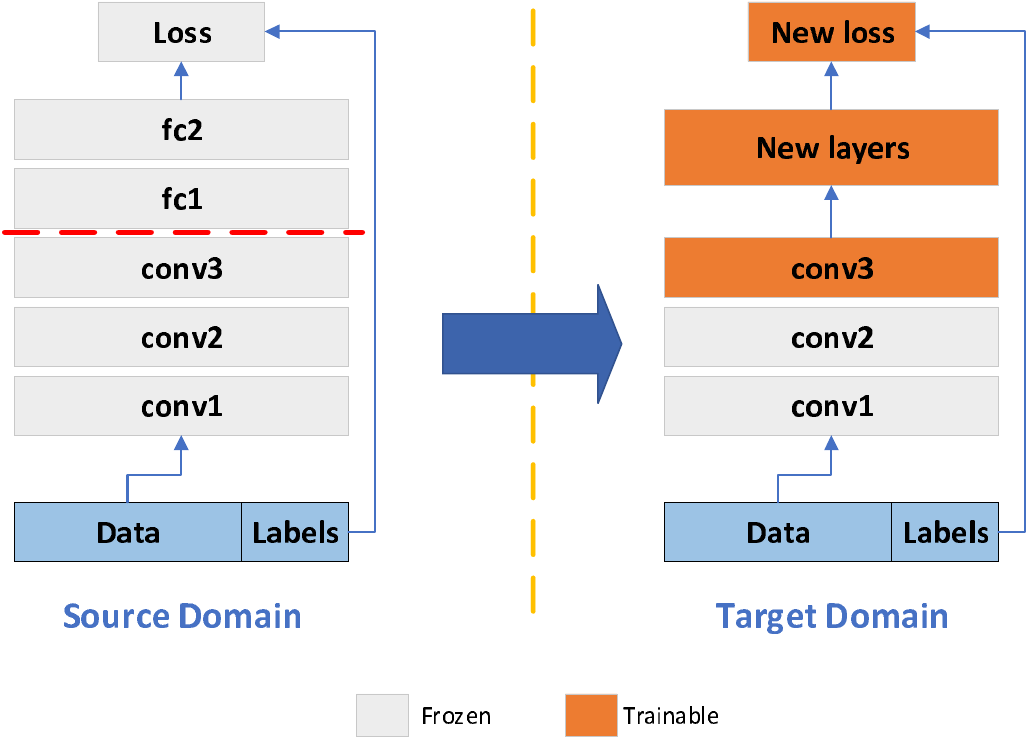} \\
			\fontsize{8}{0}\text{b) Selective fine-tuning.}
		\end{array}$
		\caption{Fine-tuning pre-trained models.}
		\label{Fig.Fine-tuning}
	\end{center}
\end{figure}

Figure~\ref{Fig.Fine-tuning} illustrates the two types of fine-tuning pre-trained models. The target model is created by replacing two top layers of the pre-trained model with new layers. For \textit{Weight Initialization}, as shown in Fig.~\ref{Fig.Fine-tuning}(a), both the pre-trained layers and new layers are trained with target data until the validation loss starts to increase. For \textit{Selective Fine-tuning} demonstrated in Fig.~\ref{Fig.Fine-tuning}(b), the first two layers are frozen during the training process in the target domain.

\subsubsection{How to select the best DTL strategy}
The comparisons of DTL strategies are provided in Table~\ref{DTL_strategies}. Choosing a suitable DTL strategy depends on several factors, among which the two most important elements are the target data size and the similarity between the source and target data. Some common rules are summarized as follows:
\begin{itemize}
	\item \textit{Target labeled data are small and similar to source data:} In this case, the fine-tuning should not be selected because the small data may lead to overfitting problems. Instead, since the source and target data are similar, the good choice is to use a pre-trained model as a feature extractor.
	\item \textit{Target data are small and very different from source data:} Since the data are small, we should choose the feature exactor approach. Moreover, it is better to transfer only the more general features from the first few layers because the source and target data are not similar.
	
	\item \textit{Target labeled data are large:} In this case, overfitting is not a big concern, and thus \textit{Weight Initialization} is an ideal choice.
\end{itemize}
\begin{table*}
	\caption{DTL Strategies} 
	\label{DTL_strategies}
	\begin{centering}
		\begin{tabular}{|>{\raggedright\arraybackslash}m{3cm}|>{\raggedright\arraybackslash}m{2cm}|>{\raggedright\arraybackslash}m{2cm}|>{\raggedright\arraybackslash}m{4.5cm}|>{\raggedright\arraybackslash}m{4cm}|}
			\hline 
			\multicolumn{1}{|>{\centering\arraybackslash}m{3cm}|}{\cellcolor{gray!35}\textbf{DTL Strategy}} &  \multicolumn{1}{>{\centering\arraybackslash}m{2cm}|}{\cellcolor{gray!35}\textbf{Source and Target Domains}} & \multicolumn{1}{>{\centering\arraybackslash}m{2cm}|}{\cellcolor{gray!35}\textbf{Source and Target Tasks}} & \multicolumn{1}{>{\centering\arraybackslash}m{4.5cm}|}{\cellcolor{gray!35}\textbf{Advantages}} & \multicolumn{1}{>{\centering\arraybackslash}m{4cm}|}{\cellcolor{gray!35}\textbf{Disadvantages}}\tabularnewline
			\hline 
			\hline
			Off-the-shelf Pre-trained Models 
			& Identical & Similar	& - Simple for implementation. \newline - No need to train with target data. 
			& It is hard to obtain source model since the target and source domain must be the same.	\tabularnewline 	\cline{2-5} 
			\hline 
			Pre-trained Models as Feature Extractors 
			& Similar & Different but related & - Flexible in model selection. \newline - No need to train with target data.
			& Challenge to select a level of feature.	\tabularnewline \cline{2-5} 
			\hline
			Fine Tuning Pre-trained Models 
			& Different but related & Different & Able to utilize models from different domains. & Need adequate target's labelled data to fine-tune the source model.	\tabularnewline \cline{2-5} 
			\hline 
		\end{tabular}
		\par\end{centering}
\end{table*}

\subsection{Transfer Learning in Deep Reinforcement Learning }
In this section, we first provide fundamental background for DRL and then discuss how TL can be applied to improve the performance of DRL algorithms.
\subsubsection{Deep Reinforcement Learning}
\paragraph{Markov decision process (MDP)} MDP is widely used to formulate dynamic and stochastic decision-making problems. Typically, an MDP is determined by four elements, including a state space $\mathcal{S}$, an action space $\mathcal{A}$, transition probabilities $\mathcal{P}$, i.e., the probability that a state $s_t$ at time $t$ moves to state $s_{t+1}$ at time $t+1$ after action $a_t$ is executed at time $t$, and a reward function $\mathcal{R}: S \times A \rightarrow \mathbb{R}$ that returns an immediate reward $r_t$ after performing action $a_t$ at state $s_t$. 
A mapping from the state space to the action space is called a policy, denoted by $\pi$.
The objective of MDP is to find an optimal policy $\pi^*$ that maximizes an expected discounted total reward, i.e., $\pi^*=\max_\pi\mathbb{E}_\pi [\sum_{t=0}^{T}\gamma^t r_t(s_t, \pi(s_t))]$ where $a_t=\pi(s_t)$, $\mathbb{E}[\cdot]$ is the expectation function, and $\gamma \in [0,1)$ is the discount factor representing the importance of future rewards. In particular, the larger the discount factor is, the more important future rewards are. In practice, due to outstanding abilities to deal with uncertainties in intelligent systems, MDP has been widely adopted to address various problems in dynamic wireless environments, such as spectrum management, cognitive radios, wireless security, and power control~\cite{alsheikh_markov_2015}.

\paragraph{Reinforcement learning (RL)} RL is a unique class of ML. An RL agent operates in a dynamic environment formulated by an MDP framework, and its goal is to learn an optimal policy to maximize the expected discounted total reward. During the learning process, the agent interacts with the environment and observes results to gradually find its optimal policy. 
Specifically, at each time step $t$, the agent observes current state $s_t$, performs action $a_t$ according to its current policy, then receives an intermediate reward $r_t$ and moves to new state $s_{t+1}$.
After that, the agent adjusts its policy based on the feedback of the environment, i.e., $r_t $ and $s_{t+1}$. This procedure is repeated until the agent's policy converges to the optimal one.

In practice, Q-learning is one of the most widely used RL algorithms. This algorithm requires evaluating a state-action value function $\mathcal{Q}^\pi(s,a)$, also called Q-function, that specifies how good of performing an action $a$ at a state $s$ under policy $\pi$. The value of each pair of state and action is called Q-value.
The Q-function under the optimal policy $\pi^*$ is called the optimal state-value function $\mathcal{Q}^*(s,a)$. 
Suppose that values of $\mathcal{Q}^*(s,a)$ for all state-action pairs $(s,a)$ are known, the RL agent can obtain the optimal policy at state $s$ by simply taking an action that maximizes $\mathcal{Q}^*(s,a)$ \cite{sutton_reinforcement_1998}.
The Q-learning employs a table, namely Q-table, to store and update Q-values. 
Specifically, each cell in the Q-table stores an estimation of the Q-value for each state-action pair.
Based on interactions between the agent and the environment at time $t$, i.e., action $a_t$, intermediate reward $r_t$, and next state $s_{t+1}$, the Q-function is updated using the temporal difference (TD), which is the difference between target Q-value, i.e., $Y_t=r_t(s_t, a_t) + \gamma\max_{a_{t+1}} \mathcal{Q}_t(s_{t+1}, a_{t+1})$, and the current estimated Q-value, i.e., $\mathcal{Q}_t(s_t,a_{s_t})$, as follows:	
\begin{equation}
	\label{Eq:updateQfunction}
	\begin{aligned}
		\mathcal{Q}_{t}(s_t,a_t) \leftarrow &\mathcal{Q}_t(s_t,a_t) + 
		\zeta_t\Big [ Y_t - \mathcal{Q}_t(s_t,a_{s_t})\Big ],
	\end{aligned}
\end{equation}
where $\zeta_t$ is the learning rate that represents the impact of new interaction, i.e., TD. 
If the Q-function is updated by \eqref{Eq:updateQfunction} and the learning rate $\zeta_t$ satisfies conditions in \eqref{Eq:rules}, it is proven that the policy learned by the Q-learning will converge to the optimal policy \cite{watkins_qlearning_1992}. 
\begin{equation}
	\label{Eq:rules}
	\zeta_t \in [0,1), ~\sum_{t=1}^{\infty}\zeta_t = \infty, \mbox{ and } \sum_{t=1}^{\infty} ( \zeta_t  )^{2} < \infty.
\end{equation}
It is worth noting that although the convergence of Q-learning is proven, this algorithm can be inefficient in the case of high dimensional state and action spaces since it uses a table for estimating $\mathcal{Q}^*(s,a)$. Thus, DRL has been introduced recently as a highly-effective solution to address the current limitations of RL algorithms. 

\paragraph{Deep reinforcement learning}In a high dimension environment, which has enormous numbers of states and actions, the traditional RL methods, e.g., Q-learning, may not handle it effectively. 
For example, considering the problem of designing an RL-based agent to play various video games without knowing the rules in advance, states can be represented by  images consisting of millions of pixels, making it impractical to construct a Q-table. Furthermore, it is inefficient to learn all state-action values in all states separately as that of the Q-learning algorithm.	
These challenges can be addressed by leveraging DNN architecture for RL algorithms, i.e., DRL. The existing DRL methods can be grouped into value- and policy-based categories. 

In the value-based category, an agent first needs to learn a value function, e.g., the Q-function, then finds the optimal policy based on this function. Most of the value-based DRL methods rely on the Q-learning algorithm, namely Deep Q-learning. 
Deep Q-learning employs a DNN instead of a Q-table to learn the Q-function $\mathcal{Q}(s, a)$. This DNN takes a state as an input and outputs Q-values for all actions so that the number of activations in the output layer equals the size of the action space. 
However, RL is unstable and can diverge if the Q-function is estimated by a non-linear function approximator such as a DNN because data obtained from the MDP are high correlation \cite{mnih_human_2015}. 
Thus, we can address this problem by applying two techniques:
\begin{itemize}
	\item \textit{Experience Replay:} At time $t$, instead of only using current transition, i.e., $(s_t, a_t, r_t, s_{t+1})$, a mini-batch of previous transitions is sampled uniformly and randomly from a memory buffer $\textbf{B}$ to train the DNN. Accordingly, the correlation in data can be loosen. 
	\item \textit{Quasi-static target Q-network:} This technique employs two DNNs, the Q-network $\mathcal{Q}(s,a; \phi)$ for estimating Q-value and the target Q-network $\hat{\mathcal{Q}}(s,a; \phi^-)$ for estimating target Q-value.
	We denote $\phi$ and $\phi^-$ as parameters of the source and target Q-network, respectively. The Q-network is trained at every time step, while the target Q-network's parameters $\phi^-$ will be copied from $\phi$ at every predefined number of time steps. Therefore, the learning process is more stable.
\end{itemize}

The Q-network is trained to minimize the gap between the estimated Q-value, i.e.,~$\mathcal{Q}(s_t,a_t;\phi_t)$, and target Q-value, i.e.,~$Y_t = r_t + \gamma \max_{a_{t+1}}\hat{\mathcal{Q}}\big(s_{t+1}, a_{t+1};\phi^-_t\big)$, so that its loss function is calculated as follows:
\begin{equation}
	\begin{aligned}
		\label{DQL-lossfunction}
		L_t(\phi_t) = \mathbb{E}_{(s,a,r,s')\sim U(\mathbf{B})}\bigg[ \bigg( Y_t
		-\mathcal{Q}(s,a;\phi_t)\bigg)^2\bigg].
	\end{aligned}
\end{equation}

The loss function in \eqref{DQL-lossfunction} can be minimized by Gradient Descent algorithm (GD), in which the Q-network's parameters are iteratively updated as follows:
\begin{equation}
	\label{eq:GDupdate}
	\phi_{t+1} = \phi_t -\eta_t\nabla L_t(\phi_t),
\end{equation}
where $\nabla(\cdot)$ is the gradient of a function, $\eta_t>0$ is a step size at step $t$. To address the sampling noises, $\eta_t$ can be decreased gradually to ensure the convergence.	Suppose that the training process is complete, the optimal policy is given according to ${\operatorname{argmax}}_a \mathcal{Q}(s, a)$. 

\begin{figure}[t]
	\begin{center}
		$\begin{array}{c}
			\includegraphics[scale=0.35]{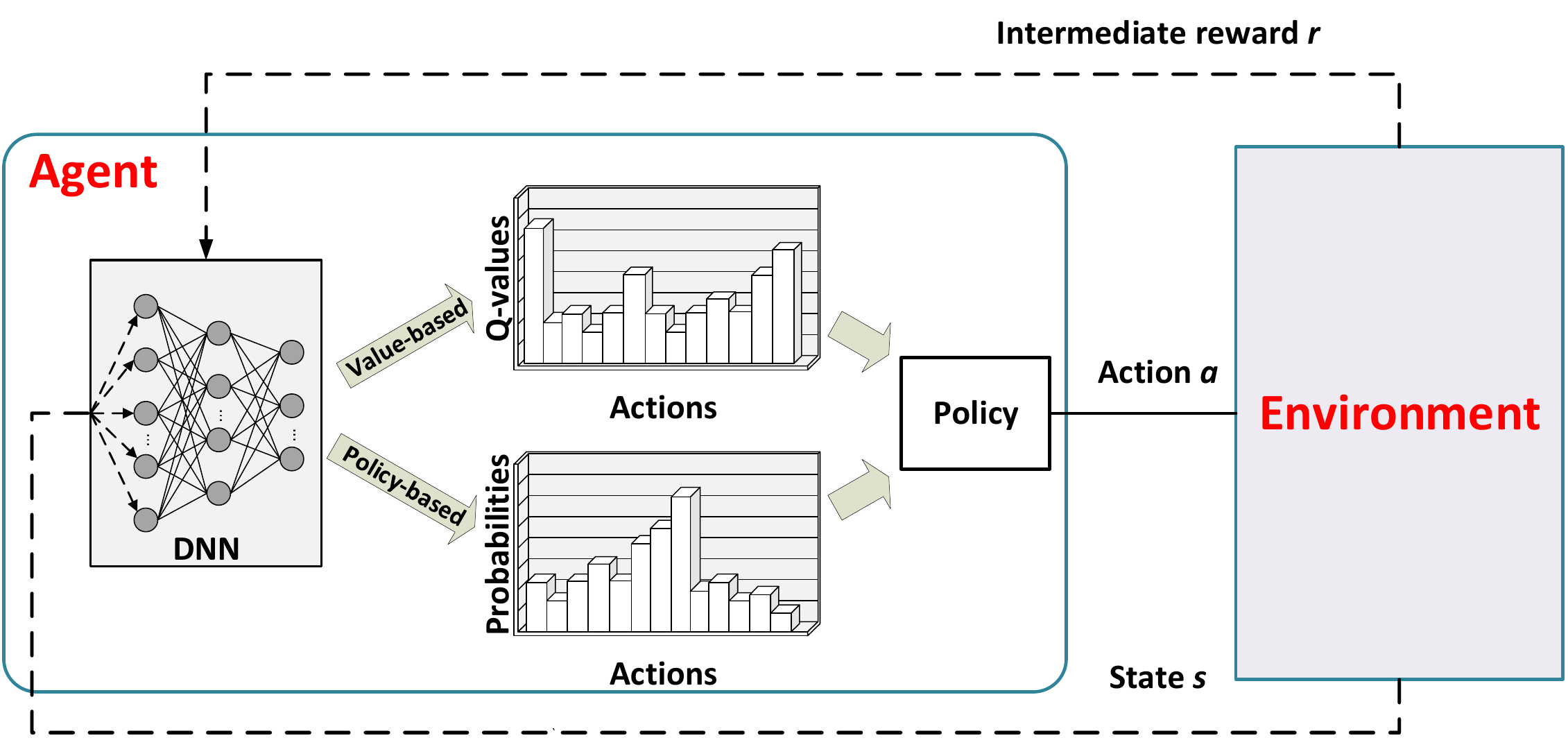} \\
		\end{array}$
		\caption{DRL process.}
		\label{Fig.DRL}
	\end{center}
\end{figure}

Although value-based DRL methods can work very well in high dimension state spaces, it may take a lot of time to calculate Q-values for all possible actions, making it impractical if the given action space is very large or continuous.
Fortunately, the policy-based DRL methods can address this problem by directly learning the optimal policy without estimating any value function. In particular, the agent starts with an inaccurate policy represented by a DNN, denoted by $\pi(a_t|s_t;\theta_t)$ where $\theta_t$ denotes parameters of the DNN at time $t$, and then updates its policy based on interactions with the environment to achieve the best performance. Specifically, a performance according to a policy $\pi$ can be defined as follows:
\begin{equation}
	\mathcal{J}(\theta)=\sum_{s \in \mathcal{S}}d^{\pi(\theta)}(s)V^{\pi(\theta)}(s),
\end{equation}
where $d^{\pi(\theta)}$ is the steady-state distribution under policy $\pi(\theta)$, and  $V^{\pi(\theta)}(s)$ is a state value function under policy $\pi(\theta)$ that specifies how good to be of a state \cite{sutton_reinforcement_1998}. The DNN is trained to maximize the performance function, so that Gradient Ascent (GA) can be used to iteratively update the DNN's parameters by
\begin{equation}
	\theta_{t+1} = \theta_t +\eta_t\nabla \mathcal{J}(\theta_t).
\end{equation}
The policy-based DRL diagram is similar to that of value-based methods, as illustrated in Fig. \ref{Fig.DRL}. However, the DNN in the policy-based methods performs a classification task instead of a regression task in value-based methods. In particular, the DNN in policy-based DRL methods takes a state as an input and outputs the probability distribution of actions. Then, actions are selected based on this distribution.

\subsubsection{Transfer Learning Strategies in Deep Reinforcement Learning}
DRL is one of the most data-hungry and time-consuming ML types. In particular, at the beginning of the learning process, an agent may spend a lot of time to explore the environment before acquiring an optimal policy. Therefore, the TL techniques can be used to speed up the learning process. However, the DRL's environment is formulated by the MDP framework making it complicated to apply TL. Specifically, the differences between source and target in DRL can be from any MDP elements, i.e., the state space, the action space, the reward function, the transition dynamics, or the task objective. Recall that in Section II.A, TL definition is based on the domain and task. In DRL, data are unlabelled and obtained from interactions between the agent and the environment, and the task is embedded in the MDP. Therefore, the MDP can be referred to both the domain and the task. Thus, TL in DRL can be defined as follows\cite{zhu_transfer_2020}:

\begin{definition}[]
	\label{def.transfer}
	\textit{``Transfer Learning in DRL: Given a source MDP $\mathcal{M}_\mathrm{S}$ and a target MDP $\mathcal{M}_\mathrm{T}$, the goal of TL in DRL (TL-DRL) is learning the optimal policy $\pi^*_\mathrm{T}$ for the target MDP by leveraging the source's knowledge $\mathcal{D}_\mathrm{S}$, i.e., the data, the policy, and the environment dynamics, along with the target information $\mathcal{D}_\mathrm{T}$:}
	\begin{equation}
		\pi^* = \underset{\pi_\mathrm{T}}{\operatorname{argmax}}~ \mathbb{E}_{s\sim S_\mathrm{T}, a \sim \pi_\mathrm{T}}[ Q^{\pi_\mathrm{T}}(s,a)],
	\end{equation}
	\textit{where $\pi_\mathrm{T} : S_\mathrm{T}  \rightarrow A_\mathrm{T}$ is a policy of $\mathcal{M}_\mathrm{T}$ and estimated by a DNN trained on both $\mathcal{D}_\mathrm{S}$ and $\mathcal{D}_\mathrm{T}$.''} 
\end{definition}
Based on the form of knowledge to be transferred, the existing strategies for TL-DRL can be categorized into the following groups:
\paragraph{Reward shaping}
In \textit{Reward Shaping} (RS), the prior knowledge from the source MDP is embedded in an additional reward function to adjust the target reward function \cite{ng_policy_99, wiewiora_principled_2003}. In particular, RS uses prior knowledge to design an additional reward function $F$ for target MDP that generates extra rewards along with the environment rewards. The goal of the additional rewards is to guide the agent to select a better action. For example, an action that leads to a good state will have a higher additional reward to direct the agent to the desired policy. As a result, the agent will receive rewards generated from a new reward function $R_{RS} = R + F $ and learn a new optimal policy for a new target MDP in which the reward function is modified. Therefore, RS technique is generally applicable to different DRL algorithms because it only requires minimum changes to the DRL framework.
\paragraph{Learning from demonstrations}
In this approach, the source knowledge forms a demonstration defined by a transition tuple $(s, a, s', r)$. A demonstration describes the interactions between the agent and the environment, i.e., the agent moves to state $s'$ and receives reward $r$ by performing action $a$ at state $s$. Demonstrations are obtained through interactions between the agent and environment under a policy that could be optimal, near-optimal, or even suboptimal. Learning from demonstrations is a method aiming to help the agent efficiently explore the environment by utilizing the demonstrations from the source MDP. It can be done in two fashions, online and offline. In the offline approach, target DRL components, e.g., value function, policy, or environment dynamics model, are initialized by using supervised DL with the demonstrations in $\mathcal{M}_\mathrm{S}$ before the learning phase in $\mathcal{M}_\mathrm{T}$. In contrast, the online methods use the source demonstrations along during the learning phase to direct the agent's actions.
\paragraph{Policy transfer}
In this approach, the optimal policy from the source MDP will be transferred to the target MDP. Two types of policy transfer are \textit{Policy Distillation} and \textit{Policy Reuse}.
\begin{itemize}
	\item \textit{Policy Distillation:} The distillation term is introduced by \cite{hilton_distilling_2015} to refer a method for compressing the knowledge ensemble from different source models to a single target model in DL. This method then is extended to DRL with the name \textit{Policy Distillatio}n, in which policies from one or more source tasks are distilled to a policy for the agent in the target MDP. Typically, a target policy is acquired by minimizing the discrepancy of action distributions between the source policy and the target policy. Because the target model is often shallower than sources models and capable of executing multiple source tasks, \textit{Policy Distillation} is also a compelling way for compressing model and multi-task RL. 
	\item \textit{Policy Reuse:} Instead of distillation from multiple policies, the second approach aims to reuse sources' policies for the target task directly. \textit{Policy Reuse} guides the target agent's learning process according to how the source policies perform on the target MDP. Specifically, the policies, including source policies and current target policy, are directly reused based on a probability distribution $Prb$, where the possibility of an individual policy depends on its performance, denoted by $G_i$ in the target MDP as follows:
	\begin{equation}
		Prb(\pi_i) = \frac{\exp(tG_i)}{\sum^{N}_{j=0}\exp(tG_j)},
	\end{equation}
	where $N$ is the number of policies, $0 \leq i \leq N$, and $t$ is a parameter that increases over time. Note that $G_0$ is the current target policy.  Typically, Q-function can be used to evaluate the performance of a policy \cite{Fernandez_probabilistic_2006, Barreto_successor_2017}.
\end{itemize}	
\paragraph{Inter-task mapping}
To facilitate TL, this approach uses functions to map between the source and target MDPs. Existing studies assume that there is a one-to-one mapping function between the source and target. The subjective of mapping can be a subset or an entire state space \cite{gupta_learning_2017,konidaris_autonomous_2006,ammar_reinforcement_2011} or the representation of the transition dynamic \cite{ammar_reinforcement_2012}, i.e., $(s, a, s')$.  
Typically, in inter-task mapping over states, the source state space is split into agent-specific and task-specific representations. Then, the agent-specific space is used to learn the mapping function, and the mapped states are leveraged to reshape the reward function in the target MDP. To preserve source knowledge as much as possible, a DNN based on autoencoder architecture is utilized to learn the mapping function \cite{gupta_learning_2017}. In inter-task mapping over transition representation, both source and target transition representation are first mapped to a higher dimension. Then, a Sparse Pseudo-Inputs Gaussian Processes is leveraged to learn the mapping function \cite{ammar_reinforcement_2012}. However, the second approach often relies on assumptions that the source and target are similar in terms of the transition dynamic and the state representation. 
\begin{table*}[ht]
	\caption{TL strategies in DRL} 
	\label{TL-DRL_strategies}
	\begin{centering}
		\begin{tabular}{|>{\raggedright\arraybackslash}m{3.5cm}|>{\raggedright\arraybackslash}m{8.5cm}|>{\raggedright\arraybackslash}m{4cm}|}
			\hline 
			\multicolumn{1}{|>{\centering\arraybackslash}m{3.5cm}|}{\cellcolor{gray!35}\textbf{TL-DRL Strategy}} &  \multicolumn{1}{>{\centering\arraybackslash}m{8.5cm}|}{\cellcolor{gray!35}\textbf{MDP Difference}} & \multicolumn{1}{>{\centering\arraybackslash}m{4cm}|}{\cellcolor{gray!35}\textbf{Knowledge to transfer}} \tabularnewline
			\hline 
			\hline
			\parbox[t]{2mm}{\multirow{1}{*}{Reward Shaping}} 
			& $\mathcal{M}_\mathrm{S}=\mathcal{M}_\mathrm{T}$ & Reward function	\tabularnewline 	\cline{2-3} 
			\hline 
			\parbox[t]{2mm}{\multirow{1}{*}{Learning from Demonstrations}} 
			& $\mathcal{M}_\mathrm{S}=\mathcal{M}_\mathrm{T}$ & Demonstrations, i.e., $(s,a,s',r)$  \tabularnewline \cline{2-3} 
			\hline
			\parbox[t]{2mm}{\multirow{1}{*}{Policy Transfer}} 
			& $\mathcal{S}_\mathrm{S} \neq \mathcal{S}_\mathrm{T}$ and $\mathcal{A}_\mathrm{S} \neq \mathcal{A}_\mathrm{T}$; $\mathcal{R}_\mathrm{S} \neq \mathcal{R}_\mathrm{T}$
			& Source agent's policy \tabularnewline \cline{2-3} 
			\hline 
			\parbox[t]{2mm}{\multirow{1}{*}{Inter-task Mapping}} 
			&$\mathcal{S}_\mathrm{S} \neq \mathcal{S}_\mathrm{T}$;
			$\mathcal{S}_\mathrm{S} \neq \mathcal{S}_\mathrm{T}$ and $\mathcal{A}_\mathrm{S} \neq \mathcal{A}_\mathrm{T}$;
			$\mathcal{R}_\mathrm{S} \neq \mathcal{R}_\mathrm{T}$ and $\mathcal{A}_\mathrm{S} \neq \mathcal{A}_\mathrm{T}$;
			$\mathcal{S}_\mathrm{S} \neq \mathcal{S}_\mathrm{T}$ and $\mathcal{R}_\mathrm{S} \neq \mathcal{R}_\mathrm{T}$;
			$\mathcal{S}_\mathrm{S} \times \mathcal{A}_\mathrm{S}  \neq \mathcal{S}_\mathrm{T} \times \mathcal{A}_\mathrm{T}$
			& Source MDP  \tabularnewline \cline{2-3} 
			\hline 
			\parbox[t]{2mm}{\multirow{1}{*}{Reusing representations}} 
			& $\mathcal{S}_\mathrm{S} \neq \mathcal{S}_\mathrm{T}$;
			$\mathcal{S}_\mathrm{S} \neq \mathcal{S}_\mathrm{T}$ and $\mathcal{A}_\mathrm{S} \neq \mathcal{A}_\mathrm{T}$;
			$\mathcal{R}_\mathrm{S} \neq \mathcal{R}_\mathrm{T}$ and $\mathcal{A}_\mathrm{S} \neq \mathcal{A}_\mathrm{T}$;
			$\mathcal{S}_\mathrm{S} \neq \mathcal{S}_\mathrm{T}$ and $\mathcal{R}_\mathrm{S} \neq \mathcal{R}_\mathrm{T}$;
			$\mathcal{S}_\mathrm{S} \times \mathcal{A}_\mathrm{S}  \neq \mathcal{S}_\mathrm{T} \times \mathcal{A}_\mathrm{T}$ & Representations of states or transition dynamics \tabularnewline \cline{2-3} 
			\hline
		\end{tabular}
		\par\end{centering}
\end{table*}
\paragraph{Reusing representations}
Instead of learning a mapping function between source and target MDPs, state and action representations can be directly 	reused in the target or indirectly through a task-invariant feature space. The first way in this strategy is conducted progressively by using a progressive neural network architecture  \cite{rusu_progressive_2016} consisting of a sequence of DNNs such that each DNN is trained for a particular task. Specifically, the first DNN learns the first task, then the second DNN is added to the sequence for learning the second task, and so on. While training a new task, the previous DNNs in the sequence are used as feature extractors to produce new representations. As a result, the size of the progressive network grows with the new tasks making it not scalable. The second way can address this disadvantage by using a modular neural network architecture with a fixed size. The authors in \cite{fernando_pathnet_2017} propose the PathNet, in which neurons are divided into pathways.
Each pathway is analog to a DNN in the progress architecture. In particular, a new pathway is created by randomly selecting neurons to learn the new task. The new pathway's parameters are only updated during the training if they are not in the previous pathways.  The modular architecture is also employed to split the policy network into a task- and environment-specific module\cite{devin_learning_2017}. The idea behind this division is the task-specific module can be applied to different environments where the task is performed, while the environment-specific module can be applied to different tasks in the same environment.	
\subsubsection{How to select the best TL-DRL strategy}
In comparison with DRL, TL-DRL is indeed more complex since it involves MDPs. To select an appropriate TL-DRL scheme, it is critical to determine the differences between $\mathcal{M}_\mathrm{S}$ and $\mathcal{M}_\mathrm{S}$, which can be in the state space, the action space, the reward function, and the environment dynamics, because this information will define the type of knowledge that can be transferred. Another important aspect needed to be considered is whether a specific TL-DRL method restricts a type of RL algorithm. For example, the authors in \cite{hester_deep_2018} propose the Deep Q-learning from the Demonstration framework that can be applied only for Deep Q-Learning. More comparisons for the aforementioned TL-DRL strategies can be found in Table~\ref{TL-DRL_strategies}.



\section{Applications of Transfer Learning for Spectrum Management}\label{sec:spectrum}
As the demands for frequency spectrum are continuously growing, effective spectrum management becomes a critical issue for wireless networks. Traditional ML techniques have been widely applied in spectrum management for various tasks, such as resource allocation, channel estimation, and spectrum sensing. Nevertheless, there are two main obstacles that hinder the effectiveness of the traditional ML techniques. First, the radio environment is subject to significant variations. This negatively impacts the performance of the ML models which are trained only for a specific scenario. Second, many spectrum management tasks are time-sensitive, and thus there is much less time available to train the ML models. To address these problems, TL has emerged to be an effective solution which can utilize the knowledge gained from similar scenarios in order to achieve highly-effective learning processes.

\subsection{Cognitive Radio Networks}
Cognitive Radio Network (CRN) is an effective solution to the ever-growing demands of spectrum-based communications. The backbone of CRN is the cognitive radio techniques that can make optimal spectrum sharing decisions under dynamics and uncertainties of radio environments. To enable such intelligent and adaptive decision makings, conventional ML techniques have been leveraged for various tasks such as spectrum sensing, handoff prediction, and interference management. However, these ML techniques require a lot of training time and data to be effective for a specific task, which is disadvantageous in the ever-changing radio environments. To improve the effectiveness and robustness of these ML techniques, various TL approaches have been proposed to utilize knowledge from different but related scenarios, i.e., nearby devices or regions. 
\subsubsection{Spectrum Sensing}
\begin{figure}[!t]
	\centering
	\includegraphics[width=\columnwidth]{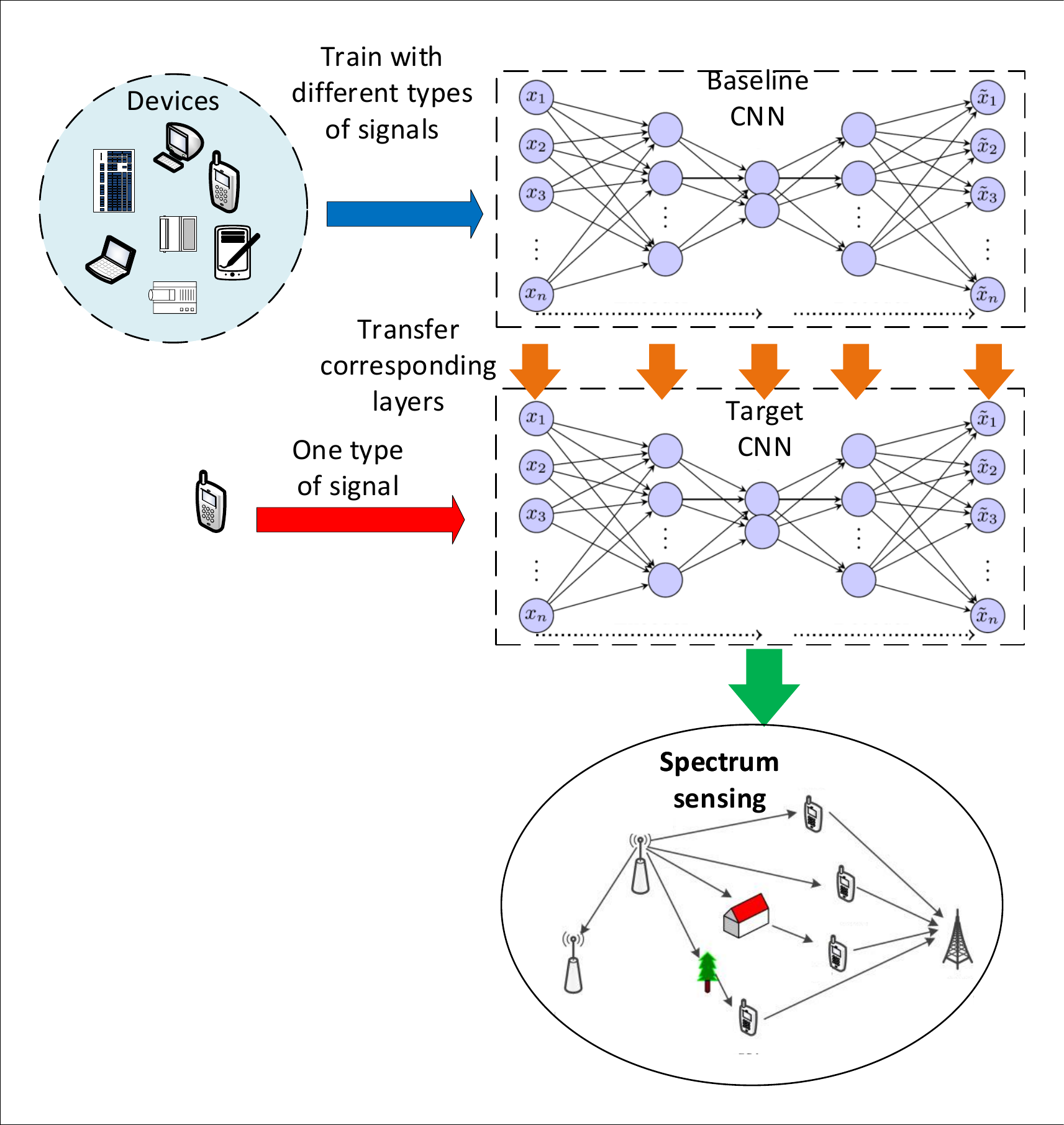}
	\caption{The TL-based spectrum sensing approach~\cite{peng_robust_2019}.}
	\vspace{-0.5em}
	\label{fig:sm1}
\end{figure}

Spectrum sensing refers to the techniques to allow the Secondary Users (SUs) to detect activities of Primary Users (PUs) on the primary channels, and thereby effectively utilizing the channels under the presence of the PUs~\cite{ali_advances_2016}. Although the traditional ML approaches have been considered effective solutions for spectrum sensing, the sensing performance is significantly reduced when the domains change, i.e., another type of signal or environment. Thus, transductive TL approaches are proposed in~\cite{peng_robust_2019} and~\cite{zheng_spectrum_2020} to utilize the knowledge gained from other signal types and environments. Particularly, as illustrated in Fig.~\ref{fig:sm1}, the authors in~\cite{peng_robust_2019} propose a \textit{Fine-tuning Pre-trained Models} TL strategy for spectrum sensing in CRNs. Particularly, a baseline Convolutional Neural Network (CNN) model trained with various types of signals data is transferred to a target task (one type of signal), and then the model is fine-tuned using the target data. Simulation results show that TL can improve the performance, i.e., detection probability and false-alarm rate, of the CNN by up to 25\%. However, in the cases of limited target data, the proposed approach performs worse than other non-ML methods, e.g., energy detectors. Similarly, a TL approach is proposed in~\cite{zheng_spectrum_2020}, which also employs the \textit{Fine-tuning pre-trained models} TL strategy. Specifically, the approach transfers a CNN trained with different types of signal data and noise data to the target task and fine-tune the model with target data. This TL approach is very similar to the one in~\cite{peng_robust_2019}, except that~\cite{peng_robust_2019} does not consider noise data. Experimental results show that the proposed TL process can improve the detection probability of the CNN by up to 30\% under low signal-to-noise ratio (SNR) levels.

Different from the approaches in~\cite{peng_robust_2019} and~\cite{zheng_spectrum_2020} that utilize knowledge from different types of signals, an inductive TL process is proposed in~\cite{pati_deep_2020} that leverages knowledge from different environments. Specifically, a four-layer CNN is developed to extract features, and a linear SVM is responsible for classifying signals based on the extracted features. To improve the robustness of the proposed approach, a CNN trained with data from a specific environment is transferred to another environment to train with the target data, while the first two layers of the CNN are frozen. The reason for freezing only the first two convolutional layers is because these layers learn the more general features which are common across different domains. Experimental results show that the TL process can significantly improve the training time and sensing time of the CNN by up to 94\% and 20\%, respectively. 
\subsubsection{Channel Selection}
In overlay CRN, the SUs can choose to switch to another channel to avoid interference with the PUs at the current channel~\cite{tsiropoulos_radio_2014}. To reduce the delay of such switching process, the SUs can predict when a channel will be occupied by the PUs and then select idle channels for switching in advance. The first step, spectrum prediction, is an important but challenging task. To predict future spectrum state, historical data can be leveraged by conventional ML techniques. However, for a certain spectrum band, historical data might be scarce, which significantly reduces the prediction accuracy. Therefore, a TL approach is proposed in~\cite{lin_cross_2020}, which utilizes historical data from other spectrum bands. Particularly, a two-layer long short-term memory (LSTM) model is developed to predict available channels in a sliding time window manner. Then, the authors propose a transductive TL process to transfer the knowledge (features) learned from different frequency bands to overcome the lack of spectrum data for learning. To further improve the effectiveness of the TL process, the authors utilize the transfer component analysis to choose suitable features for transferring. Experimental results show that the proposed TL approach can improve the accuracy of the LSTM model by up to 2.5 times, especially when the target task data are limited. However, when the domain data are sufficient, training the LSTM without TL could be better. Moreover, the TL is only effective when the knowledge is transferred between certain pairs of frequency bands, and thus similarity analysis of the frequency bands is needed.

For the second step, i.e., channel selection, RL is a very effective learning technique for the SUs to find the best channel under the dynamic and uncertainty of primary channels. However, RL techniques usually require a long learning time to find the optimal strategy. Therefore, several TL approaches have been recently proposed to speed up the learning process. Specifically, Q-learning approaches are developed in~\cite{zhao_spectrum_2020,cao_spectrum_2020,koushik_intelligent_2017} to find the optimal handoff strategies for SUs. To speed up the learning process of the newly joined SUs, both~\cite{zhao_spectrum_2020} and~\cite{cao_spectrum_2020} employ the \textit{offline Learning from Demonstration} TL strategy. Particularly, the authors propose to transfer the complete Q-table of the nearest SU to serve as an initial point for the new SU. Simulation results show that the TL approaches in~\cite{zhao_spectrum_2020} and~\cite{cao_spectrum_2020} can improve the convergence rate of the new SU's learning process by up to 30\% and 14\%, respectively. Unlike~\cite{zhao_spectrum_2020} and~\cite{cao_spectrum_2020},~\cite{koushik_intelligent_2017} proposes to transfer the knowledge of an expert SU selected based on its similarities with the new SU in terms of channel statistics, node statistics, and application statistics. Particularly, the level of similarity is determined using the Bregman divergence of the channel statistics, node statistics, and application statistics among the SUs. Then, the SU with the highest similarity is chosen as the expert SU for the TL process. This TL process aggregates the expert SU's policy $\pi_S$ and the new SU's policy $\pi_T$ with a transfer rate $\theta$, i.e., $\pi=(1-\theta)\pi_T+\theta\pi_S$. At the beginning, $\pi_S$ performs better since the time to learn the new policy is still short. As the learning time increases, the $\pi_T$ gradually becomes better. Thus, the authors propose to diminish the transfer rate as the learning proceeds. Simulations are carried out to evaluate the effects of different transfer rates on the Q-learning model. The results show that a high transfer rate ($\theta=0.8$) achieves better results at the beginning of the learning process than that of the lower transfer rates ($\theta=0.2$ and $\theta=0.5$). However, their performances become equal as the Q-learning model has more time to train. Another TL-based deep Q-network (DQN) is developed in~\cite{zhao_transfer_2015} for channel selection in CRNs. Particularly, a DQN is proposed to select channels with high QoS for file transmission. The authors propose to transfer the Q-values of the adjacent agents to the newly joined agent. A source agent selection policy is proposed, which selects the agents causing large interference. Then, at the TL process, the Q-values are reversed (i.e., positive rewards become negative) to encourage the new agent to avoid the channels that the source agents have chosen. Simulation results show that TL can significantly improve the throughput of the RL approach, especially in high traffic scenarios.

\subsubsection{Interference Management}
In CRN underlay, the SUs can share the same channel as the PUs, as long as the signals from the SUs do not cause unacceptable interference for the PUs. To control the interference, the SUs can control their transmission power~\cite{tsiropoulos_radio_2014}. However, in such constantly changing environments, finding an optimal power control policy is not an easy task for conventional ML techniques because of the unsteady number of users sharing the same resource (different domains) and the small time-window to react (short time to learn). To overcome these limitations, a TL approach is proposed in~\cite{galindo_distributed_2011}. Particularly, a distributed Q-learning approach is proposed for interference management in femtocells (SUs) that share the available radio resources with macrocells (PUs), aiming to find the optimal power levels that femtocells can assign to each resource block. Since the domains of the Q-learning models of the resource blocks are similar, the authors propose an unsupervised TL approach which transfers the full Q-table of the previous resource block to the target task resource block. Simulation results show that the TL process can reduce the probability that the macrocell capacity falls below a certain threshold by up to 20 times.

Different from~\cite{galindo_distributed_2011} that utilizes the knowledge from previous experiences, the TL approaches developed in~\cite{shah_deep_2018} and~\cite{shah_fast_2020} transfer the knowledge from nearby SUs. Specifically,  in~\cite{shah_deep_2018}, a DQN, which uses a neural network to estimate the Q-function, is developed for the SUs to determine their optimal power allocation. In CRNs, SUs that are close to each other may have similar environment parameters, and thus their Q-values are also similar. Thus, to speed up the learning process of the DQN, when a new SU joins the network, the Q-function and parameters of the nearest SU can be transferred to the new SU. Similarly, a TL-based DQN is proposed for optimizing spectrum utilization in~\cite{shah_fast_2020}, which transfers the Q-values from the nearest SU to the new SU. Simulation results show that TL can reduce the learning time by approximately 25\% and 80\% in~\cite{shah_deep_2018} and~\cite{shah_fast_2020}, respectively.
\subsubsection{Radio Map Construction}
\begin{figure}[!t]
	\centering
	\includegraphics[width=\columnwidth]{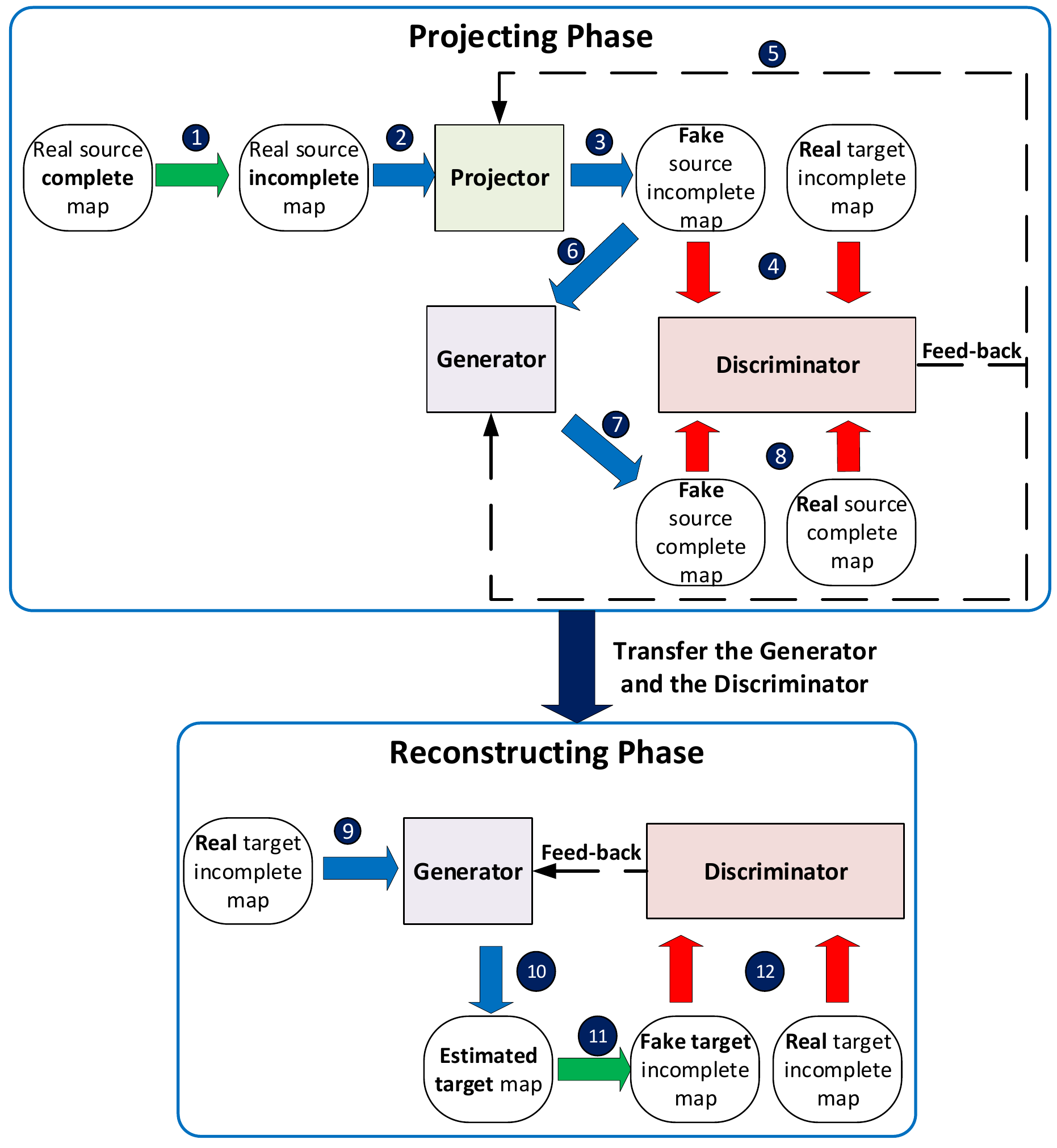}
	\caption{The GAN-based TL approach~\cite{zhao_using_2015}.}
	\vspace{-0.5em}
	\label{fig:sm3}
\end{figure}

Accurate radio map construction can play a crucial role in making optimal resource allocation decisions~\cite{phillips_survey_2012}. Practically, the radio maps are often constructed using measurements from a limited number of measuring devices. To construct the complete radio map from the limited measurements, ML techniques are effective solutions. However, conventional ML techniques are facing serious challenges such as the time-consuming training process and the scarcity of data. To address these problems, transductive TL approaches are proposed in~\cite{parera_transfer_2020b,han_two_2020}. Particularly,~\cite{parera_transfer_2020b} proposes a feed-forward NN to reconstruct the radio map corresponding to an antenna tilt configuration, i.e., antenna tilt degree. To address the lack of data in new tilt configurations, the authors propose a TL strategy to transfer knowledge from existing tilt configurations of the same antenna. Specifically, the authors propose to transfer a number of layers of the feed-forward NN and then retrain the model for the new target task. Experiments are conducted with different numbers of transferred layers to find the optimal TL strategy. The results show that transferring only the first two layers of the feed-forward NN yields the best result, and generally the proposed TL can achieve higher prediction accuracy (up to 12\%) compared to that of other approaches such as K-nearest Neighbors (KNN) and random forest. Different from~\cite{parera_transfer_2020b}, a TL approach is presented in~\cite{han_two_2020}, which transfers the knowledge gained from the other regions. In particular, a generative adversarial network (GAN) is developed to estimate the power spectrum maps for underlay CRNs. To address the lack of data, the authors propose to transfer the radio environment features gained from the other regions to the target region. Particularly, as illustrated in Fig.~\ref{fig:sm3}, the TL process consists of two phases. In the projecting phase, the complete maps from the source are first fed into a projector, i.e., a DNN, to create (adversarial) incomplete maps. Then, the discriminator will try to distinguish the created adversarial maps and the real maps from the target task. Once the training process is completed, the projector can create incomplete maps very similar to those of the real target task. Then, these fake incomplete maps are fed into a generator to create fake complete maps. The discriminator then is fed with the fake and real complete maps to train the generator. At the end of the projecting phase, the generator can create complete maps that are very similar to the real complete maps. At the beginning of the reconstructing phase, the real target incomplete map is fed into the generator to create the estimated full map. Then, these estimated full maps are transformed back into incomplete maps and fed into the discriminator to train the generator. Once the discriminator cannot distinguish between the fake and real incomplete maps, the training is completed. The generator is finally capable of creating estimated maps based on real incomplete maps. Simulation results show that the proposed TL-based approach provides a more accurate power spectrum map reconstruction performance than those of the traditional methods without TL. However, the effect of the TL process is not comprehensively evaluated.

%
\subsection{Resource Allocation}
In wireless networks, effective resource allocation is an important but also challenging task. Making optimal decisions regarding spectrum, resource block, and power allocation, load balancing, on-off switching, etc., can significantly improve the network's operation efficiency and user experience. For such optimization problems, ML has emerged to be an effective solution in recent years due to its ability to quickly approximate the optimal solution without requiring expert knowledge of the problem~\cite{bengio_machine_2020}. However, traditional ML models often do not perform well when data is lacking and require a long time to train. To address these problems, TL has been leveraged for many applications as discussed in the following.
\subsubsection{Channel Assignment}
\begin{figure}[!t]
	\centering
	\includegraphics[width=\columnwidth]{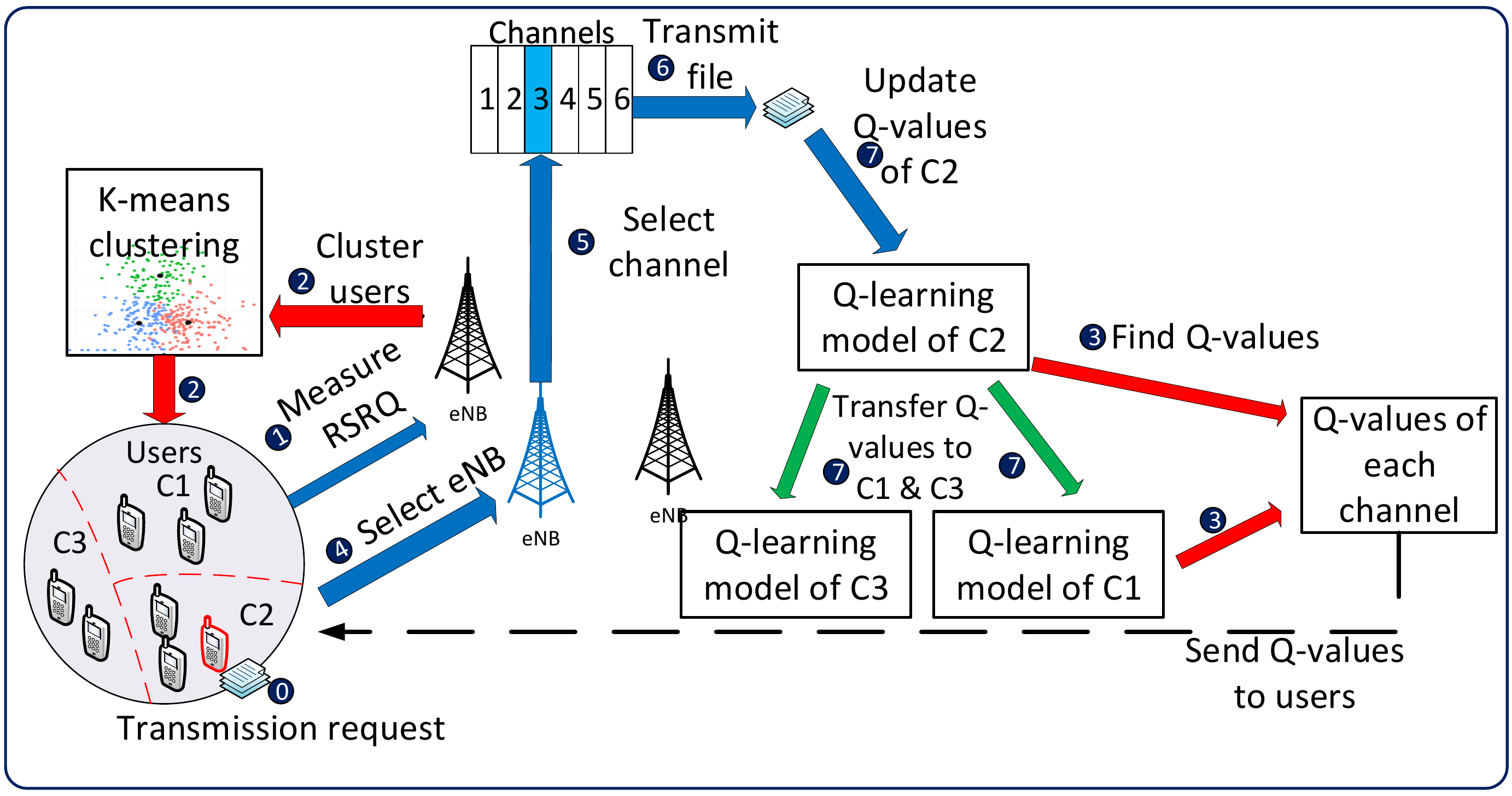}
	\caption{The TL-based enB and channel selection approach~\cite{zhao_using_2015}.}
	\vspace{-0.5em}
	\label{fig:sm2}
\end{figure}

Although RL approaches have been widely used due to their outstanding features in handling dynamic issues in resource allocation, there are still some limitations that can be addressed by TL. In~\cite{zhao_transfer_2013}, a Q-learning algorithm is developed to find the best channel for file transmissions in flexible opportunistic networks. In such flexible opportunistic networks, the network topology may change significantly over time, e.g., due to the aerial eNBs leaving the network. This requires a long time for the traditional Q-learning to learn the optimal channel selection for the new topology. Therefore, to speed up the learning process, a TL strategy is proposed to transfer the Q-values from previous experiences to the current model. Simulation results show that the TL process can improve both the convergence speed and the performance of the Q-learning model. Extended from~\cite{zhao_transfer_2013}, an unsupervised TL approach is proposed in~\cite{zhao_using_2015} to optimize the eNB and channel selection for file transmissions. As illustrated in Fig.~\ref{fig:sm2}, when there is a file transmission request, the users are clustered based on their Reference Signal Received Quality (RSRQ) measurements. Then, Q-learning is adopted by the eNBs to find the Q-values of each channel for each cluster. The obtained Q-values are then sent to the users for eNB selection. When an eNB is selected, it transmits the file using the channel with the highest Q-value. The result of the transmission, i.e., success or failure, is used to update the Q-values of that cluster. Finally, the Q-values are transferred to other clusters. Simulation results show that the proposed TL approach can reduce the delay and transmission failure rate up to 1.5 times compared to the approach without TL. 

\subsubsection{Power Allocation and Energy Efficiency}
TL-based approaches have also been developed for effectively learning the optimal control policy, e.g., transmit power allocation and on-off switching of wireless stations and devices, to improve the energy efficiency of wireless networks. Particularly, in~\cite{sharma_transfer_2016}, a TL-based RL approach is developed for managing access points in Wi-Fi networks, aiming to improve the energy efficiency by switching on/off the access points. Particularly, an actor-critic RL model is proposed to learn the optimal AP switching scheme based on Wi-Fi traffic data. To improve the learning speed of the RL model, the authors propose to transfer the knowledge gained from previous experience, i.e., previous policy, to the current model. Similar to~\cite{koushik_intelligent_2017}, the previous policy and the new policy are aggregated with a transfer rate, and the transfer rate is diminished as the learning proceeds. Experimental results show that the proposed TL process can improve the energy efficiency of the approach without TL by up to 22\%. In~\cite{li_tact_2014}, a TL-based RL approach is proposed for managing Base Stations (BSs) in radio access networks. Particularly, an actor-critic RL model is proposed to learn the optimal BS switching scheme. To further improve the efficiency of the RL model, the authors propose to transfer the knowledge gained from previous experience and neighboring regions. Similar to~\cite{sharma_transfer_2016}, the authors propose to aggregate the previous and new policies with a predefined transfer rate, and this transfer rate can be diminished over time. Simulation results show that the proposed TL process can improve the energy efficiency of the actor-critic model by up to 60\%, and a high transfer rate (0.5) achieves better results. In~\cite{sun_deep_2018}, a DRL approach is proposed for mode selection and resource management in Fog Radio Access Networks (F-RANs), with the main aim to reduce the network's power consumption. In the considered system, there is a network controller that will determine the user equipments' communication modes and processors' on-off states to minimize the system's long-term power consumption. To this end, the controller utilizes a DQN to learn and find the optimal control policy for the UEs. To speed up this learning process, the authors propose to use TL to transfer the knowledge of the previously trained DQN to the new DQN when the environment changes. Particularly, the authors employ \textit{Weight Initialization}, i.e., transfer the whole weights of the source model to serve as an initial point for the target model. Simulation results show that applying TL can help to speed up the learning process by 37.5\% when the environment changes moderately. However, in this work, TL is ineffective when the environment has major changes. In~\cite{dong_deep_2020}, a DL approach is proposed to find the optimal resource allocation policy to minimize the total power consumption of a BS in 5G networks. To this end, the authors develop a cascaded structure of NNs, where the tasks of optimizing bandwidth and transmit power allocation are distributed to the first and second NNs, respectively. Since the wireless environment changes frequently, a DTL approach is proposed. With this TL approach, the NNs are trained in offline mode, and then the first several layers of the two NNs are transferred to the NNs of the target task. Then, the NNs are trained online with the target data. Simulation results show that the TL approach can help the training process to converge much faster, i.e., by 2000-3000 epochs.

\subsubsection{Resource Block Allocation}
TL-based approaches can also be used to address resource block allocation problems in wireless networks. In~\cite{sahin_vrls_2019}, an actor-critic model is developed for assigning transmission blocks to vehicles in Vehicle-to-vehicle (V2V) communications. To this end, the authors propose to employ two DNNs as the actor and critic to train RL agents, each of which is responsible for assigning resource blocks to vehicles in its designated area. To speed up the learning process of the DNNs in new environments, the DNNs are trained in simplified environments and then transferred to more complex environments to fine-tune. Specifically, the authors employ \textit{Weight Initialization} and fine-tuning all the layers of the DNNs. Simulation results show that the proposed approach outperforms the other methods in terms of packet reception ratio. However, in this work, the impact of the TL process on the DNNs is not evaluated. Similarly, an echo-state-network-based approach is developed in~\cite{chen_echo_2017} for resource block allocation for wireless virtual reality networks. The authors first formulate the resource allocation problem as a non-cooperative game among the considered small BSs. Then, a RL model based on echo state network is proposed to find the Nash equilibrium of the game. In this game, once the environment changes, e.g., the number of small BSs increases, the RL model has to learn new utility values again, which is time-consuming. Thus, the authors propose a TL approach to transfer previous knowledge, i.e., previous outputs of the RL model, to reduce the time it takes to train the model on new utility values. Particularly, an echo state network is developed to find the relationship between the changes in the environments and the resulting changes in the utility functions. In other words, the effects of the environments on the utility function can be estimated, and thus the RL can learn the new utility values faster. Simulation results show that the TL approach can help to speed up the learning process by 18.2\%. 

\subsubsection{Resource Utilization Prediction}In addition to optimizing resource allocation, TL approaches can also be employed to predict resource utilization in wireless networks. In particular, the authors in~\cite{parera_transfer_2020a} propose an LSTM to sequentially predict a number of resource utilization parameters. A TL process is employed, which transfers all the layers of a pre-trained model to the target model. Then, the target model is trained with the target data to fine-tune the several last layers, while the first layers are fixed. This TL strategy is expected to be beneficial because the first layers capture more general features, and training only the last layers is more computationally efficient. Simulation results show that the TL process can reduce the training time of the target model approximately by half. 

\subsection{Channel Estimation and Prediction}
The physical properties of the wireless communication channels are always changing, which causes attenuation, distortion, delays, and phase shift of the signals. Therefore, it is essential to provide accurate and up-to-date predictions of the channel to overcome these limitations and facilitate signal processing. For such prediction tasks, conventional ML techniques have been widely applied. However, these ML techniques usually require a lot of data, have a low learning speed, and are not robust to environmental changes. To address these issues, various TL approaches have been developed. In~\cite{zeng_traffic_2020}, a DL approach is proposed for 5G/B5G cellular traffic prediction. Particularly, a spatial-temporal cross-domain neural network model is developed for traffic load prediction of various services such as SMS, call, and the Internet. The proposed model consists of multiple smaller models to simultaneously learn the features of different datasets including spatio-temporal (traffic level at different places over time), timestamp (weekdays or weekends), and cross-domain (BSs and their coordinates) datasets. The extracted features from these datasets are finally fed into a densely connected CNN for traffic level prediction. Since the traffic level of different services has a high correlation, e.g., all services' traffic levels are lower at midnight, a transductive TL process is proposed to transfer a model of one service to another service. Another transductive TL approach is proposed to predict traffic levels of different cellular services in~\cite{zhang_deep_2019}. Similar to~\cite{zeng_traffic_2020}, this approach also extracts features from different datasets and fuse them to feed into a densely connected CNN. The proposed TL process in~\cite{zhang_deep_2019} is also similar to that in~\cite{zeng_traffic_2020}, which transfers the model of one service to another. However, the main difference between the two approaches is that~\cite{zhang_deep_2019} does not use the timestamp dataset. Experimental results show that TL can help to improve the performance, in terms of Root Mean Square Error (RMSE) and Mean Absolute Error (MAE), of the proposed model in~\cite{zeng_traffic_2020} and~\cite{zhang_deep_2019} by up to 17.1\% and 12.9\%, respectively. 

Apart from service traffic level, ML techniques can also be leveraged to predict other parameters such as Channel State Information (CSI), channel quality, and Radio Frequency (RF) distributions. In such tasks, the knowledge from similar scenarios, e.g., frequency bands, can be leveraged to reduce the learning time and improve the prediction accuracy. For example, a transductive TL approach is proposed in~\cite{wagle_transfer_2012} to adaptively learn and predict dynamic RF environments. Particularly, a spatio-temporal Gaussian process model is developed to predict the distribution of RF variations. Since the RF environments are dynamic, the authors adopt TL to leverage the knowledge gained from different environments to reduce the training time for a new environment. Moreover, the authors investigate the trade-off between training time and prediction accuracy, the authors propose and compare three TL approaches with varying degrees of transferring, i.e., transferring all hyperparameters, transfering lengthscales and noise hyperparameters, transferring only the lengthscales parameters. Experimental results show that the more hyperparameters are transferred, the more training time can be reduced because the new model needs to learn fewer parameters. However, more transferred parameters may lead to a worse prediction accuracy because of the differences between the environments. Thus, more investigations on fine-turning methods are needed. 

The authors in~\cite{parera_transfer_2019} focus on evaluating different TL approaches for traffic prediction in wireless networks. Specifically, the authors develop and compare a CNN and an LSTM to predict channel quality indicator (CQI) in two scenarios, i.e., different frequency bands of the same 4G cell and different cells using the same frequency band. To address the scarcity of labeled data in the target domains, e.g., a particular frequency band or cell, the authors propose a transductive TL strategy to utilize knowledge from similar domains. Particularly, the authors propose to transfer the first layers of the source models, e.g., a cell or frequency band with sufficient data, to the target domain for fine-tuning. For both the CNN and the LSTM, all the layers except the last one are transferred to the source model, and the target models are then fine-tuned with target data. Simulation results show that the TL process can improve the prediction accuracy, and the CNN performs better than the LSTM if there is no target data, i.e., the CNN's RMSE value is 6\% lower than that of the LSTM. However, when there is more data available, the LSTM's RMSE value is 3\% lower than that of the CNN.  

\subsection{Other Issues in Spectrum Management}
In~\cite{yang_machine_2019}, a TL approach is proposed to predict path loss and delay spread for Unmanned Aerial Vehicle (UAV) communication systems. Specifically, the authors propose two algorithms, i.e., random forest and KNN, for path loss and delay spread prediction based on features such as the UAV's coordinates and the presence of buildings between two UAVs. To address the lack of data issue, especially in a new scenario or at a new frequency, the authors develop two TL strategies, namely frequency-based TL and scene-based TL. The first strategy transfers the models trained at different frequencies to the target frequency, whereas the scene-based TL transfers the models trained at different scenarios but at the same frequency. In both strategies, all the weights of the pre-trained models are transferred to the target models, and then the target models are fine-tuned with target task data, i.e., \textit{Weight Initialization}. Simulation results show that both the frequency-based TL and scene-based TL achieve lower RMSE compared to other approaches.

A TL approach is also used to address the lack of data for configuration troubleshooting problems in femtocell networks~\cite{wang_transfer_2011}. Particularly, an SVM is developed to classify the misconfiguration cases. Since the data in a femtocell network is often scarce due to the small number of users, the authors propose to transfer the output of the other femtocells' models to the target task. Specifically, the authors first determine the similarities among the femtocells by calculating the Kullback-Leibler divergence among their data. Then, each femtocell's model is used to classify the target task. The outputs of these models are then aggregated to be the the output of the target task. Particularly, the final output is the weighted sum of all the models' outputs, and the weights are the similarities among the femtocells. Simulation results show that the proposed TL approach can improve the SVM's prediction accuracy by up to 20\%.

\textit{Summary:} In this section, we discuss diverse applications of TL approaches in order to address existing limitations and challenges of conventional ML techniques for spectrum management problems such as the lack of labeled data, the heterogeneity of devices and environmental conditions, and the slow learning process. As summarized in Table~\ref{tab:Summary_spectrum}, it can be observed that knowledge from previous experiences and similar network conditions can be leveraged to address the lack of data. Furthermore, to improve the ML techniques' robustness to different environments, a baseline model trained with generalized datasets can be utilized. Alternatively, to improve the learning speed, especially in distributed learning systems, expert knowledge from similar learning agents can be used to initialize the new agent's model. Although the effectiveness of the TL approaches has been demonstrated for many spectrum management problems, there are still several challenges. For example, TL is only effective when the source and the target tasks share many common features. Thus, finding the right source for transfer necessitates proper analysis of the feature spaces of both tasks. However, most of the aforementioned works do not analyze the tasks' domains. Moreover, once the source is selected, there is still a need to determine which part of the source model to be transferred. For example, several approaches transfer only the first layers of the ML models because these layers capture more general features, whereas other TL strategies transfer the whole source model. 

\begin{table*}[]
	\caption{Summary of TL Approaches for Spectrum Management}
	\renewcommand{\arraystretch}{.5}
	\begin{tabular}{|>{\raggedright\arraybackslash}m{0.5cm}|>{\raggedright\arraybackslash}m{4cm}|>{\raggedright\arraybackslash}m{7cm}|>{\raggedright\arraybackslash}m{2.5cm}|>{\raggedright\arraybackslash}m{1.8cm}|}
		\hline 
		\multicolumn{1}{|>{\centering\arraybackslash}m{0.5cm}|}{\multirow{2}{*}{\textbf{No.}}} &
		\multicolumn{1}{>{\centering\arraybackslash}m{4cm}|}{\multirow{2}{*}{\textbf{Problem}}}&
		\multicolumn{1}{>{\centering\arraybackslash}m{7cm}|}{\multirow{2}{*}{\textbf{Knowledge to transfer}}} & \multicolumn{1}{>{\centering\arraybackslash}m{2.5cm}|}{\multirow{2}{*}{\textbf{ML method}}}& \multicolumn{1}{>{\centering\arraybackslash}m{1.8cm}|}{\multirow{2}{*}{\textbf{TL type}}}\\
		&&&&\\
		\hline 
		\hline 
		
		\cite{zhao_transfer_2013} & \multirow{29}{*}{Resource allocation} & Q-parameters from previous experiences & Q-learning  & Unsupervised\\ \cline{1-1} \cline{3-5}
		\cite{zhao_using_2015} &   & Q-values from other user clusters &  Q-learning  & Unsupervised   \\ \cline{1-1} \cline{3-5}

		\cite{sharma_transfer_2016} & & Knowledge from previous policies & Actor-critic model  & Unsupervised\\ \cline{1-1} \cline{3-5}
		\cite{li_tact_2014} &  & Knowledge from previous policies and neighbor regions & Actor-critic model  & Unsupervised\\ \cline{1-1} \cline{3-5}
		\cite{sun_deep_2018} &  & Q-values from other environments  & DQN  & Unsupervised\\ \cline{1-1} \cline{3-5}
		\cite{dong_deep_2020} &  & Transfer the NNs pre-trained with offline data to use online & Cascaded NNs  & Inductive\\ \cline{1-1} \cline{3-5}	
		
		\cite{sahin_vrls_2019} &  & Transfer a pre-trained model & Actor-critic model  & Unsupervised\\ \cline{1-1} \cline{3-5}
		\cite{chen_echo_2017} &  & Knowledge from previous experience & Echo state network  &  Unsupervised   \\ \cline{1-1} \cline{3-5}

		\cite{zappone_model_2019} &  & Transfer a model pre-trained with offline data to use online & ANN  & Inductive\\ \cline{1-1} \cline{3-5}

		\cite{parera_transfer_2020a} &  & Transfer a pre-trained model & LSTM  & Transductive\\ 
		\hline
		\hline
		
		\cite{peng_robust_2019} & \multirow{5}{*}{Cognitive Radio Network:} & Knowledge from different types of signals & CNN  & Transductive\\ \cline{1-1} \cline{3-5}
		\cite{zheng_spectrum_2020} &\multirow{5}{*}{Spectrum sensing}  & Knowledge from different types of signals and noise data & CNN  & Transductive\\ \cline{1-1} \cline{3-5}
		\cite{pati_deep_2020} &  & Knowledge from other environments & CNN, SVM  & Transductive\\ \hline
		
		\cite{lin_cross_2020} & \multirow{12}{*}{Cognitive Radio Network: } & Knowledge from other frequency bands & LSTM  & Transductive\\ \cline{1-1} \cline{3-5}
		\cite{zhao_spectrum_2020} & \multirow{12}{*}{Channel selection} & Q-table from similar SUs & Q-learning  & Unsupervised\\ \cline{1-1} \cline{3-5}
		\cite{cao_spectrum_2020} & & Q-function from nearest SUs & Q-learning  & Unsupervised\\ \cline{1-1} \cline{3-5}
		\cite{koushik_intelligent_2017} & & Knowledge from a similar SU & Actor-critic model  & Unsupervised\\ \cline{1-1} \cline{3-5}
		\cite{zhao_transfer_2015} &  & Knowledge from adjacent agents & DQN  & Unsupervised\\ \hline
		
		\cite{galindo_distributed_2011} & \multirow{5}{*}{Cognitive Radio Network}& Q-table from previous resource blocks & Q-learning  & Unsupervised\\ \cline{1-1} \cline{3-5}
		\cite{shah_deep_2018} & \multirow{5}{*}{Interference management} & Q-values of nearby SUs  & DQN &  Unsupervised    \\ \cline{1-1} \cline{3-5}
		\cite{shah_fast_2020} &  & Q-values of the nearest SU  & DQN  & Unsupervised\\ \hline
		
		\cite{parera_transfer_2020b} & \multirow{2}{*}{Cognitive Radio Network:}  & Knowledge from other tilt configuration & Feed-forward NN  & Transductive\\ \cline{1-1} \cline{3-5}
		\cite{han_two_2020} & \multirow{2}{*}{Radio map construction} & Knowledge from other regions & GAN  & Transductive\\

		\hline		
		\hline

		\cite{zeng_traffic_2020} & \multirow{11}{*}{Channel Estimation and Prediction} & Transfer a pre-trained model & LSTM, NN, CNN  & Transductive\\ \cline{1-1} \cline{3-5}
		\cite{zhang_deep_2019} &  & Transfer a model trained with a different dataset  & LSTM, CNN  & Transductive\\ \cline{1-1} \cline{3-5}
		\cite{wagle_transfer_2012} &  & Transfer parameters of a pre-trained model & LSTM, CNN  & Transductive\\ \cline{1-1} \cline{3-5}
		\cite{parera_transfer_2019} &  & Knowledge from different cells and different frequencies & CNN, LSTM  & Transductive\\ \hline
		\hline


		\cite{yang_machine_2019} & Predict path loss and delay spread for UAV communication system & Knowledge from different frequencies and scenarios & Random forest, KNN  & Transductive\\ \hline
		
		\cite{wang_transfer_2011} & Configuration troubleshooting in femtocells & Knowledge from different femtocells & SVM  & Transductive\\ \hline
		
		
	\end{tabular}
	\label{tab:Summary_spectrum}
	
\end{table*}

\section{Applications of Transfer Learning for Localization}\label{sec:localization}

Localization plays a very significant role for many applications in order to estimate the location/position of objects, e.g., mobile users, in future wireless networks including smart homes and smart cities. Typically, to implement the localization, wireless signals, e.g., received signal strength (RSS), CSI, and sensor signals, from wireless devices can be used to determine the position/location of the target objects in considered areas. Nonetheless, due to the time-variant signal characteristics of wireless environments, it is costly and impractical to collect and calibrate up-to-date signal data frequently. To address these issues, TL approaches can be applied with minimum signal data re-collection and re-calibration. In this section, we discuss applications of TL to address limitations of localization problems for both indoor and outdoor environments.


\subsection{Indoor Localization}

\begin{figure}[!t]
	\centering
	\includegraphics[scale=0.4]{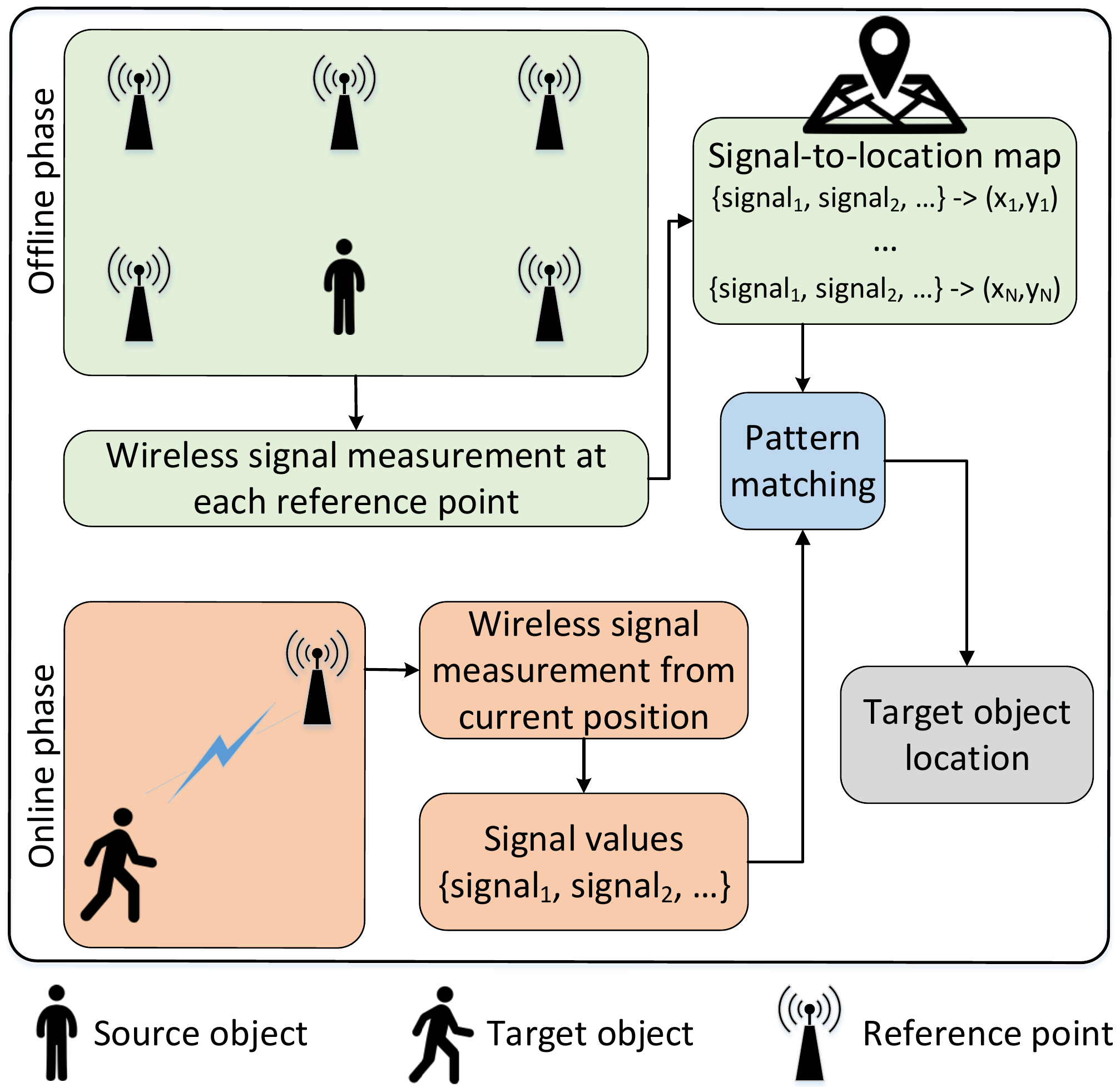}
	\caption{The general indoor localization framework.}
	\vspace{-0.5em}
	\label{fig:localization_4}
\end{figure}  

In indoor environments, the location of an object, e.g., a mobile device, can be estimated by using wireless signals from diverse reference points, e.g., Wi-Fi hotspots through considering offline and online phases. As illustrated in Fig.~\ref{fig:localization_4}, the reference points can first measure the set of wireless signals received from a source object. Then, a signal-to-location map, i.e., a correlation map to localize an object based on the captured signals, can be deployed offline. In the online phase, wireless signals from a target object in a specific position/location can be measured using a reference point. Upon obtaining the set of wireless signals from the target object, they can be stored for the signal pattern matching, i.e., mapping the most similar wireless signal patterns in the online phase with the ones in the offline signal-to-location map, to estimate the current position/location of the target object. However, the existence of dynamic signal behavior in indoor enviroments may trigger outdated signal-to-location map. For that, TL approaches can be used efficiently to address that problem for indoor localization without the need for frequent signal data updates. In the following, we discuss how wireless signals can be utilized to perform the TL process efficiently.

\subsubsection{RSS Map-Based Localization}

In a localization problem, the RSS map, i.e., a map which contains the correlation between RSSs and positions/locations of the object, can be generated based on the measurement of RSSs estimated by the reference points (when detecting objects within their considered areas). The RSS map can be easily produced without requiring additional hardware. Additionally, the use of RSS can provide more accurate location than that of the Global Positioning System (GPS) due to the robustness to the obstacles in a small-scale wireless coverage. As a result, RSS map has been widely used as a transferred knowledge in TL methods. 

In~\cite{pan_adaptive_2007}, the authors introduce an adaptive localization prediction leveraging inductive TL approach. In particular, a mapping function between RSS and physical location data in different periods is first derived based on a Manifold co-Regularization learning method~\cite{sindhwani_co_2005}. Then, using the TL method, the mapping function is adjusted for various discrete time dynamically by extracting the meaningful knowledge used to estimate location in another time with minimum re-calibration. Extended from the work in~\cite{pan_adaptive_2007}, the authors in~\cite{pan_transfer_2008} develop the Wi-Fi-based indoor localization problem (WILP) as an inductive TL problem considering the RSS map transfer in different scenarios, i.e., across time periods, locations, and devices. This aims to build an accurate localization model for different periods, areas, and devices with few labeled data separately. Specifically, if the user information, e.g., timestamp of the captured RSS, is available offline and online, a hidden Markov learning model is used to predict discrete location grids. Otherwise, the Manifold co-Regularization learning method can be applied to classify the location grids. With the real experiments, the proposed dynamic TL method can outperform the static non-TL method by 10\% in terms of the cumulative probability. 

Despite considering the indoor localization problem in different time period, location, and device scenarios, the TL approach in~\cite{pan_transfer_2008} is implemented in each scenario independently. To accomodate more complex conditions adaptively, the authors in~\cite{sun_adaptive_2008} study an inductive TL method considering both time and devices simultaneously due to variation of signal distributions. Particularly, a dimensionality reduction method, i.e., manifold alignment, is used to build a mapping function between different RSS and physical location spaces in a low-dimensional space. Then, the built RSS map is utilized to transfer knowledge from an old model to a new model. Through experiments, the joint consideration of time and devices for the TL approach can gradually improve the localization accuracy compared with those of non-TL methods. To investigate the impacts of environment changes on the fingerprinting values, the joint use of fuzzy clustering and inductive TL methods for indoor localization through updating the RSS map is studied in~\cite{zhang_tl-fcma_2018}. Particularly, the use of fuzzy clustering can reduce the effect of environment changes to the considered areas. With the obtained result from the fuzzy clustering, the RSS map can be updated using the manifold alignment-based TL approach to improve the accuracy of RSS map for the localization. The performance evaluation shows that the mean location error can be improved up to 71\% compared with that of the work in~\cite{sun_adaptive_2008}. 

Although the aforementioned works can reduce the training cost and complexity (through applying RSS map adaptation over various time and devices), they still need to extensively deploy a complete map through fingerprinting to directly localize mobile devices with high accuracy. Additionally, the map accuracy may be reduced when the map adaptation is repeated for several iterations, leading to higher localization errors. To address this problem, the authors in~\cite{sorour_joint_2014} design an inductive TL-based indoor localization scheme without creating a complete RSS map. Particularly, a source dataset with the same spatial correlation of the RSSs is first generated. Then, using the manifold alignment, the RSS map is transferred to a few fingerprints and several RSSs observations with unknown locations, aiming at executing the direct localization of the observed information. As such, the localization accuracy of the moving users can be enhanced through exploring the correlation of the observations. The experimental results then demonstrate that the proposed solution can reduce the mean localization error up to 20\% compared with that of the simulated radio map scenario.  

\begin{figure}[!t]
	\centering
	\includegraphics[scale=0.35]{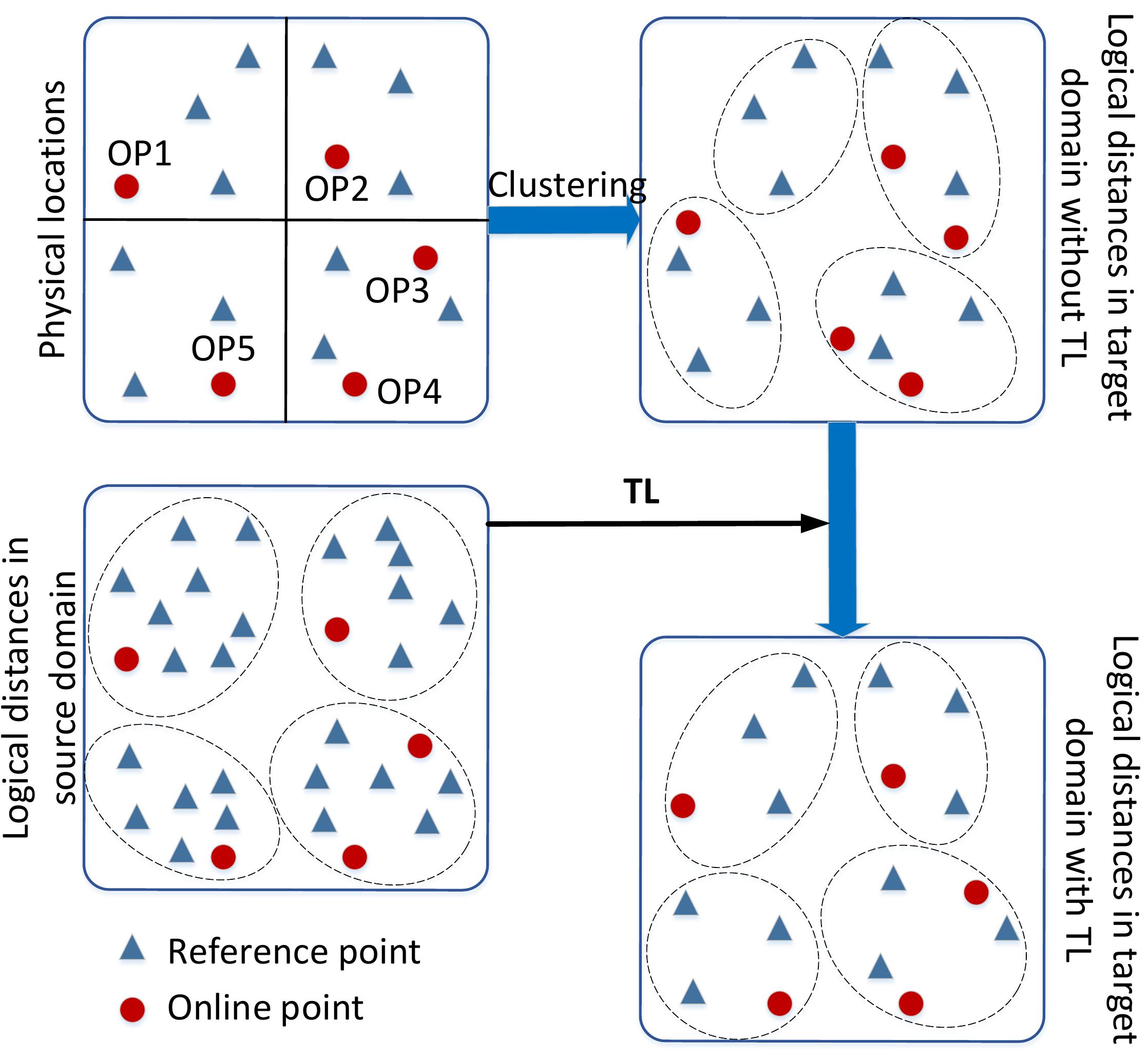}
	\caption{The TL-based framework for the localization scheme with clustering~\cite{liu_toward_2017}.}
	\vspace{-0.5em}
	\label{fig:localization_2}
\end{figure}

In practice, the indoor localization method using the fingerprint may incur a massive time overhead in the offline training step. For that, the authors in~\cite{liu_toward_2017} propose an inductive TL-based framework using KNN and weighted-KNN (WKNN) for indoor localization problem which can reduce the offline training overhead, thereby improving the system scalability. Specifically, the localization based on the fingerprint is first defined as a clustering problem where the online points' real locations are classified into specific areas according to the real-time RSSs and offline fingerprint dataset. In this case, the source domain can learn the distance metrics through optimizing the statistical dependence between the features of RSS and ground-truth labels. Afterwards, the logical distributions/distances between the online and reference points in the target domain can be calculated and reshaped according to the RSS map from the source domain as illustrated in Fig.~\ref{fig:localization_2}. As such, the points which are possessed by the same cluster will be closer to each other while the ones which belong to the different clusters will be farther from the others. Using an actual testbed environment, the proposed framework can achieve 85\% localization accuracy level with less offline training time overhead up to 30\% compared with that of the non-TL method.

\begin{figure}[!t]
	\centering
	\includegraphics[scale=0.206]{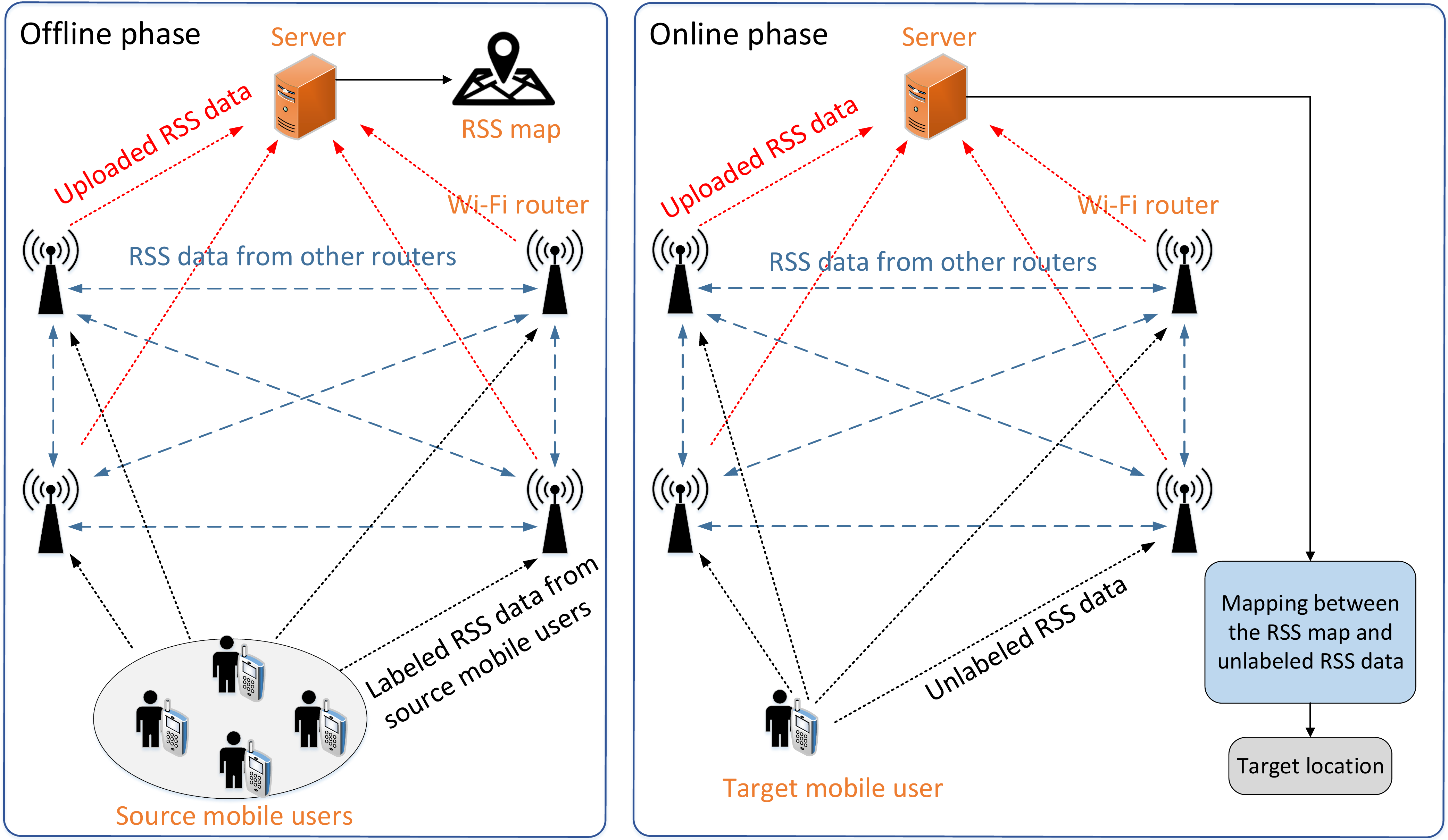}
	\caption{The TKL-based framework for the robust localization model using Wi-Fi routers~\cite{zou_adaptive_2017}.}
	\vspace{-0.5em}
	\label{fig:localization_1}
\end{figure} 

Another adaptive localization scheme for dynamic indoor environment leveraging a transfer kernel learning (TKL) with support vector regression (SVR) approach, i.e., TKL-WinSMS, is discussed in~\cite{zou_adaptive_2017}. In this case, the commercial-off-the-shelf (COTS) Wi-Fi routers can be used as the online reference points whose locations are extracted from the RSS observations. Leveraging online data together with labeled source data from offline calibrated RSS map, the TKL can be applied to build a robust localization model considering unlabeled RSS data from the target mobile user as shown in Fig.~\ref{fig:localization_1}. This model can produce a domain-invariant kernel matrix, i.e., the learned matrix built through extrapolating target eigensystem on source data, which can minimize the estimation error to the ground truth source data and has the most similar feature-space distribution between source and target domains. Given the kernel matrix as the input for the SVR training, the trained model from the online process can be used to adaptively improve the offline RSS map. The experiments reveal that the proposed TKL-based method can boost the localization accuracy up to 26\% compared with that of SVR without TKL method. Another adaptive indoor localization considering instance-based inductive TL approach is discussed in~\cite{hu_adaptive_2019}. When the environment changes, some old instances of RSS data may be obsolete, thereby degrading the localization accuracy. To address this issue, some parts of the old RSS training data which have the same distribution with the current RSS testing data can be selected through fine-tuning the weights of instances. Compared with the methods when the training as well as testing data have different distributions and when insufficient data are utilized, the proposed method can improve the localization accuracy up to 18\% and 10\%, respectively.

From all the above works, it can be observed that they only utilize linear/shallow representation features, and thus may lead to the poor performance when RSS variations due to the environment changes and heterogeneous hardware, e.g., device A is used in the offline phase while device B is utilized in the online phase, occur more frequently. To solve this problem, a DTL can be applied to extract more robust hidden representation features from the RSS data by learning transferable features deeply. In~\cite{wang_an_2019}, the authors propose a deep mean correlation alignment (DMCA) network to reduce the domain variance through aligning the hidden representation features between source and target domain in the specific DNN layers. Specifically, the mean embeddings and the second-order statics of the hidden representation features are aligned between two domains. With the DMCA, the effect of dynamic environment and various hardware can be alleviated without fingerprint and hardware calibrations. Compared with the non-DL-based learning, the DMCA can achieve a lower average error distance up to 2.76 meters in the experimental evaluation using smartphones on various days. 

For a certain real indoor localization scenario, the use of DTL may trigger a heavy burden in building the RSS fingerprint map due to the requirement of a great amount of samples. In practice, it is intractable to obtain sufficient samples for the real localization due to sensor fault or dynamic environment, and thus the DTL may not be applicable in this scenario. To this end, the authors in~\cite{yong_indoor_2020} introduce a hybrid domain-based transductive TL framework, i.e., HDTL, which can transfer important knowledge from multiple source domains to the single target domain. Particularly, the framework first distinguishes the source domains into homogeneous and heterogeneous features with respect to the target domain. Then, a mapping between the homogeneous and heterogeneous features in the source domains can be developed using homogeneous knowledge transfer between the source and target domains. To produce a common homogeneous feature space in both source and target domains, the mapping is applied to fill up some missing heterogeneous data in the target domain. Using the homogeneous feature space, the location of target can be determined accurately under the existence of data loss in the target domain. Compared with the conventional TL with stand-alone source domain, the use of TL with multiple source domains can provide higher localization accuracy, i.e., lower average error distance. 


\subsubsection{CSI Map-Based Localization}

Instead of using the RSS map for the indoor localization problem, the utilization of CSI map, i.e., a map which contains the correlation between CSIs and positions/locations of the object, can bring more stable signal amplitude, smaller multipath effect, and better representation of the location-dependent feature under the frequency diversity. For example, the authors in~\cite{gao_crisloc_2020} introduce an inductive TL-based localization scheme based on CSI fingerprinting, i.e., CRISLoc, using smartphones. Particularly, the CRISLoc first incorporates a passive mode of the smartphones to capture the over-the-air packets for obtaining CSI samples. Then, using the TL method, the high-dimensional CSI fingerprint samples can be re-modeled based on the obsolete fingerprint database and some up-to-date measurements. To track down the location of the smartphone, the KNN algorithm is also implemented. The experiments reveal that the proposed scheme can obtain small mean error within 0.29 meters with the variance up to 8.6 cm when the locations of two APs are changed, thereby ensuring the robustness of the CRISLoc towards dynamic environment. 

Another work considering the DTL for device-free passive wireless indoor localization (DFPWL) framework is investigated in~\cite{rao_device_2020} under time-varying CSI behaviors. In particular, the CSI is first extracted from a single link without deploying/wearing any devices or using any pre-processed CSI samples. Then, using the DTL method, the framework can learn new hidden representations from the CSI samples as fingerprints, aiming at minimizing the distribution discrepancy between the training and testing fingerprint samples. To predict the location of the target, the KNN algorithm is used to compare the training and testing fingerprint samples. Based on the experiments, the proposed framework can provide stable localization accuracy continuously, i.e., with  the mean localization error up to 1.5 meters, without re-collecting the CSI fingerprint samples. 

\subsubsection{Sensor-Based Localization}

In addition to the RSS and CSI, sensors embedded in smartphones or IoT devices, e.g., accelerator and camera sensors, also can be applied to localize target objects. For example, the inductive TL-based localization problem considering the knowledge transfer from monitoring outdoor movements to the indoor environment, i.e., iLoom, is introduced in~\cite{qiu_self_2017}. Specifically, accelerator sensors on the user's smartphone are first employed to obtain the indoor movement traces of the user leveraging a dead reckoning approach~\cite{steinhoff_dead_2010}, i.e., current position calculation of some moving objects based on the previous positions, speed estimation, and heading direction for particular periods. Nonetheless, due to high noise in the data sensing, the use of dead reckoning may suffer from the inaccurate position results. Thus, relatively accurate data obtained from outdoor movements using GPS are added. Then, the inductive TL process from the outdoor dataset to the indoor dataset is executed. Particularly, the TL process is motivated by the fact that for a specific person, his/her movement behaviors, e.g., walking style, do not change significantly whether he/she is in an indoor or outdoor area. As such, the movement knowledge using large training weights from the outdoor environment can be transferred to the indoor environment properly. To further improve the dead reckoning approach with lower error calibration, an acceleration range box (which is built by the TL approach) and some optimization methods are implemented. This acceleration range box works as an acceleration filter which can determine
an incorrect acceleration from a set of accelerations in various directions. Via both indoor and outdoor experiments, the proposed method can obtain higher localization accuracy up to 60\% with low-complexity training compared with that of the conventional method without acceleration range box.   

\begin{figure}[!t]
	\centering
	\includegraphics[scale=0.42]{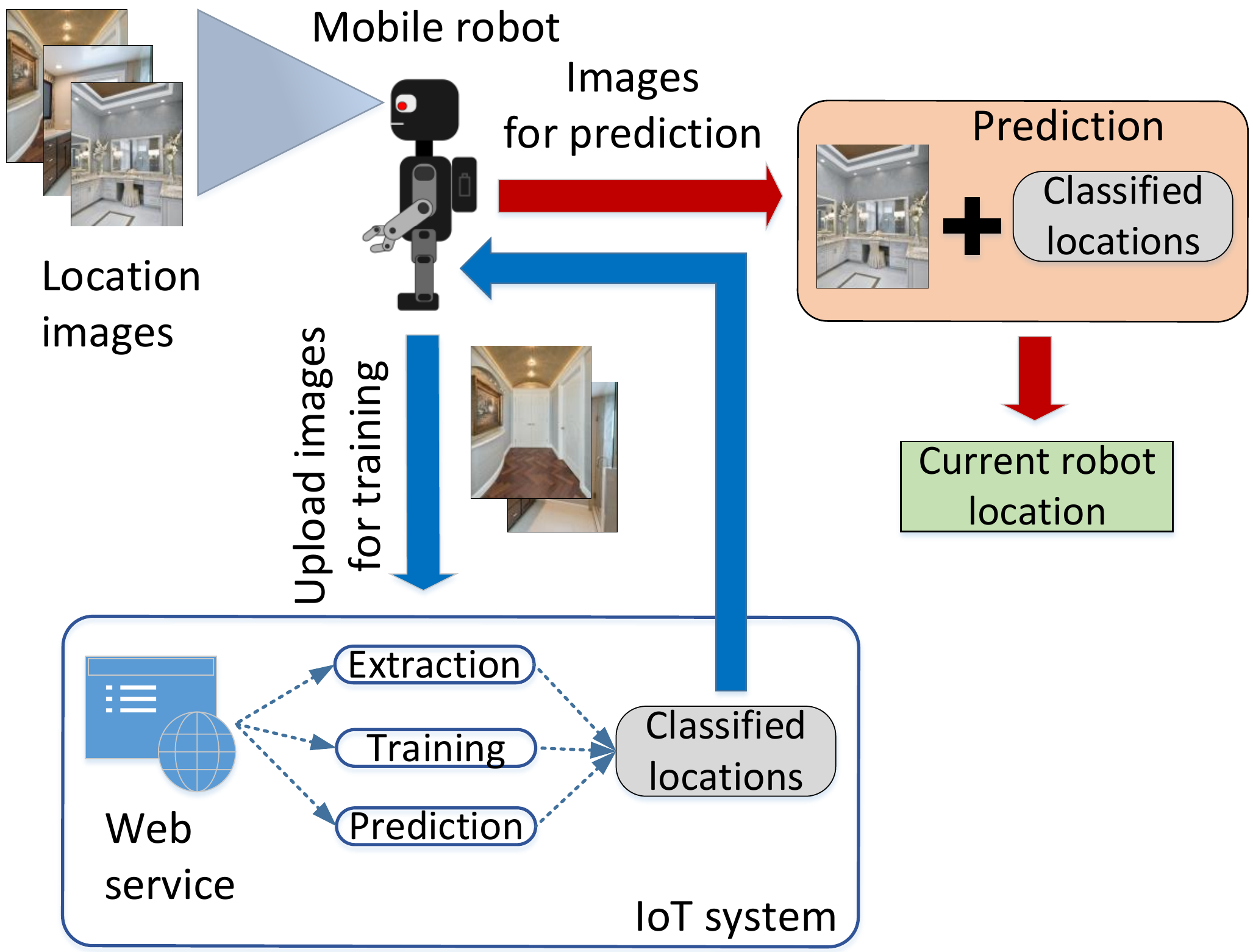}
	\caption{The TL framework for mobile robot localization~\cite{junior_a_2019}.}
	\vspace{-0.5em}
	\label{fig:localization_3}
\end{figure} 


The application of camera sensors to detect the location of an object has also been investigated recently. In~\cite{junior_a_2019}, the authors introduce an inductive DTL using CNN for indoor IoT-based mobile robot localization. As illustrated in Fig.~\ref{fig:localization_3}, the mobile robot can first capture images of its current locations using embedded GoPro and omnidirectional cameras. With the development of IoT system, the following processes are then conducted. First, the captured images are uploaded to the IoT system using Wi-Fi, and the useful features from them are extracted and trained using CNN. In this case, the last fully connected layer of the pre-trained CNN model can be removed and modified into a one-dimensional vector layer. The modified model can then be transferred to perform the TL process through a classification task using popular ML methods, e.g., Bayes, Multi-Layer Perceptron (MLP), SVM, random forest, and KNN. Based on this classification, the location of the mobile robot within the topological map coverage (based on the new real-time images of the mobile robot) can be predicted. Through experiments, the proposed method can achieve the localization accuracy up to 100\% especially when the images of an apartment are captured using the GoPro camera.

\subsection{Outdoor Localization}


From the previous section, it is observed that the indoor localization problems are only limited to small-scale wireless environments through using Wi-Fi or wireless local area networks (WLANs). To further increase the localization coverage, the localization can also be adopted in outdoor environments. Specifically, the location of various mobile nodes within a certain large-scale network coverage can be first trained in the offline phase through capturing the RSS. Upon building the RSS map in a similar way as that of the indoor localization, the locations of new mobile nodes can be estimated in the online phase. However, there exists an inherent challenge of using outdoor localization in which the mobile nodes' density and SNR change dramatically, leading to higher localization estimation errors. To cope with this problem, TL approaches can be used through transferring and fine-tuning the RSS map between source and target domain (similar to the indoor localization). In this case, instead of using small WLANs, the outdoor localization employs larger networks, e.g., low power wide area networks (LoRaWANs) to localize an object/a node within the same WAN area.

\begin{figure}[!t]
	\centering
	\includegraphics[scale=0.4]{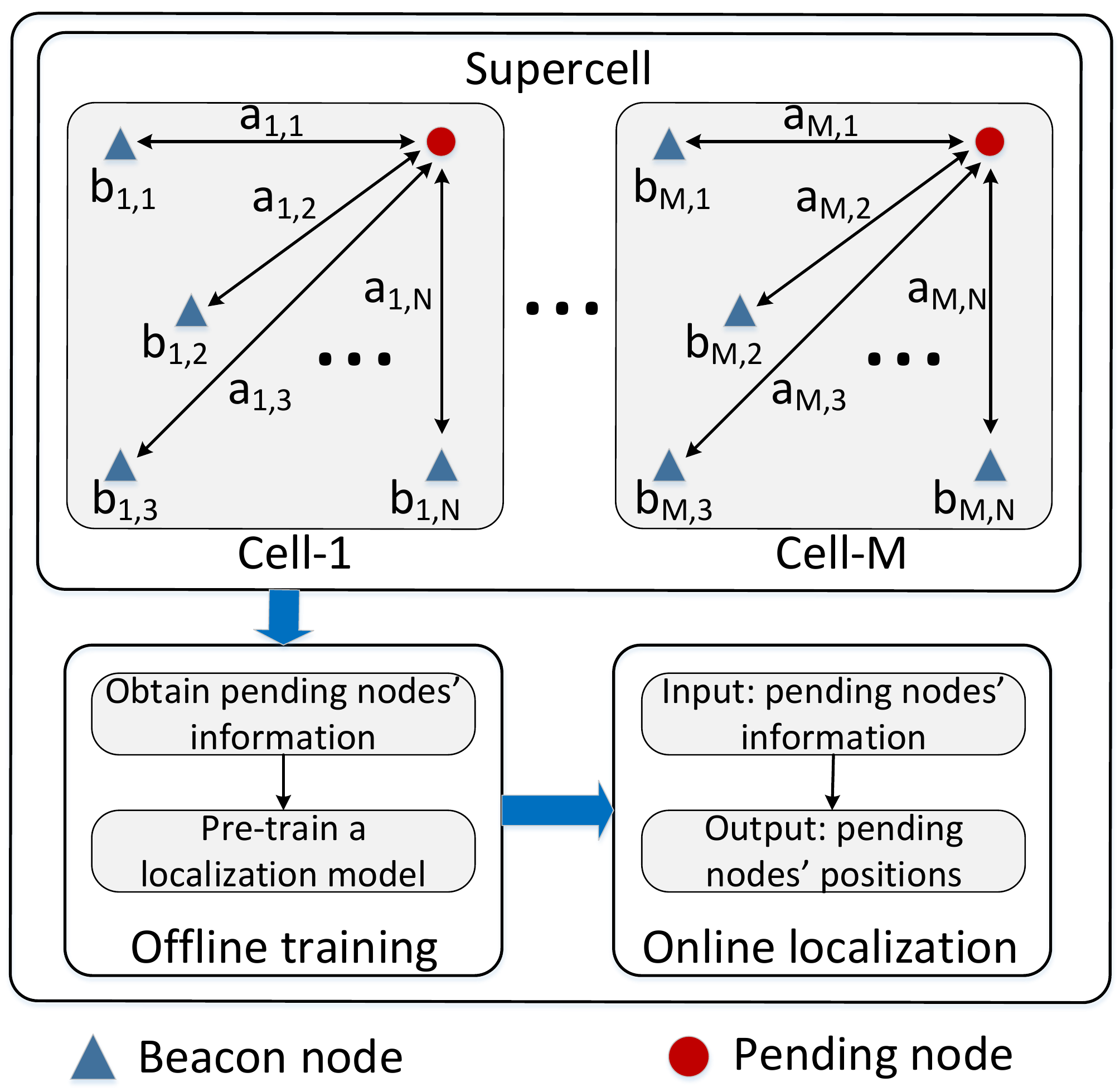}
	\caption{The CTTL framework for large-scale outdoor localization~\cite{tang_ctll_2015}.}
	\vspace{-0.5em}
	\label{fig:localization_6}
\end{figure}

In~\cite{tang_ctll_2015}, the authors propose a cell-based inductive TL (CTTL) method for a large-scale localization in a wireless sensor network. Specifically, as illustrated in Fig.~\ref{fig:localization_6}, the CTTL system incorporates beacon nodes with known location vector $\mathbf{b} = [b_{1,1},\ldots,b_{M,N}]$. When a pending node requests the information about its position, it can make a communication with the beacon nodes in the same cell, e.g., Cell-1 with the distance vector $\mathbf{a} = [a_{1,1},\ldots,a_{1,N}]$, aiming at reducing the communication cost. Using the RSS information from the pending nodes, the beacon nodes can estimate the pending nodes' locations. To address dynamic changes of pending nodes' density and SNR, a robust localization mechanism using a TL approach along with SVR model is applied to achieve accurate positioning. As shown in Fig.~\ref{fig:localization_6}, a supercell containing multi cells is first deployed to train many location samples within a big cell coverage in the offline phase. Then, in the online phase, the extracted location knowledge can be used by each smaller cell through modifying the trained model's weights to determine the pending nodes' positions within each cell. Using the CTTL, the positioning accuracy can reach 95\% with high robustness and low communication cost.

To boost the accuracy of the outdoor localization technique, the authors in~\cite{chen_outdoor_2019} investigate an inductive DTL approach using low power wireless area networks (LoRaWANs). Specifically, a grid segmentation method is first developed to obtain a sufficient number of virtual labeled locations based on the correlation between the labeled and unlabeled locations. In this case, some features including the RSS information, SNR, and current timestamps are used to pre-train a model. Leveraging the knowledge of labeled-unlabeled location correlation from the pre-trained model, the target model can be fine-tuned through adding new labeled locations, aiming at providing higher location prediction accuracy. Compared with SVM, the proposed DTL can improve the outdoor location accuracy up to 75\% after 5000 epochs in a large area. Motivated from the above work, the authors in~\cite{pimpinella_machine_2020} propose an inductive TL-based localization accuracy augmentation using RSS fingerprint samples in the co-located low power WAN technologies, i.e., LoRaWANs and wireless M-Bus. Particularly, by using an inter-technology method, RSS map calibration can be transferred from one technology to another one by using a pre-trained DNN model with two hidden layers. In this case, the weights of the first hidden layer remain unchanged while the ones of the second hidden layer are fine-tuned according to the RSS samples in the target domain. Through the real experiments, it is demonstrated that the proposed method can improve the localization accuracy up to 16\% compared with those of traditional ML methods without knowledge transfer.

%

\emph{Summary}: In this section, we discuss the use of TL for various localization applications. 
It can be observed that for the indoor localization problem, using the conventional inductive TL methods, e.g., SVM-based TL, KNN-based TL, and SVR-based TL, can improve the localization accuracy through transferring the knowledge map, i.e., RSS map, CSI map, sensing data knowledge, from one location/environment to another one. Additionally, the use of inductive DTL methods, e.g., CNN-based TL and DMCA-based TL, can further boost the accuracy with the minimum update of the map and more robust performance against the dynamic environment. Meanwhile, for the outdoor localization problem, the inductive DTL approach can be applied to enhance the positioning accuracy of mobile devices by fine-tuning the RSS map in large-scale wireless areas, e.g., cellular networks and LoRaWANs. The reason is that the semi-supervised TL method can perform well with the existence of a large amount of unlabeled datasets and few amount of labeled datasets. Nonetheless, most of the existing works in localization problems incorporate the inductive TL in which similar environments are used. To enhance the use of transductive and unsupervised TL approaches, the utilization of different environments, e.g., knowledge transfer from indoor environments to outdoor environments and vice versa can be further investigated. Additionally, the adoption of signal data from the embedded sensors to RSS/CSI map and vice versa may further prove the efficiency of TL approaches. The summary of all the TL approaches for localization applications is presented in Table~\ref{tab:Summary_localization}.

\begin{table*}[]
	\caption{Summary of TL Approach for Localization}
	\begin{tabular}{|>{\raggedright\arraybackslash}m{1.2cm}|>{\raggedright\arraybackslash}m{2.7cm}|>{\raggedright\arraybackslash}m{6.5cm}|>{\raggedright\arraybackslash}m{4.2cm}|>{\raggedright\arraybackslash}m{1.3cm}|}
		\hline 
		\multicolumn{1}{|>{\centering\arraybackslash}m{1.2cm}|}{\multirow{1}{*}{\textbf{No.}}} &
		\multicolumn{1}{>{\centering\arraybackslash}m{2.7cm}|}{\multirow{1}{*}{\textbf{Problem}}}&
		\multicolumn{1}{>{\centering\arraybackslash}m{6.5cm}|}{\multirow{1}{*}{\textbf{Knowledge to transfer}}} & \multicolumn{1}{>{\centering\arraybackslash}m{4.2cm}|}{\multirow{1}{*}{\textbf{ML method}}}&
		\multicolumn{1}{>{\centering\arraybackslash}m{1.3cm}|}{\multirow{1}{*}{\textbf{TL type}}}\\
		\hline 
		\hline 
		\cite{pan_adaptive_2007,pan_transfer_2008} & \multirow{12}{*}{RSS map-based indoor}  & RSS map & TL with manifold co-regularization & Inductive      \\ \cline{1-1}\cline{3-5}
		\cite{sun_adaptive_2008} & \multirow{12}{*}{localization} & RSS map & TL with manifold alignment & Inductive      \\ \cline{1-1}\cline{3-5}
		\cite{zhang_tl-fcma_2018} &  & RSS map & TL with fuzzy clustering and manifold alignment & Inductive      \\ \cline{1-1}\cline{3-5}
		\cite{sorour_joint_2014} &  & Incomplete RSS map & TL with manifold alignment & Inductive      \\ \cline{1-1}\cline{3-5}
		\cite{liu_toward_2017} &  & RSS map  & KNN and WKNN-based TL & Inductive      \\ \cline{1-1}\cline{3-5}
		\cite{zou_a_2016,zou_adaptive_2017} &  & RSS map & TKL with SVR & Transductive      \\ \cline{1-1}\cline{3-5}
		\cite{hu_adaptive_2019} &  & Signal strength distribution & Instance-based TL & Inductive      \\ \cline{1-1}\cline{3-5}
		\cite{wang_an_2019} &  & Knowledge from RSS fingerprint with location labels & DMCA-based TL & Inductive      \\ \cline{1-1}\cline{3-5}
		\cite{yong_indoor_2020} &  & Knowledge from homogeneous and heterogeneous features of multiple source domains & Hybrid domain-based TL & Transductive      \\ \hline
		\cite{gao_crisloc_2020} & \multirow{1}{*}{CSI map-based indoor} & Outdated CSI fingerprints & TL with KNN    & Inductive      \\ \cline{1-1}\cline{3-5}
		\cite{rao_device_2020} &\multirow{1}{*}{localization} & CSI map & DTL& Inductive    \\ \hline
		\cite{qiu_self_2017} & \multirow{1}{*}{Sensor-based indoor} & Knowledge learned from outdoor motion data & SVM-based TL      & Transductive      \\ \cline{1-1}\cline{3-5}
		\cite{junior_a_2019} &  \multirow{1}{*}{localization} & Knowledge learned from location images & CNN-based TL & Inductive      \\ \hline\hline
		
		\cite{tang_ctll_2015} & \multirow{4}{*}{Outdoor localization}  & Extracted location knowledge from a supercell training & SVR-based TL & Inductive      \\ \cline{1-1}\cline{3-5}
		\cite{chen_outdoor_2019} & & Knowledge learned from virtual labeled-unlabeled data based on RSSI, SNR, and timestamp & DTL & Inductive      \\ \cline{1-1}\cline{3-5}
		\cite{pimpinella_machine_2020} & & RSS map calibration from a LoRaWAN technology & DNN-based TL & Inductive      \\ \hline\hline
		
	\end{tabular}
	\label{tab:Summary_localization}
	
\end{table*}



\section{Applications of Transfer Learning for Signal Classification and Modulation Recognition}\label{sec:signal_regconition}
Radio signal classification and modulation recognition play a critical role in many wireless communication systems, especially in non-cooperative communications~\cite{wang_transfer_2020},~\cite{xu_deep_2020}. In particular, signal classification aims to detect information bits sent from the transmitter by analyzing received signals at the receiver. Differently, the goal of modulation recognition is to recognize the modulation technique of the transmitted signals from the transmitter and then derive the original signals by using the corresponding demodulation mechanism~\cite{wang_transfer_2020}. From the beginning of the wireless communication era, modulation recognition and signal classification have been carefully studied in many signal processing applications. However, traditional approaches require sophisticated mathematical models for specific signal types and properties which limit their applications. With the rapid development in neural network architectures, DL-based methods have been widely proposed to address the existing problems of traditional signal and modulation classification~\cite{bu_adversarial_2020}. However, DL-based detectors have several drawbacks. Specifically, DL solutions usually assume that data distribution is i.i.d distributed, which is infeasible in many cases. For example, the receiver may capture transmitted signals at different sampling rates due to the system properties and hardware configuration~\cite{bu_adversarial_2020}. As a result, the data distribution changes the result in a low detection performance. In addition, DL applications often requires long training time over large datasets to guarantee good detection performance. However, collecting and processing data are costly and sometimes impossible~\cite{selver_transferring_2019}. Recently, TL has been emerging as a promising solution to overcome these challenges. Therefore, in this section, we will discuss potential applications of TL for signal recognition in wireless networks.

\subsection{Signal Classification}

\begin{figure}[!]
	\centering
	\includegraphics[scale=0.27]{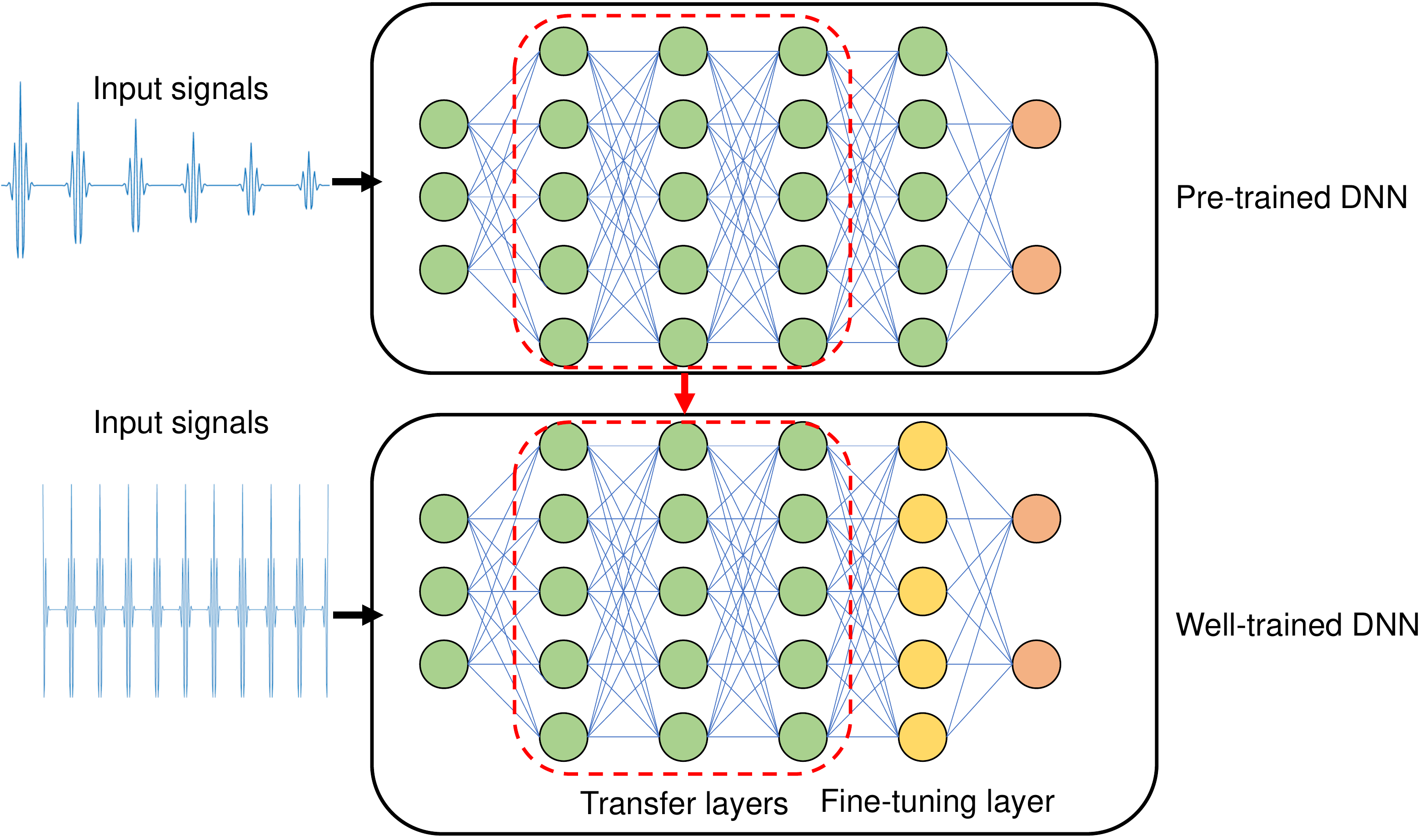}
	\caption{TL for signal classification~\cite{liu_deep_2020}.}
	\label{Fig.signal_classfication}
\end{figure}

The effectiveness of TL in signal classification has been demonstrated in several works in the literature. In~\cite{o_over_2018}, the authors propose a signal classification method considering the real propagation effects (i.e., over-the-air measurement). In particular, the authors point out that training a model from the beginning requires a very long time over large datasets to achieve a good detection performance. As such, the authors propose to use TL as a convenient alternative. Specifically, the model is first trained with common datasets, e.g., synthetic data. The trained mode is then transferred to classify over-the-air data with real propagation effects through fine-tune techniques. The simulation results demonstrate that by using TL, when training on the real data, the DL model can achieve the accuracy close to that of training on synthetic data. In~\cite{xu_deep_2020}, the authors study that the signal classification accuracy of a DNN is greatly relied on signal characteristics, surrounding environment, and signal conversion in time and frequency domains. Nevertheless, wireless channels may have different properties in in different settings and environments. Moreover, unexpected hardware errors may randomly occur, especially in low-cost devices, and thus a single analytical channel/hardware model cannot achieve good classification in all scenarios. For that, the authors in~\cite{xu_deep_2020} and~\cite{chen_deep_2019} propose to use TL to transfer knowledge trained in a source task to improve the classification accuracy of new tasks. In particular, a part of the source neural network is transferred to the target neural network to leverage the knowledge gained in the source task. Then, the remaining network layers will be fine-tuned to adapt with the target environment and setting.

Although these works can achieve good classification performance, they may not be effective in dealing with weak signals in sensor networks, especially low-cost and low-power wireless communication systems. For that, the authors in~\cite{liu_deep_2020} introduce a universal detection framework for backscatter communication networks based on DTL with offline training and online detection. In particular, a pre-trained DNN is first established to obtain the similar features between channel models by training offline over labeled data. After that, a part of the pre-trained model is frozen while the other part is fine-tuned to adapt the current channel coefficients by using TL as illustrated in Fig.~\ref{Fig.signal_classfication}. Finally, the well-trained model is used for signal classification. Simulation results then demonstrate that the proposed DTL-based detection framework can achieve the bit-error-rate close to that of the optimal detection with perfect channel state information.

\subsection{Modulation Recognition}

TL-based applications in modulation recognition have also attracted significant attention from researchers. In~\cite{xu_deep_2019}, the authors propose to use a CNN with TL for automatic modulation recognition. The authors demonstrate that the DNN can achieve a good detection performance when the SNR is high. However, when the wireless channel quality is unstable, the accuracy may drop below 40\%. To address this issue, the weights trained in the high SNR scenarios are transferred for training the model when the channel quality is not good, i.e., low SNR. Through experiments, the authors then demonstrate that TL can effectively address the unstable training problem of conventional DL approaches. Similarly, the authors in~\cite{bu_adversarial_2020} also adopt DTL to enhance the modulation classification performance of the target model. In particular, the authors demonstrate that with a simpler DNN architecture (compared to that of the source model), the target model can achieve a good detection performance by using TL. To do that, during the TL phase, the author freeze the source model and perform an adversarial training between the target model and a discriminator with the inputs are the output of the softmax layers in the source and target models. Based on experiment results, the authors then demonstrate that the proposed TL solution can achieve a higher detection performance compared to other existing methods.

In~\cite{wang_transfer_2020}, the authors propose a semi-supervised classification framework for zero-forcing aided multiple-input and multiple-output systems with the help of TL. In particular, the authors point out that conventional DL approaches require large datasets with labeled samples for effective recognition performance. Nevertheless, in practice, labeled data is not always available. For that, a TL model is proposed to enhance the recognition performance in real-life settings. Specifically, the proposed TL model has two components: (i) a traditional CNN to train a few labeled samples and (ii) a convolutional auto-encoder utilize massive unlabeled data. To improve the training accuracy and avoid overfitting, the encoder layer can share knowledge (e.g., weights) with the convolutional layers. Through experiments, the authors demonstrate that the proposed TL solution can outperform conventional approaches and achieve classification accuracy close to that of the case in which the channel quality is good.

The authors in \cite{ujan_efficient_2020} study a problem of recognizing radio frequency interference type and its characteristics, i.e., modulation class. They first compute scalograms, i.e., a visual representation of a waveform, for collected signal data, then feed them into a CNN-based classifier. To take advantage of pre-trained DL models, they use several famous models for image classification, including AlexNet, GoogleNet, VGG, and Resnet, as \textit{Feature Extractors} (section II.D). Specifically, classification layers of these pre-trained models, which are fully connected layers after the last convolution layer, are replaced by new classification layers. Simulation results then show that the accuracy of the system greatly relies on data type and feature extraction method.

\subsection{Signal Recognition for Radar Systems}

Most of the applications of TL for radio signal recognition are in the radar detection networks.
In \cite{yang_robust_2016}, the authors address the radar emitter signal recognition under a noisy environment recognition. 
The proposed system consists of four phases, including time-frequency analysis, feature extraction, feature reconstruction, and classification. 
To overcome the dynamic environment noise, the \textit{Feature Extractor} strategy in DTL is applied in the feature reconstruction phase. 
Specifically, a shared-hidden-layer autoencoder network is trained first, then the encoder layer of this network is used to reconstruct the features. 
In this way, the reconstructed features are robust again the dynamic noise of the environment. 
Then, these features are fed into a relevance vector machine (RVM)-based classifier. 
From the simulations, it is demonstrated that the proposed approach can achieve more robustness and accuracy than those of other methods using SVM \cite{zhang_new_2012}, K-means \cite{yang_hybrid_2013}, and fuzzy theory \cite{wang_fuzzy_2013}.

In \cite{xiao_radar_2019}, the authors study an application of DRL for recognizing radio waveforms. 
The signal data is first transformed into time-frequency images by using Choi-Williams distribution (CWD). 
Then features of these images are extracted by a feature fusion algorithm.
This algorithm aims to combine two type of features, including texture feature, which contains information about the spatial distribution of colors or intensities, and the depth characteristics, which is extracted by a CNN model. 
To cope with the lack of training data, they use \textit{Fine-tuning} strategy (section II.D) to transfer the knowledge of pre-trained models, including AlexNet and VGGNet, to their CNN model. 
The results show that this method achieves good accuracy in a noisy environment, $98.89\%$ and $99.72\%$ when using AlexNet and VGGNet, respectively.
Similarly, the authors in \cite{wang_transferred_2019} also focus on waveforms recognition for cognitive passive radar systems. 
However, in this work, they introduce a two-channel CNN with bidirectional LSTM (B-LSTM) architecture, named TCNN-BL.
In this architect, input data is first fed parallelly into two different CNN-based channels (T-CNN) to obtain multi-scale reception field.
Then, the output of T-CNN goes through a B-LSTM layer along with two classification layers. 
Since the sample rate of a cognitive passive radar highly depends on both operation mode, i.e., detection, parameter estimation, or imaging, and configuration, i.e., bistatic/multistatic and individual/cooperation, it is challenging to obtain adequate training data. 
To overcome this situation, they first train the model on a specific sample rate and then apply the \textit{Weight Initialization} technique to transfer the knowledge from this pre-trained model to the target model operating at a different sample rate.  
From simulation results, the proposed approach outperforms both conventional methods~\cite{lunden_automatic_2007,wang_automatic_2017}, in which features are extracted manually, and DNN-based methods \cite{oshea_introduction_2017,west_deep_2017, oshea_over_2018}.



\textit{Summary:} In this section, we have discussed the applications of TL for various signal classification and modulation recognition applications. In particular, the signal detection performance largely relies on the signal features, time frequency domain conversion, and wireless environment. However, in different settings and scenarios, wireless channels possess different characteristics. Moreover, signals may be transmitted at different sampling rates due to the system's properties and unexpected hardware errors and configurations. As such, conventional DL approaches may achieve a low detection accuracy as the assumption of independent and identical data distribution is violated. In addition, collecting sufficient labeled data for effective training is costly and even impossible in many scenarios. Finally, training the DL model from the beginning usually takes a long time and requires high-computing hardware. To tackle these problems, TL can be applied. The most common approach is transferring the trained source model to the target domain. One can reuse the whole trained model or a part of the trained model as the initiation of the target model. Moreover, the trained model can be frozen and then combined with new layers for training in the target model. It is worth noting that the existing works mostly assume that labeled data is available at both the target and the source domains. Nevertheless, in practice, labeled data may not be available at the target domains or even at the source domain. To address this problem, transductive TL and unsupervised TL can be adopted. The summary of the TL approaches for signal classification is highlighted in Table~\ref{tab:summary_signal_classification}.

\begin{table*}[!]
	\centering
	\caption{Summary of Transfer Learning Approaches for Signal Classification}
	\begin{tabular}{|>{\raggedright\arraybackslash}m{1.5cm}|>{\raggedright\arraybackslash}m{2.7cm}|>{\raggedright\arraybackslash}m{5cm}|>{\raggedright\arraybackslash}m{3cm}|>{\raggedright\arraybackslash}m{2cm}|}
		\hline 
		\multicolumn{1}{|>{\centering\arraybackslash}m{1.8cm}|}{\multirow{1}{*}{\textbf{Paper number}}} &
		\multicolumn{1}{>{\centering\arraybackslash}m{2.7cm}|}{\multirow{1}{*}{\textbf{Problem}}}&
		\multicolumn{1}{>{\centering\arraybackslash}m{5cm}|}{\multirow{1}{*}{\textbf{Knowledge to transfer}}} &
		\multicolumn{1}{>{\centering\arraybackslash}m{3cm}|}{\multirow{1}{*}{\textbf{ML method}}} & \multicolumn{1}{>{\centering\arraybackslash}m{2cm}|}{\multirow{1}{*}{\textbf{TL type}}}\\
		\hline 
		\hline 
		\cite{o_over_2018} & \multirow{5}{*}{Signal classification}  & A part of the pre-trained model & CNN  &  Inductive  \\ \cline{1-1}\cline{3-5}
		\cite{xu_deep_2020} &  & A part of the pre-trained model & CNN &   Inductive  \\ \cline{1-1}\cline{3-5}
		\cite{chen_deep_2019} &  & The pre-trained model & CNN &  Inductive   \\ \cline{1-1}\cline{3-5}
		\cite{liu_deep_2020} &  & A part of the pre-trained model & CNN &Inductive \\ \hline
		
		\cite{ujan_efficient_2020} &  & A part of the pre-trained model & CNN &   Transductive  \\ \cline{1-1}\cline{3-5}
		
		\cite{xu_deep_2019} & \multirow{3}{*}{Modulation Recognition} & Weights of the pre-trained model & CNN& Inductive     \\ \cline{1-1}\cline{3-5}
		\cite{bu_adversarial_2020} &   & A part of the pre-trained model & CNN &   Inductive   \\ \cline{1-1}\cline{3-5}
		\cite{wang_transfer_2020} &   & A part of the pre-trained model & CNN & Inductive      \\ \cline{1-1}\cline{3-5}
		\cite{yang_robust_2016}	&   & A part of the pre-trained model & CNN & Inductive     \\ \cline{1-1}\cline{3-5}
		\cite{xiao_radar_2019}&   & A part of the pre-trained model & CNN & Transductive		\\ \cline{1-1}\cline{3-5}
		\cite{wang_transferred_2019} &   & Parameters of the pre-trained model & CNN and B-LSTM & Inductive
		\\ \hline
	\end{tabular}
	\label{tab:summary_signal_classification}
	
\end{table*}


\section{Applications of Transfer Learning for Wireless Security}\label{sec:security}
In the last few years, the rapid development of wireless technologies such as 5G and IoT on the one hand has brought great benefits for human beings, but on the other hand has led to many serious concerns regarding emerging security issues for users as well as service providers~\cite{cao_survey_2019}. Although conventional ML techniques have been widely employed for wireless security, they have been facing some major challenges. Specifically, the great diversity of devices and environmental conditions have hindered ML's applicability because conventional ML techniques require a large amount of data and training resources to be effective for only a specific problem. To address these limitations, TL has emerged to be an effective solution which can utilize knowledge not only from past experiences but also from other domains, i.e., different types of devices and environments. Therefore, in this section, we discuss applications of TL in security issues in wireless networks, such as intrusion detection, jamming attacks and eavesdropping.

\subsection{Intrusion Detection}
Wireless intrusion detection techniques can be classified into signature-based and anomaly-based techniques~\cite{buczak_a_2015}. Signature-based intrusion detection relies on the known characteristics (signatures) of the attacks to detect intrusion. Although these methods can effectively detect intrusions and generate few false alarms, they require manual updates of the signatures database, and they cannot detect (zero-day) novel attacks~\cite{buczak_a_2015,vu_deep_2020}. On the other hand, anomaly-based intrusion detection methods first try to model normal network behaviors and then detect intrusion by identifying abnormal behaviors. This type of method can detect zero-day attacks, and the data collected from these zero-day attack cases can be useful to build the signature database for new attacks. Nevertheless, this type of intrusion detection techniques is prone to creating many false alarms because new legitimate behaviors might also raise the alarm~\cite{buczak_a_2015}. 

\begin{figure}[!t]
	\centering
	\includegraphics[scale=0.4]{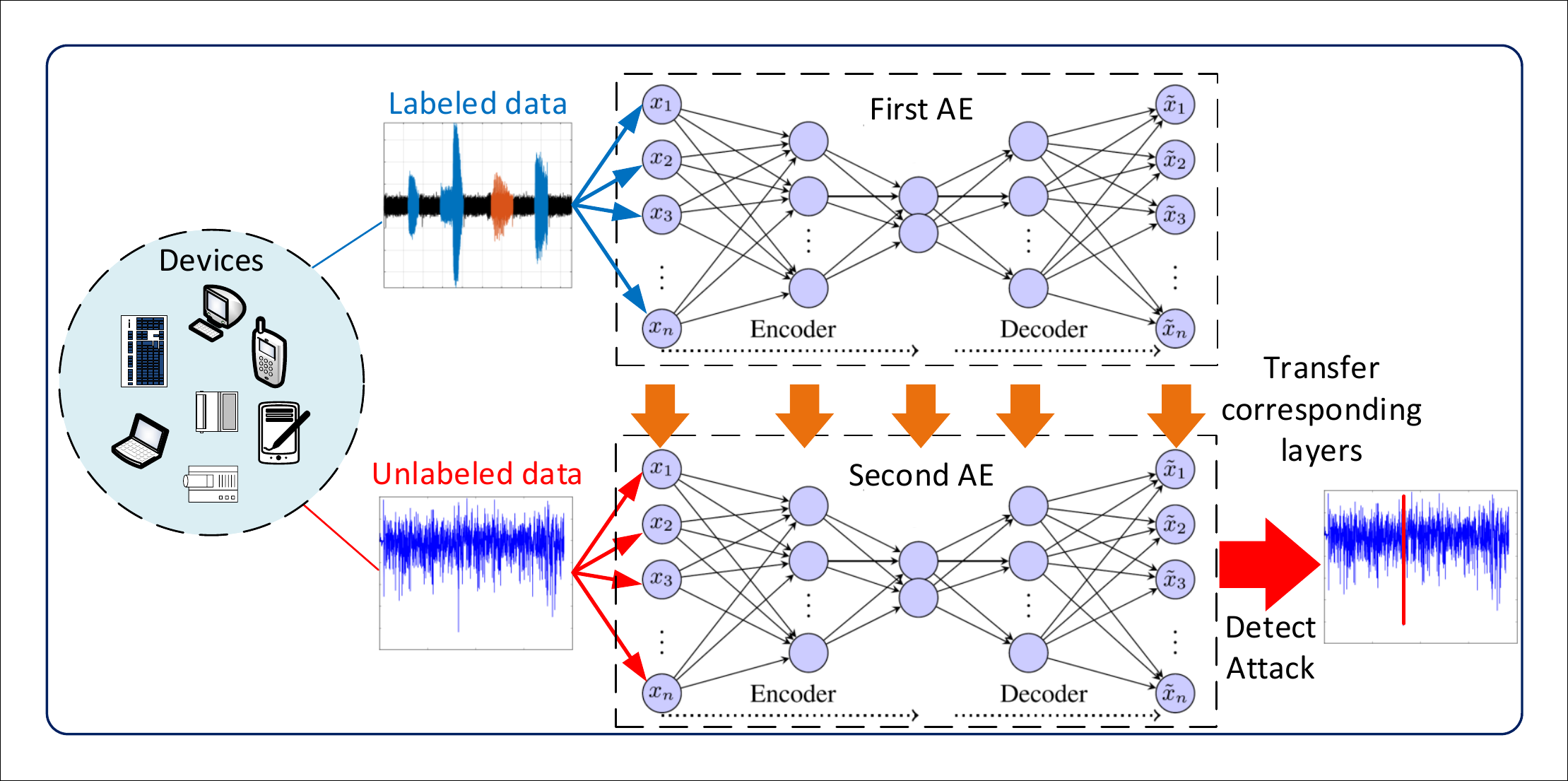}
	\caption{The TL-based intrusion detection approach~\cite{vu_deep_2020}.}
	\vspace{-0.5em}
	\label{fig:sec_1}
\end{figure}

With the ability to classify behaviors automatically, ML techniques have been considered one of the most effective techniques for anomaly-based intrusion detection~\cite{buczak_a_2015,vu_deep_2020}. However, ML techniques can only perform well if the training data's and the actual data's distribution are similar, which is highly unlikely in the cases of zero-day attacks~\cite{vu_deep_2020}. To mitigate this problem, the ML models must be trained with a huge amount of labeled data. However, this is often intractable to achieve in practice and consequently hinders the applicability of ML in intrusion detection~\cite{vu_deep_2020}. To address this problem, a DTL approach is proposed in~\cite{vu_deep_2020} for IoT intrusion detection, which can utilize both labeled and unlabeled data to detect attacks such as DDoS, UDP flooding, TCP flooding, and sending spam data. Specifically, the proposed model consists of two AutoEncoders (AEs) as illustrated in Fig.~\ref{fig:sec_1}. The first AE conducts supervised learning of the source data, whereas the second AE is trained with the unlabeled target data. The knowledge gained from the source domain will be transferred to the target domain (transductive TL) by minimizing the Maximum Mean Discrepancy (MMD) distances between each layer of the first AE and its corresponding layer of the second AE. Another DTL approach is proposed in~\cite{wen_time_2019}, which also analyzes network traffic to detect anomalies. Particularly, a CNN model is developed to detect abnormal segments of the time series, which is similar to detecting desired objects in images. Moreover, the authors propose a TL strategy that transfers the first 12 layers of a pre-trained CNN to the target model and fine-tune the target model with target data. This TL strategy is expected to be more efficient than transferring the whole model because the first layers usually capture the common features between the tasks, whereas the remaining layers capture the more task-specific features. Experiments results show that the TL strategy can improve the intersection over union (IOU) score of the CNN by up to 21\%.

Typically, anomalies in traffic data are the indicators of the attacks that create an abnormal amount of traffic such as DDoS. However, there is another type of attack, namely emulation attack, that cannot be detected using traffic data. For example, in Primary User Emulation (PUE) attack, an attacker pretends to be a PU to prevent the other SUs from competing for a spectrum band, thereby maximizing the attacker's spectrum usage. For such kind of attack, ML techniques can be leveraged to classify the signals coming from the real device and the attacker's device. In~\cite{sharaf_on_2016} and~\cite{dabbagh_authentication_2019}, the authors develop approaches to identify emulation attacks based on device fingerprint (i.e., unique features of each device) and environmental effects (i.e., the effects of the physical environment on the device's fingerprint). Specifically, the proposed approaches estimate and model the environmental effects on each device's fingerprint and detect abnormal differences between the expected and the measured environmental effects. Since the network devices often have similar environmental effects when they are physically close together, both~\cite{sharaf_on_2016} and~\cite{dabbagh_authentication_2019} propose transductive TL approaches to transfer the knowledge of environmental effects among nearby devices. Particularly, the authors in~\cite{sharaf_on_2016} propose to identify the changes in the environments of nearby devices and use this knowledge to filter out the effects of the environments on the devices' fingerprints. Simulation results show that the implemented TL scheme helps to filter out the effects of environmental changes, which leads to a lower false-positive rate compared to that of the approach without TL. Alternatively, the author in~\cite{dabbagh_authentication_2019} propose a TL strategy that transfers the knowledge learned from a device (e.g., smartphone) to a nearby device of a different type (e.g., sensor). The reason is that if the two devices are physically close to each other, they might share similar environmental effects. Experimental results show that the applied TL approach can improve the detection performance by 70\%, and higher transfer weights (i.e., the percent of the source model's weight transferred to the target model) can lead to better results.

Similarly, the authors in~\cite{sharaf_transfer_2015,zhao_the_2018} also propose approaches to identify IoT devices based on fingerprints. However, these approaches do not consider environmental effects. Instead, they utilize the knowledge gained from past experiences. Particularly, a Bayesian model is developed to classify device fingerprints to detect attacks in~\cite{sharaf_transfer_2015}. The authors first build an abstract knowledge database which is composed of fingerprints of all devices in each time frame. At a new time frame, new fingerprints are received. Then, an inductive TL strategy is applied to combine the new fingerprints with the old fingerprints stored in the abstract knowledge database to cluster and assign the new fingerprints to the devices. In~\cite{zhao_the_2018}, the authors develop an inductive TL technique that combines a large amount of source data, i.e., data collected in the past, and the smaller target dataset, i.e., current data, to train an SVM model for the classification of IoT devices. The main difference between the two approaches is that the changes over a long time (i.e., one year) in the physical features are taken into account in~\cite{zhao_the_2018}, whereas the time frames in~\cite{sharaf_transfer_2015} are much shorter. Simulation results show that the TL approaches proposed in~\cite{sharaf_transfer_2015} and~\cite{zhao_the_2018} can improve the recognition rate by 3.5\% and 9\%, respectively. From these results, we can observe that using the knowledge from a longer time frame might be more beneficial. However, this TL strategy is also much more time-consuming.

Aside from fingerprints, wireless devices can also be recognized using transient signals (which include the amplitude, phase, and frequency) due to the lower hardware requirements than that of the steady-state signals. Nonetheless, the wireless device recognition may suffer from limited number of transient signal data as the wireless devices to be recognized are often non-cooperative. To this end, the inductive TL approach for individual recognition of wireless devices with small amount of samples is discussed in~\cite{yu_individual_2019}. In particular, a wireless device as the reference AP first captures transient signals from other wireless devices and then classifies signals in the source and target domains. Afterwards, the feature weights of transient signals in the source domain are extracted by using the transient envelope feature extraction with entropy weighting method, and transferred to the target domain. For that, a new KNN-based model can be trained using the transferred knowledge by modifying the feature weights of the source model samples. Through the experiments using 5 wireless devices and an oscilloscope to obtain 200-360 samples and 40 samples of source and target domain, respectively, the proposed TL method can improve the classification accuracy up to 13.2\% compared with that of typical KNN without TL method. Additionally, it is possible to recognize similar RF devices based on their emitted signals, even when these devices are in the same batch built by the same manufacturer. The differences in emitted signals come from device noises and manufacturing errors. Therefore, the unique identification of an RF device can be derived from the statistical characteristics of its emitted signals. In \cite{wang_radio_2020}, the authors introduce a TL-based model to identify an RF device. In particular, they employ the \textit{Weight Initialization} technique to a LSTM model, namely TL-LSTM. They then use eight RF devices, in which four devices generate the source data and the rest of devices generate the target data. The source model is first trained in the source domain. Then, the first layer's parameters of the target model are cloned from that of the trained source model. The results demonstrate that the proposed TL can improve the accuracy in identifying RF devices regardless the noise of the environment.
\subsection{Jamming Attacks}
Jamming attack is a common type of security issues in wireless networks, mainly because attackers can easily launch jamming attacks without requiring complete information regarding the target system. To mitigate jamming attacks, conventional methods such as changing the spectrum and transmission power have been widely applied~\cite{vadlamani_jamming_2016}. However, with the development of ML techniques, attackers can launch adaptive jamming attacks to counter these fixed anti-jamming methods. To address these adaptive attacks, reinforcement learning algorithms have been considered to be a very potential solution to dynamically learn and find the optimal anti-jamming control policies. Nevertheless, reinforcement learning algorithms usually require a long learning time to be effective. To address this issue, TL can be used to speed up the learning process of traditional reinforcement learning. In~\cite{chen_reinforcement_2018}, a TL-based approach is proposed for anti-jamming power control of wireless body area networks (WBANs). Specifically, in the considered WBAN, there is a WBAN coordinator which determines the power level and time slot of the transmission of sensing data while taking into account jamming attacks. Since this anti-jamming power control process can be formulated as a Markov decision process, an advanced Q-learning approach is developed to find the optimal power control policy. This Q-learning approach utilizes unsupervised TL to determine the initial Q-learning parameters based on previous experiences, thereby avoiding unnecessary exploration at the beginning and speeding up the learning process. Simulations results show that the proposed approach always outperforms the Q-learning approach without TL in terms of bit error rate (BER), and SINR, and the long-term utility of the jammer can be significantly reduced.

Another approach that combines Q-learning and TL is proposed in~\cite{han_multi_2019} for anti-jamming multi-regional communication. Specifically, the approach first applies Q-learning to learn the jamming rules of the attackers and then determine anti-jamming policies accordingly. However, this method is only effective for a particular wireless environment (region). When the wireless environment changes (in a new region), the whole Q-learning process has to start again. Thus, the authors propose to implement unsupervised TL to transfer the previous knowledge, i.e., Q table, gained from other regions to initialize the learning process of a new region. To evaluate anti-jamming performance, simulations are carried out to compare the proposed approach with an independent learning algorithm (without TL). The results show that the proposed approach can converge to the optimal policy faster and achieve better BER compared to those of the independent learning algorithm.

In~\cite{lu_uav_2020}, a DRL approach is developed for UAV aided anti-jamming cellular communications. Specifically, UAVs are deployed to dynamically relay messages of mobile users in the case where the BS is severely jammed. To choose the optimal relay policy, the UAVs employ a CNN that takes into account current BER, channel gains, and jamming power estimation to determine the Q-values for different relay policies. Knowledge from previous experiences is used to initialize the weights and learning parameters of the new CNN, i.e., inductive TL. Simulation results show that the proposed TL strategy can not only reduce the learning speed of the Q-learning but also improve the performance of the Q-learning approach in terms of BER and energy consumption.

\subsection{Other Security Issues}
Apart from intrusion detection and anti-jamming, TL approaches have also been developed for other security issues such as eavesdropping and spam voice calls. In~\cite{do_learning_2018}, a framework for energy-efficient data communications in energy-harvesting CRNs is proposed, which takes into account hidden eavesdroppers. Specifically, a model-free RL, i.e., the actor-critic model, is developed to find the optimal data encryption policy for transmission. TL is also implemented in this RL process, where the action is determined not only by the reward of the last action but also from the rewards of previous actions. This unsupervised TL of the previous experience is expected to increase the convergence speed of the RL process. Simulations demonstrate that the proposed approach can achieve better channel utilization and transmit more data packets compared to other conventional approaches. In~\cite{bhowmik_mtrust_2020}, an unsupervised learning approach is developed to filter spam voice calls by assigning trust scores among users. In particular, the approach extracts data from call logs to determine trust scores using an MLP model. Since call data are user-specific, a model that is accurate for one user might not be accurate for another user. Thus, a transductive TL approach is implemented, which first creates a baseline model and then trains the model with a large amount of data. Then, this model is transferred to each specific user, and it will be further trained using the user-specific call data to achieve an adaptive and personalized call filter for each user. Experiment results show that the proposed approach can achieve a prediction accuracy up to 99.88\% on a validation dataset.

\textit{Summary:} As summarized in Table~\ref{tab:Summary_security}, we have discussed various TL-based approaches to address security issues such as intrusion detection, jamming attacks mitigation, anti-eavesdropping data encryption, and spam voice calls filter in wireless networks in this section. For intrusion detection, TL approaches are developed to address the lack of labeled data (by transferring a pre-trained model), the heterogeneity of devices and environment conditions (by transferring knowledge of different device types and environments), and the low learning speed (by transferring knowledge from the past experiences). For jamming attacks mitigation and anti-eavesdropping data encryption, RL is the most commonly applied technique. Since the conventional RL techniques require a long learning time to be effective, knowledge from past experiences is transferred to speed up the learning process. For spam voice calls filter, the call data are scarce and user-specific. An effective solution to simultaneously address those two challenges is to first train a baseline model with data from many users. Then, the baseline model can be fine-tuned with user-specific data to build a personalized model for each user. Although the proposed TL approaches are effective solutions to their respective problems, there are still some challenges. For example, the baseline models used for transfer require a long time to train because of the large and generalized datasets used for training. Moreover, choosing the right TL parameters, e.g., transfer rates, is crucial to the effectiveness of TL, but it is not analyzed in most of the discussed approaches. 

\begin{table*}[]
	\caption{Summary of TL Approaches for Security in Wireless Networks}
	\renewcommand{\arraystretch}{.5}
	\begin{tabular}{|>{\raggedright\arraybackslash}m{0.5cm}|>{\raggedright\arraybackslash}m{4cm}|>{\raggedright\arraybackslash}m{5cm}|>{\raggedright\arraybackslash}m{4cm}|>{\raggedright\arraybackslash}m{1.5cm}|}
		\hline 
		\multicolumn{1}{|>{\centering\arraybackslash}m{0.5cm}|}{\multirow{1}{*}{\textbf{No.}}} &
		\multicolumn{1}{>{\centering\arraybackslash}m{4cm}|}{\multirow{1}{*}{\textbf{Problem}}}&
		\multicolumn{1}{>{\centering\arraybackslash}m{5cm}|}{\multirow{1}{*}{\textbf{Knowledge to transfer}}} & \multicolumn{1}{>{\centering\arraybackslash}m{4cm}|}{\multirow{1}{*}{\textbf{ML method}}}& \multicolumn{1}{>{\centering\arraybackslash}m{1.5cm}|}{\multirow{1}{*}{\textbf{TL type}}}\\
		&&&&\\
		\hline 
		\hline 
		\cite{vu_deep_2020} & \multirow{15}{*}{Intrusion detection} & Layers of a pre-trained model &  Auto Encoders  & Transductive   \\ \cline{1-1} \cline{3-5}
		\cite{wen_time_2019} &  & Layers of a pre-trained model & CNN  &  Transductive   \\ \cline{1-1} \cline{3-5} 
		\cite{dabbagh_authentication_2019} &  & Environmental effects of nearby devices  & TL &  Transductive    \\ \cline{1-1} \cline{3-5}
		\cite{sharaf_on_2016} &  & Environmental effects  & Infinite Gaussian Mixture Model   & Transductive\\ \cline{1-1} \cline{3-5}
		\cite{zhao_the_2018} &  & Device's physical features from the past & SVM    & Inductive  \\ \cline{1-1} \cline{3-5}
		\cite{sharaf_transfer_2015} & & Fingerprint from the past & Bayesian model &    Inductive  \\ \cline{1-1} \cline{3-5}
		\cite{yu_individual_2019} & & Feature weights of transient signals & KNN &    Inductive  \\ \cline{1-1} \cline{3-5}
		\cite{wang_radio_2020} & & Knowledge from similar devices & LSTM &    Transductive  \\ \hline
		\cite{chen_reinforcement_2018} &  \multirow{6}{*}{Anti-jamming} & Knowledge from previous experiences & Q-learning  &   Unsupervised  \\ \cline{1-1} \cline{3-5} 
		\cite{han_multi_2019} &  &Knowledge from nearby regions & Q-learning   & Unsupervised   \\ \cline{1-1} \cline{3-5} 
		\cite{lu_uav_2020} &  & Knowledge from previous experiences & CNN and  Q-learning   &  Unsupervised  \\ \hline 
		\cite{do_learning_2018} & Anti-eavesdropper data encryption & Knowledge from previous experiences & Actor-critic model   &  Unsupervised \\ \hline 
		\cite{bhowmik_mtrust_2020} & Filter spam voice calls & Knowledge from other users & MLP    &  Transductive \\ \hline 
	\end{tabular}
	\label{tab:Summary_security}
	
\end{table*}

\section{Applications of Transfer Learning for Activity Recognition}\label{sec:activity_recognition}

Wireless networks, e.g., Wi-Fi or wireless sensor network, have a capability to provide low-cost wireless sensing for various activity recognition applications by capturing CSI or sensing data. Generally, the activity recognition using wireless signals can only work efficiently for a specific environment and may not be efficient to use in other environments. To this end, the TL approach can be applied as a useful learning method to provide common wireless-based activity recognition system including human activity, handwritten signature, and mobile device recognition without requiring large computing resources.    

\subsection{Human Activity Recognition}

Human activity recognition (HAR) is commonly utilized to detect various human activities using wireless signals and smart sensors in smartphones or mobile devices, aiming at improving the quality of human life including safety and health services. Specifically, the human movement data in forms of CSI signals or sensing data can be first collected from the wireless devices. Then, using the HAR system powered by ML methods, the human activities can be classified by transforming the CSI signals or sensing data into certain labels for various purposes, e.g., older people tracking, medical diagnosis, crime monitoring, and driving behaviors. Nonetheless, a human activity classification model usually only works well in one specific environment. When a new environment is employed, a new training process may be labor-intensive and time-consuming as the sufficient data and trained model require to be collected and re-built from the beginning, respectively. In this way, TL approaches can be used as very useful methods to recognize the human activities accurately while addressing the above challenges efficiently.

\subsubsection{CSI Signal-Based Human Activity Recognition}

The CSI, which describes how the wireless signals propagate between transmitters and receivers in the communication link, can be used to recognize human acticity. The CSI as the channel matrix can be estimated using transmitted and received signals of wireless devices after passing several signal processings, e.g., cyclic prefix removal, demapping, and OFDM demodulation. Compared to the RSS, the CSI can observe the change of fine-grained environment (including finding the causes of the environment changes and recognizing how the environment changes) more effectively, and thus it is applicable to be used for human activity recognition. However, due to the lack of sufficient CSI signal data and high cost to obtain them when the environment changes, the use of TL approach can help to improve the efficiency of using CSI to recognize the human activity. 

The adoption of TL approach leveraging CSI signals has been
investigated recently to recognize hand and arm movements. In~\cite{shang_a_2017}, the authors present an inductive TL-based sign-language recognition system, i.e., WiSign, using time-series of CSI with a sparse labeled dataset. In particular, the system can recognize and extract the meaning of particular hand or arm movements, e.g., ``Yes/No'', ``Hello/Goodbye'', ``Thanks'', and ``Sorry'', according to the study of various CSI waveforms. In this case, an inductive TL method, i.e., by transferring knowledge of the CSI waveforms from other users to train a new classifier for a new user, can be applied to achieve a higher sign language accuracy. Using the TL method, the system only needs few labeled training samples and can prevent the use of a generative model. Through real experiments using two Wi-Fi devices, it is demonstrated that the proposed system can achieve a higher mean prediction accuracy than that of classical SVM. Instead of using the CSI waveforms directly for the TL process, the authors in~\cite{bu_wi-fi_2018} propose a DTL-based classification framework for gesture recognition through CSI waveforms-to-image conversion. Specifically, the framework first converts the amplitude of CSI into an image matrix. Using an existing CNN pre-trained model on a large number of images, the TL method can be applied to reduce the training time and extract the pre-trained model's important features (by implementing fine-tuning process for the classification tasks). From the real experiments using 8 gesture positions, the gesture recognition can achieve up to 100\% accuracy.

\begin{figure}[!t]
	\centering
	\includegraphics[scale=0.4]{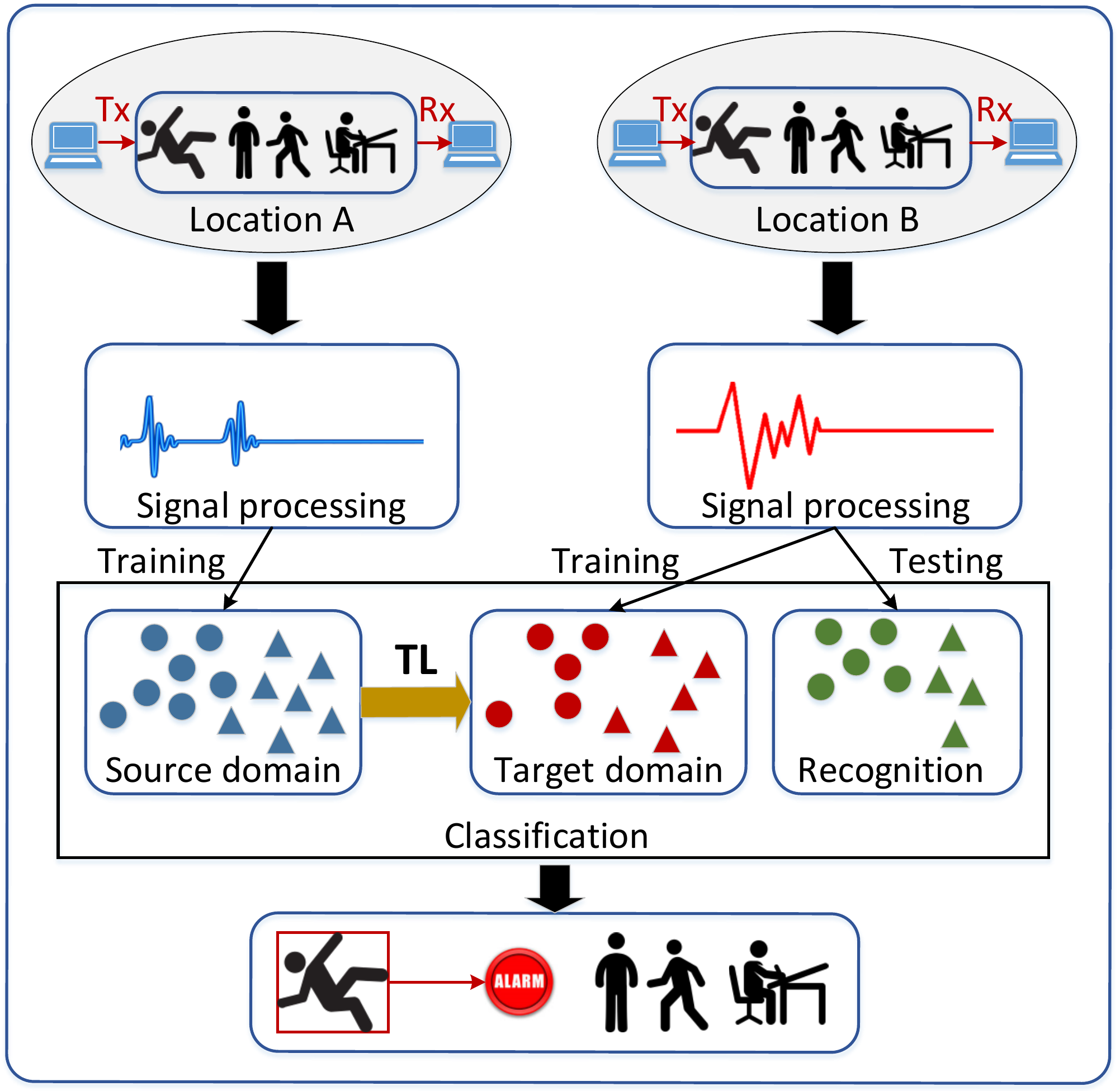}
	\caption{The TL framework for human fall detection~\cite{zhang_commercial_2019}.}
	\vspace{-0.5em}
	\label{fig:human_1}
\end{figure}

Instead of using the full component of CSI signals in the above works, the frequency component of CSI signals can be utilized to detect the velocity of human body movement for human fall detection. Specifically, the authors in~\cite{zhang_commercial_2019} discuss an inductive TL-based human activity recognition for fall detection system, i.e., TL-Fall. Typically, a fall detection system in a certain condition does not work well for other conditions due to unique information where the falling occurs, i.e., when a specific person (e.g., elderly or disable person) loses his/her balance and hits the ground. To cope with this problem, the TL-Fall accounts for an SVM-based TL to transfer falling velocity changes knowledge from the trained labeled data in the old condition to the new condition with few labeled data (as illustrated in Fig.~\ref{fig:human_1}). In this case, the system first collects the amplitude of CSI from multiple OFDM subcarriers of Wi-Fi signals with IEEE 802.11n standard. Then, the frequency component of the CSI can be extracted to determine the velocity of human body changes when falling happens. Based on the experimental results in an indoor environment, the TL-Fall outperforms other baseline ML-based fall detection methods in terms of fall detection sensitivity, i.e., the percentage of correctly detecting falling activities. 

Motivated by the CSI extraction in~\cite{zhang_commercial_2019} and CSI-image transformation method in~\cite{bu_wi-fi_2018}, the authors in~\cite{arshad_leveraging_2019} propose an inductive TL-based framework to detect multiple human activities, i.e., TL-HAR. Specifically, using the variance of MIMO subcarriers, the activity-based CSI is first collected and transformed into CSI images. Then, the specific features from CSI images of human activities are trained using deep-CNN and inferred-CNN. This pre-trained model can be modified by fine-tuning several layers which are closer to the output of this model. The modified model can then be transferred to perform the TL process for other human activity dataset training with shorter training time. Via the simulation, it is shown that the TL-HAR can enhance the classification accuracy up to 3\% and 5.6\% for single and multiple MIMO links, respectively, compared with those of other conventional ML methods. Instead of transforming the CSI data to images, the authors in~\cite{ding_device_2020} investigate a location-independent inductive TL with CNN-based human activity recognition model using the estimation of channel frequency response containing the amplitude attenuation. Particularly, the CSI data from certain locations are first collected and denoised to remove the noise from raw CSI data. Then, a pre-trained CNN model is built to obtain the feature extractor based on the characteristic of the CSI amplitudes. From this feature extractor, a perceptive-based TL approach is used to predict/classify new i.i.d samples online in different locations. Applying such the TL method, the new activities can be recognized without training the new samples or with only using few training samples. Through the experiments, the proposed framework can reach the activity classification accuracy more than 90\% and outperform those of other baseline ML methods, e.g., Singular Value Decomposition (SVD), Non-negative Matrix Factorization (NMF), and KNN, for six different activities.

To investigate the correlation between the spatial and temporal information from CSI, the authors in~\cite{sheng_deep_2020} introduce an inductive DTL framework for human body movement which jointly considers the spatial features trained from CNN and temporal features learned from B-LSTM. In particular, the spatial features first can be extracted from the CSI streams using the CNN. Then, the fully-connected layer of CNN is used as the B-LSTM input to extract the temporal features and produce a pre-trained model. Some specific layers from the spatio-temporal pre-trained model can be fine-tuned (while other layers remain fixed) to implement the inductive TL process, aiming at training a new environment with a smaller number of samples. Through experiments using wireless router to classify four different human movements, the proposed framework can achieve overall accuracy 96.96\% with low computational cost.   

\subsubsection{Sensor-Based Human Activity Recognition}

In addition to the reflected signals, i.e., CSI, wearable sensor devices can be used to detect human activities. In~\cite{zhou_abnormal_2020}, the authors introduce a light DTL framework to detect abnormal human activities, e.g., fall detection, at edge servers. Specifically, the users' smart devices first send raw data (captured via triaxial accelerator sensors) to their nearest edge servers. Then, a CNN-based prediction model is generated to recognize the abnormal activities  at the edge servers. Using the pre-trained CNN model, the high-level embedding features can be extracted to perform the inductive TL process for other abnormal activity models using conventional ML methods, i.e., LR, KNN, DT, NV, RF, and SVM. From the experiments, the framework can provide higher prediction accuracy up to 28\% compared with those of the conventional ML methods without using TL. The development of TL approach without using the edge servers is also discussed in~\cite{rokni_autonomous_2018} where collaborative wearable devices can be used to automatically detect real-time physical human movements. In particular, there are three main steps to apply an inductive TL method. First, two wearable sensors (referred to as \emph{source view}) are attached to train a base model for the movement recognition. Then, a new wearable sensor (referred to as \emph{target view}) is embedded to train a new model at the target view. Upon completing the new model training, the source and target views (referred to as \emph{multi-view learning}) can be used together to construct a new classification model for activity recognition with a higher accuracy. Via the experiments, it is shown that the activity recognition accuracy can be improved up to 9.3\% at the level 83.7\% compared with those of conventional approaches where the model from the source view is re-used directly to the target view without any refinement. 

Instead of sending raw data to the edge data center, the authors in~\cite{valerio_hypothesis_2016} propose a distributed ML approach where the collected raw measurement data, i.e., 3-axial linear acceleration and 3-axial angular velocity, from embedded accelerator and gyroscope sensors can be trained locally at each user's smartphone. In particular, the collected data from the smartphone's sensors are first pre-processed and sampled locally to obtain a vector of multiple features in time and frequency domain. Then, by adopting a hypothesis transfer learning (HTL), some smartphones with sufficient number of datasets can train local partial models (referred to as \emph{source models}) locally. From the trained partial models, a unique complete model can be obtained by accumulating the source models from the training smartphones. This complete model can then be transferred via wireless network to learn a more accurate model at another smartphone (referred to as \emph{target model}) with insufficient number of datasets, thereby improving the human activity classification accuracy at the smartphone. Considering three static activities, (i.e., standing, sitting, and lying), and dynamic activities (i.e., walking, walking downstairs, and walking upstairs, and static-to-dynamic transitions), the proposed TL method can achieve the comparable classification accuracy performance with that of centralized learning while reducing the wireless network overhead significantly.   

\begin{figure}[!t]
	\centering
	\includegraphics[scale=0.32]{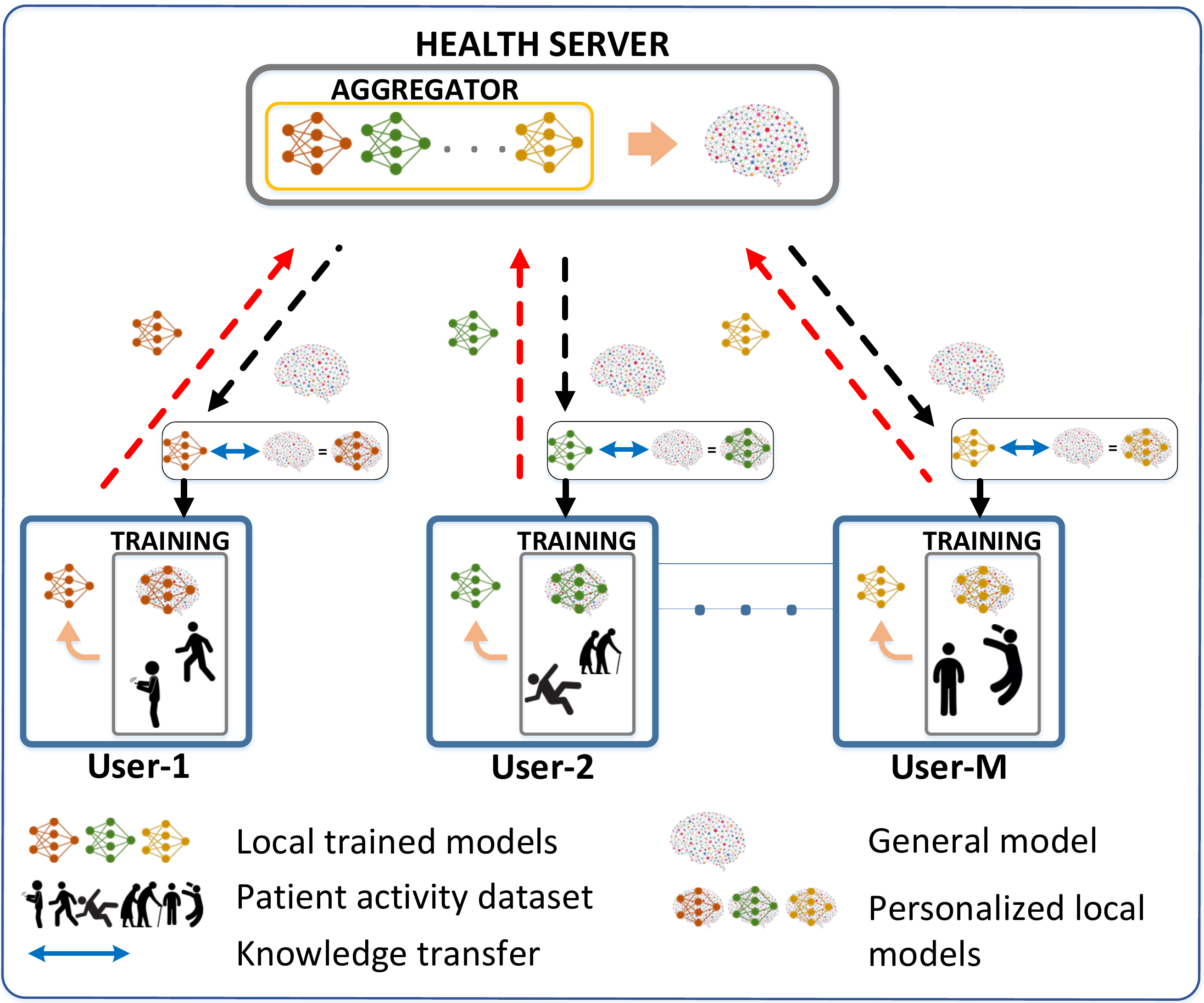}
	\caption{The federated TL framework for healthcare activity system~\cite{chen_fedhealth_2020}.}
	\vspace{-0.5em}
	\label{fig:human_2}
\end{figure}

Due to the benefits of using distributed ML for sensitive data privacy and security, the authors in~\cite{chen_fedhealth_2020} introduce a federated TL-based framework, i.e., FedHealth, leveraging CNN for wearable activity recognition in the healthcare application system. This framework can deal with personalization issues where the daily health activity pattern and physical features for each individual user can be trained differently. Particularly, each user with a wearable device can train a lightweight model locally without sharing any actual data with other users. The local trained models from all the users are then collected and aggregated by the cloud server with a homomorphic encryption to update the global cloud model. To incorporate a unique new model for each user, the useful common knowledge from the cloud model can be utilized by the previous local trained models of the users as shown in Fig.~\ref{fig:human_2}. In this case, the low-level layers of CNN, i.e., convolution and max-pooling layers, which contain the common activity recognition are frozen. Meanwhile, the fully connected layers which consist of more specific features for each user can be updated using an additional correlation aligment layer between the cloud model and the user's local trained model. Using human activity UCI Smartphone dataset with 6 activities from 30 users~\cite{anguita_human_2012} and self-producing Parkinson's disease dataset, the proposed TL approach can improve the classification accuracy up to 5.3\% and 21.6\%, respectively, compared with those of other conventional ML methods, i.e., traditional DL, KNN, SVM, and RF.

\subsubsection{Probe Response Signal-Based Human Activity Recognition}

Generally, capturing CSI waveforms may require user access to the real Wi-Fi hotspots, and thus some authentication procedures need to be performed. Furthermore, the use of real Wi-Fi network needs continuous data traffic even when the Wi-Fi hotspots are idle through transmitting beacon signals. In practice, some emergency services, e.g., search-rescue applications, may have urgent cooperation with the wireless network without using the beacon signals. For that, the authors in~\cite{shi_passive_2019} present a TL-based passive activity recognition using only Wi-Fi probe response signals generated from a local Wi-Fi device, e.g., Raspberry Pi, without connecting to the actual Wi-Fi network. Specifically, the probe request-response mechanism of Wi-Fi can enable the information exchanges between the local Wi-Fi AP and its users without any encryption or modification to the AP. Using the probe response signals, Doppler measurements to implement human activity recognition can be employed. Nonetheless, due to the low duty cycle of the probe response signal transmission, the classification accuracy may be degraded. For that, an inductive TL approach can be used to improve the accuracy through fine-tuning the convolution layer of the Doppler CNN model. By utilizing the TL method with a probe response signal dataset containing six human activities, the classification accuracy can be improved up to 7\% with significant training time reduction.

\subsection{Handwritten Signature Recognition}

In addition to the human activity recognition, the inductive TL method can be utilized to verify a specific person's identity based on his/her handwritten signature as discussed in~\cite{jung_transfer_2019}. Specifically, the Wi-Fi gesture signals of the handwritten signature for a specific position can be first collected through the reflected Wi-Fi CSI. Then, a gradient descent CNN-based algorithm is performed to pre-train a model (as the feature extractor). In this case, the feature extractor contains convolution layers, activation functions, max-pooling layers, and fully connected layers which correspond to the classifier of the pre-trained model. As such, only the common knowledge with the best classification accuracy will be used for the TL process. Upon obtaining the feature extractor, the common knowledge from this particular position can be transferred to detect signature signals from other positions using the kernel and range space projection learning~\cite{toh_kernel_2018}. Since the Wi-Fi signature signals for diverse positions have the same data domain, only the model classifier is required to be re-trained. Based on the experimental study using thousands of Wi-Fi signature signals from 50 user IDs, the proposed method can achieve 96.75\% accuracy with faster training CPU/GPU computation time compared with those of typical gradient descent and convolutional network algorithms.

\emph{Summary}: In this section, we have investigated TL approaches for the applications of wireless-based human activity recognition in wireless networks based on the full/partial component of CSI signals, sensing data, and probe response signals. In particular, due to the diverse human activities in various conditions, e.g., locations and environments, the TL method can be efficiently applied to provide generalized knowledge transfer which can improve the recognition accuracy with less training data and shorter learning time. In this case, the inductive DTL approaches, e.g., CNN and LSTM, have been widely used to improve the activity recognition accuracy by extracting the non-linear features deeply. Additionally, the adoption of distributed ML methods, e.g., HTL and FTL, can further improve the accuracy of activity recognition model for each individual user due to the mutual sharing of local trained models with minimum private data disclosure. Based on the above works, the inductive TL is considered as the most popular TL approach to recognize the human activity where the activities in the target domain are a subset of those are in the source domain. The extension capability of practical human activity recognition to new activities unseen in the source domain using transductive and unsupervised TL approaches will be potential to explore for future research works. The summary of all the TL approaches for human activity recognition applications is presented in Table~\ref{tab:Summary_human}.

\begin{table*}[]
	\caption{Summary of TL Approach for Activity Recognition}
	\renewcommand{\arraystretch}{.5}
	\begin{tabular}{|>{\raggedright\arraybackslash}m{0.5cm}|>{\raggedright\arraybackslash}m{2.8cm}|>{\raggedright\arraybackslash}m{6.8cm}|>{\raggedright\arraybackslash}m{4.7cm}|>{\raggedright\arraybackslash}m{1cm}|}
		\hline 
		\multicolumn{1}{|>{\centering\arraybackslash}m{0.5cm}|}{\multirow{1}{*}{\textbf{No.}}} &
		\multicolumn{1}{>{\centering\arraybackslash}m{2.8cm}|}{\multirow{1}{*}{\textbf{Problem}}}&
		\multicolumn{1}{>{\centering\arraybackslash}m{6.8cm}|}{\multirow{1}{*}{\textbf{Knowledge to transfer}}} & \multicolumn{1}{>{\centering\arraybackslash}m{4.7cm}|}{\multirow{1}{*}{\textbf{ML method}}}&
		\multicolumn{1}{>{\centering\arraybackslash}m{1cm}|}{\multirow{1}{*}{\textbf{TL type}}}\\
		&&&&\\
		\hline 
		\hline 
		\cite{shang_a_2017} &  \multirow{10}{*}{CSI signal-based human } & CSI waveform map &  SVM-based TL &Inductive      \\ \cline{1-1}\cline{3-5}
		\cite{bu_wi-fi_2018} & \multirow{10}{*}{activity recognition} & CSI map & CNN-based TL &Inductive      \\ \cline{1-1}\cline{3-5}
		\cite{zhang_commercial_2019} &  & Knowledge from falling velocity changes  & SVM-based TL &Inductive      \\ \cline{1-1}\cline{3-5}
		\cite{arshad_leveraging_2019} &  & Knowledge from CSI images & Deep and inferred CNN-based TL& Inductive      \\ \cline{1-1}\cline{3-5}
		\cite{ding_device_2020} &  & Knowledge from the characteristic of CSI amplitudes & CNN-based TL& Inductive      \\ \cline{1-1}\cline{3-5}
		\cite{sheng_deep_2020} &  & Knowledge extracted from spatial and temporal features & CNN and B-LSTM-based TL &Inductive      \\ \hline
		\cite{shi_passive_2019} & Probe response signal-based human activity recognition & Knowledge from probe response signals & CNN-based TL &Inductive      \\ \hline 
		\cite{zhou_abnormal_2020} & \multirow{8}{*}{Sensor-based human } & Knowledge from embedding features of sensing data & CNN-based TL & Inductive      \\ \cline{1-1}\cline{3-5}
		\cite{rokni_autonomous_2018} & \multirow{8}{*}{activity recognition} & Knowledge from movement detection of wearable sensors & Synchronous multi-view TL & Inductive      \\ \cline{1-1}\cline{3-5}
		\cite{valerio_hypothesis_2016} &  & Knowledge from features in time and frequency domain & Hypothesis TL & Inductive      \\ \cline{1-1}\cline{3-5}
		\cite{chen_fedhealth_2020} &  & Common knowledge from cloud model to obtain personalized model & CNN-based TL & Inductive      \\ \hline \hline
		\cite{jung_transfer_2019} & Identity recognition & Knowledge from reflected or interfered Wi-Fi CSI signals of in-air handwritten signatures & CNN-based TL & Inductive      \\ \hline
	\end{tabular}
	\label{tab:Summary_human}
	
\end{table*}


\section{Applications of Transfer Learning for Wireless Caching}\label{sec:caching}

\begin{figure*}[!t]
	\centering
	\includegraphics[scale=0.37]{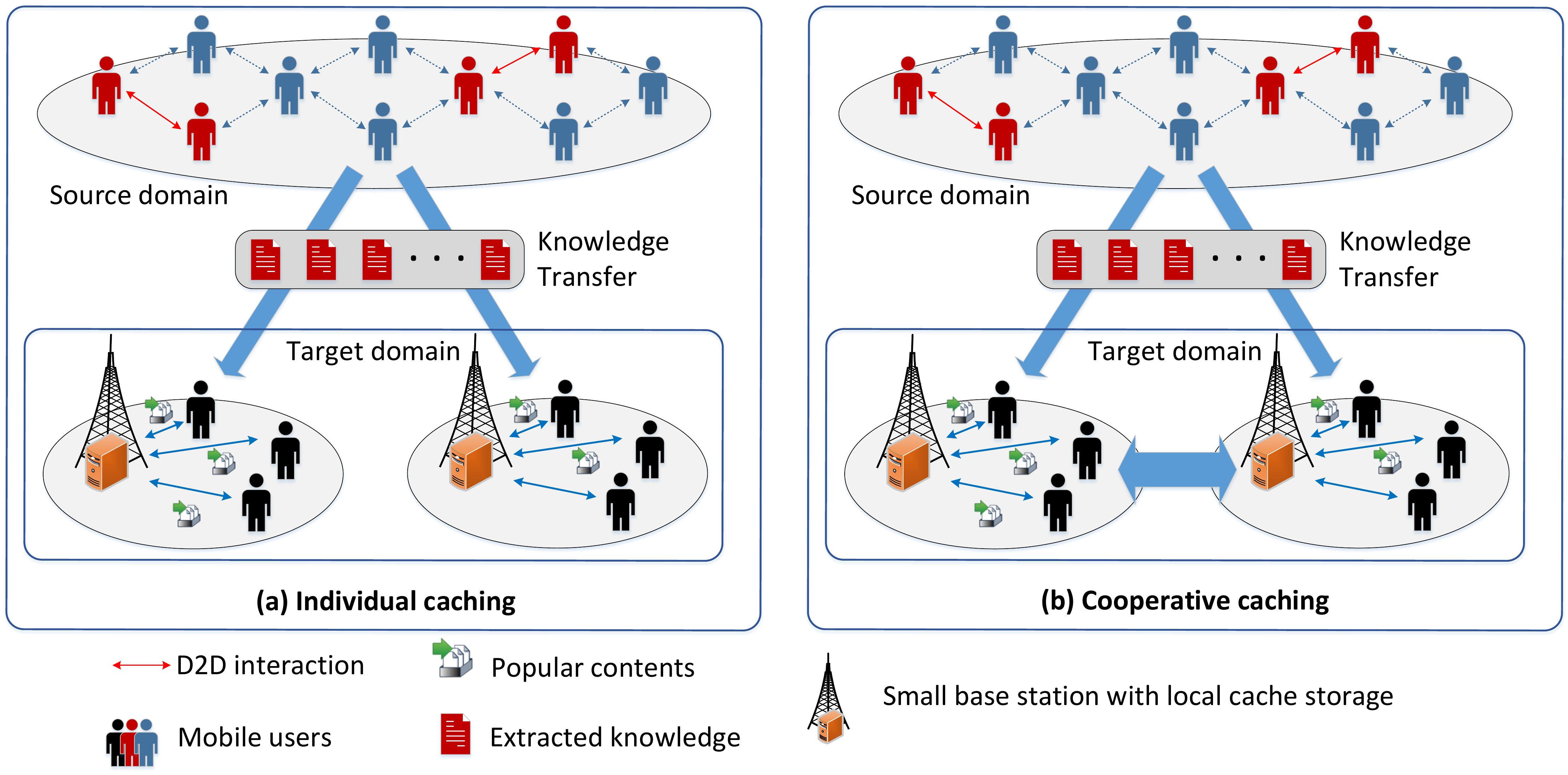}
	\caption{The knowledge transfer of D2D interactions in the individual and cooperative caching scenarios.}
	\vspace{-0.5em}
	\label{fig:caching}
\end{figure*} 


Content caching in wireless edge networks has been proposed as one of the most effective paradigms to deal with a massive data traffic demand through distributing popular contents at edge servers (which are deployed in a close proximity to requesting users). To predict the popularity characteristic of contents at the edge servers with high accuracy, proactive caching leveraging ML-based methods can be applied~\cite{shuja_applying_2020}. However, in practice, it is costly and intractable to collect sufficient number of training dataset to create efficient prediction models for content popularity. Moreover, the content popularity matrix containing the correlation between users and contents frequently suffers from data sparsity problem. For that, TL approaches can be used to address these issues. In this section, we discuss the use of TL to cope with the drawbacks of caching applications in wireless networks for two scenarios, i.e., individual caching and cooperative caching.

\subsection{Individual Caching}

In this individual caching scenario, the knowledge produced by the TL method can be transferred separately to each individual caching storage of the small cells, e.g., small BSs (SBSs). Thus, there is no cached-content sharing among the participating small cells. The content popularity in the network can be predicted at both small and macro BSs as follows.

\subsubsection{Content Popularity Prediction at Small BSs}

In~\cite{bastug_anticipatory_2014}, the authors propose an inductive TL-based content popularity prediction policy as illustrated in Fig.~\ref{fig:caching}(a). Specifically, the TL approach can first extract knowledge from content-access history of Device-to-Device (D2D) users within their social community (source domain). Then, using the extracted knowledge, the learning process for the large-scale content popularity matrix in SBS networks (target domain) can be improved. Compared to the random caching and collaborative filtering methods, the proposed TL method can achieve much closer backhaul offloading gain to the ground truth of known popularity matrix for all scenarios with various storage ratios. Extended from the work in~\cite{bastug_anticipatory_2014}, the authors in~\cite{bastug_atra_2015} develop an inductive TL-based caching strategy executed at SBSs under the storage constraints, workload, data sparsity, and unknown spatio-temporal traffic demands. In particular, the hidden latent contextual information including the users' content-access history and social ties is obtained from the interactions of D2D users (source domain). To efficiently cache well-received contents at the SBSs, the obtained information is then utilized to predict the content popularity matrix with unknown users' ratings in the SBS networks (target domain). Based on the simulation results, it is shown that the proposed TL method can improve the satisfaction ratio (i.e., the total requests that are served immediately within a time window over the total requests from all users) and backhaul offloading gain up to 22\% compared with those of the random caching and collaborative filtering methods when various storage sizes, traffic intensity, and backhaul capacity are taken into account. 

\subsubsection{Content Popularity Prediction at Macro BSs}   

The use of inductive TL to predict the content popularity in heterogeneous networks considering the minimum training time is proposed in~\cite{nagaraja_caching_2015, nagaraja_a_2016}. Instead of estimating the content popularity at each SBS, the SBS only incorporates a random caching strategy. Meanwhile, the content popularity prediction is conducted at a macro BS in a centralized manner. The predicted content popularity is then used to optimize the caching strategy through minimizing the training time due to the requested content unavailability. Nonetheless, the minimum training time can be very high to achieve a required prediction accuracy when the training process is performed from the scratch. As such, the TL approach can be applied to further reduce the training time by transferring important knowledge from the user-content interaction (source domain), e.g., popular cartoon movies for kids and popular action movies for adults, to the users' request pattern (target domain). When the number of source domain samples increases, the proposed TL method can reduce the training time linearly compared with the learning without TL method where the training time remains the same for the whole performance.
In~\cite{zhang_cache_2018}, the authors investigate an efficient dynamic video rate allocation mechanism considering a combination of DRL, imitation learning, and unsupervised TL approaches. In particular, the allocation problem is first modeled as a Markov desicion process problem. Then, a DQN, imitation learning, and the unsupervised TL approaches are utilized to find sub-optimal allocation policy. In this case, the TL approach can transfer the knowledge of DQN parameters and fine-tune them to map the video rate from source to target domain. Leveraging the TL method, the mechanism can obtain the sub-optimal allocation policy faster. From the performance evaluation, the proposed method can outperform other conventional learning methods, e.g., DNNs and random rate allocation, in terms of the average user's reward, i.e., QoS of video streaming. 

\subsection{Cooperative Caching}

Although the works in~\cite{bastug_anticipatory_2014,bastug_atra_2015} demonstrate the superiority of using TL approaches, they assume that the content popularity is predicted independently at each SBS, leading to the redundant caching problem. Furthermore, the authors in~\cite{nagaraja_caching_2015,nagaraja_a_2016,zhang_cache_2018} consider that the content popularity estimation is performed at the centralized BS, thereby increasing the workload and operational cost at the BS. To address these challenges, the authors in~\cite{hou_proactive_2017} propose an inductive TL-based proactive caching mechanism under the cooperative caching scenario among edge servers as shown in Fig.~\ref{fig:caching}(b). In this case, the following two-step proactive caching optimization framework is considered. First, the framework trains the content popularity model by using the TL approach with small training time at multiple edge servers in a cooperative scenario, i.e., where each edge server
with a local caching storage can share its cached contents with other edge servers. Using this trained model, the requested content information at another edge server can be predicted. In other words, the target and source domains are the content-access information to be predicted at an edge server and those are at other edge servers in the cooperative scenario, respectively. Then, a content placement policy can be derived using a greedy algorithm. By using such a framework, the cache hit rate can be improved up to 177\% compared with those of the baseline caching methods.
To further improve the prediction accuracy, a joint ML utilization using a regularized singular value decomposition-based collaborative filtering (RSVD-based CF) and inductive TL methods is investigated in~\cite{wang_a_2017}. Particularly, the content popularity matrix is first predicted by using the RSVD-based CF method. The TL method is then utilized to enhance the prediction accuracy under the data sparsity through transferring hidden features, e.g., a social network or a D2D network, from source to target domain. Based on the predicted content popularity, a decentralized iterative algorithm is executed to obtain the optimal caching policy. Via the simulation results, the proposed approach can reduce the content request delay, and thus improve the users' satisfaction.

\emph{Summary}: In this section, we have discussed the use of TL to address the limitations of caching applications in wireless networks. Specifically, the content popularity prediction may suffer from low accuracy and long training time when typical ML methods are utilized. In this case, the TL approaches are applicable to solve the aforementioned problems through transferring knowledge, e.g., user-content interaction, social network, and video rate map, without a need of training process from the scratch. Furthermore, the use of TL-DRL and RSVD-based TL (inductive TL) can be very efficient to improve the average user's reward and prediction accuracy, respectively. With the TL approaches, the training time of the content popularity prediction can also be reduced, thereby minimizing the content-access delay for the requesting users. However, from all the above works, they assume that the popularity matrix is constant over time. In practice, the popularity matrix may change frequently due to dynamic preferences of many mobile users. To this end, the use of TL approaches which are robust to the popularity matrix changes needs to be further evaluated to maintain high cache hit ratio and low delay of the requested contents for different time. The summary of the TL approaches for caching applications is presented in Table~\ref{tab:Summary_caching}.


\begin{table*}[]
	\centering
	\caption{Summary of TL Approach for Caching}
	\begin{tabular}{|>{\raggedright\arraybackslash}m{1.5cm}|>{\raggedright\arraybackslash}m{3cm}|>{\raggedright\arraybackslash}m{5cm}|>{\raggedright\arraybackslash}m{4cm}|>{\raggedright\arraybackslash}m{1.5cm}|}
		\hline 
		\multicolumn{1}{|>{\centering\arraybackslash}m{1.5cm}|}{\multirow{1}{*}{\textbf{No.}}} &
		\multicolumn{1}{>{\centering\arraybackslash}m{3cm}|}{\multirow{1}{*}{\textbf{Problem}}}&
		\multicolumn{1}{>{\centering\arraybackslash}m{5cm}|}{\multirow{1}{*}{\textbf{Knowledge to transfer}}} & \multicolumn{1}{>{\centering\arraybackslash}m{4cm}|}{\multirow{1}{*}{\textbf{ML method}}}&
		\multicolumn{1}{>{\centering\arraybackslash}m{1.5cm}|}{\multirow{1}{*}{\textbf{TL type}}}\\
		\hline 
		\hline 
		\cite{bastug_anticipatory_2014, bastug_atra_2015} &   & D2D user interactions & Collaborative Filtering      &Inductive      \\ \cline{1-1}\cline{3-5}
		\cite{nagaraja_caching_2015,nagaraja_a_2016} & Non-cooperative caching & User-content interaction & TL-based popularity estimator       &Inductive      \\ \cline{1-1}\cline{3-5}
		\cite{zhang_cache_2018} & & Video rate map & TL-DRL&Unsupervised      \\ \hline \hline
		\cite{hou_proactive_2017} & \multirow{2}{*}{Cooperative caching} & Knowledge from content popularity & K-means clustering &Inductive      \\ \cline{1-1}\cline{3-5}
		\cite{wang_a_2017} & & Hidden features of
		popularity matrix & Regularized
		SVD-based CF &Inductive      \\ \hline
	\end{tabular}
	\label{tab:Summary_caching}
	
\end{table*}



\section{Challenges, Open Issues, and Future Research Directions}\label{sec:challenge}
Although the surveyed works have demonstrated the effectiveness of TL in solving various problems in wireless networks, there are still many challenges and open issues. In the following, we discuss the current challenges, open issues, and future research direction of TL for future wireless networks.

\subsection{Challenges and Open Issues}
\subsubsection{Determine the Source Tasks in Wireless Networks} The effectiveness of TL depends greatly on the similarity between the source and the target task. Nevertheless, choosing the right source task is usually not trivial, especially in wireless network applications where there are multiple sources available, e.g., from nearby users, access points, femtocells, and cellular base stations. For example,~\cite{zhao_spectrum_2020,cao_spectrum_2020,shah_fast_2020,shah_deep_2018}choose sources based on their physical distances to the target. The reason is that in this case, the surrounding environments of the source and the target have high similarities, and thus their Q-values might be similar. Alternatively, TL approaches such as~\cite{parera_transfer_2019} and~\cite{yang_machine_2019} choose to transfer knowledge of the same task in different frequency bands, whereas some other approaches transfer knowledge from previous experiences. However, we can observe that most of current works do not well justify why the sources are chosen. If the source and target's domain are not highly related, TL approaches can even make the target model performing worse than just training with the target data. Thus, solutions and performance analyses for selecting sources in wireless networks need to be further investigated.
\subsubsection{Determine What to Transfer in Wireless Networks} Once the source is chosen, the next step is to determine which part of the source's model to be transferred. 
Generally, the first layers of the source model capture the more general features that are similar between the two domains, whereas the last layers are more domain-specific. For example, in wireless applications where there are multiple cognitive agents in the same radio environment, more layers can be transferred. However, in other applications, such as~\cite{li_tact_2014}, where the knowledge is transferred from neighboring regions, only the first few layers should be transferred. Therefore, determining how many layers to transfer is a very challenging task, because it requires an in-depth analysis of the similarities between the source and target. Despite its importance, there are few works that attempt to address this challenge. In~\cite{parera_transfer_2019} and~\cite{parera_transfer_2020b}, the authors determine the number of layers to transfer through simulations, i.e., testing different numbers of layers and choose the one with the best results. Therefore, more experiments and simulations should be performed to better determine layers to transfer according to different wireless applications, e.g., cognitive radio, localization, and activity recognition. 
\subsubsection{Determine the TL Parameters in Wireless Networks} The transfer rate parameter has a significant impact on the effectiveness of the TL process. This parameter dictates how many percent of the source model's weights are transferred to the target model's weights. Few approaches such as~\cite{koushik_intelligent_2017} test different transfer rates by simulations and conclude that a high transfer rate (0.8) leads to better results compared to those of lower transfer rates (0.2 and 0.5). Nevertheless, it is shown in~\cite{li_tact_2014} that a transfer rate of 0.5 achieves more desirable results. Thus, we can observe that the setting of TL parameters largely depends on the specific context of the application. Moreover, most of the surveyed approaches transfer the whole weights (i.e., the transfer rate is 1) without any clarification. Tuning such an important parameter is crucial to the effectiveness of the TL process, yet attempts to address this challenge are still limited.
\subsubsection{Trade-off between Performance and Training Time} Among the surveyed TL approaches, some approaches freeze the first layers of the target model and only fine-tune the last layers, whereas the other approaches fine-tune all the layers. Generally, the more layers we fine-tune, the better the model can perform because all the weights are adjusted to the target's domain. However, this also necessitates more training time. Thus, determining the number of layers for fine-tuning depends greatly on the wireless network applications. For example, in applications with a longer time window for training, such as channel prediction~\cite{parera_transfer_2019} and radio map construction~\cite{parera_transfer_2020b}, more layers can be fine-tuned compared to latency-sensitive applications such as embedded IoT~\cite{restuccia_deepwierl_2020}. Moreover, in the case where the target data is limited, e.g., new environments, fine-tuning all the layers may even lead to overfitting (as shown in the results of~\cite{yuan_transfer_2020} and~\cite{yang_deep_2020}) since the available labeled data do not represent the domain well. Thus, choosing an appropriate number of layers for fine-tuning remains an open issue.
\subsection{Future Research Directions}
\subsubsection{Beamforming}Beamforming refers to the techniques that focus the signal to the receiving devices instead of spreading the signal in all directions. To find the optimal beamforming solution, conventional ML techniques have been widely adopted. However, the obtained beamforming solution is only optimal for a specific network setting, and training ML models for new settings takes a lot of time. To address this issue, TL can be a potential solution. Particularly, a DL model consisting of a conventional CNN and a signal processing module is proposed in~\cite{xia_model_2014}. The signal processing module can be placed before the input or after the output of the CNN to convert key features from expert knowledge to the target beamforming matrix. A TL process is then proposed to transfer the knowledge from different network settings to the target task. Specifically, the CNN model of the source task is transferred to the target task, and then new input and output layers will replace the corresponding layers of the source model. Finally, the new target model is fine-tuned with the target task data. However, there are no simulations or experiments conducted to evaluate the TL process. In~\cite{yuan_transfer_2020}, a beamforming NN is proposed for beamforming optimization for SINR load balancing, which includes a conventional CNN and a beamforming recovery module at the output of the CNN. Since the SINR load balancing problem has some common features across different wireless environments, a transductive TL process is proposed to utilize the knowledge from different environments. Different from the TL approach in~\cite{xia_model_2014},~\cite{yuan_transfer_2020} proposes to transfer a pre-trained model to the target task. Specifically, all the layers except the fully connected layer, of the pre-trained network are frozen, and the target data is only used to train the fully connected layer. The reason for training only the last layer is that the lack of data at the target task may lead to overfitting of the target model if all the layers are trained~\cite{yuan_transfer_2020}. Moreover, this TL strategy is also beneficial in terms of training time, as only the last layer needs to be trained. Simulation results demonstrate that the proposed TL process helps to improve the convergence rate of the CNN. Although
beamforming optimization consists of a wide variety of tasks, there are currently very few works that utilize TL for beamforming. The wireless network environments frequently change, and consequently new data need to be collected and the beamforming ML models need to be trained again. Thus, TL is a very promising solution to reduce the need for new data and training time for beamforming designs.
\subsubsection{Embedded IoT Devices} 
\begin{figure}[!t]
	\centering
	\includegraphics[width=\columnwidth]{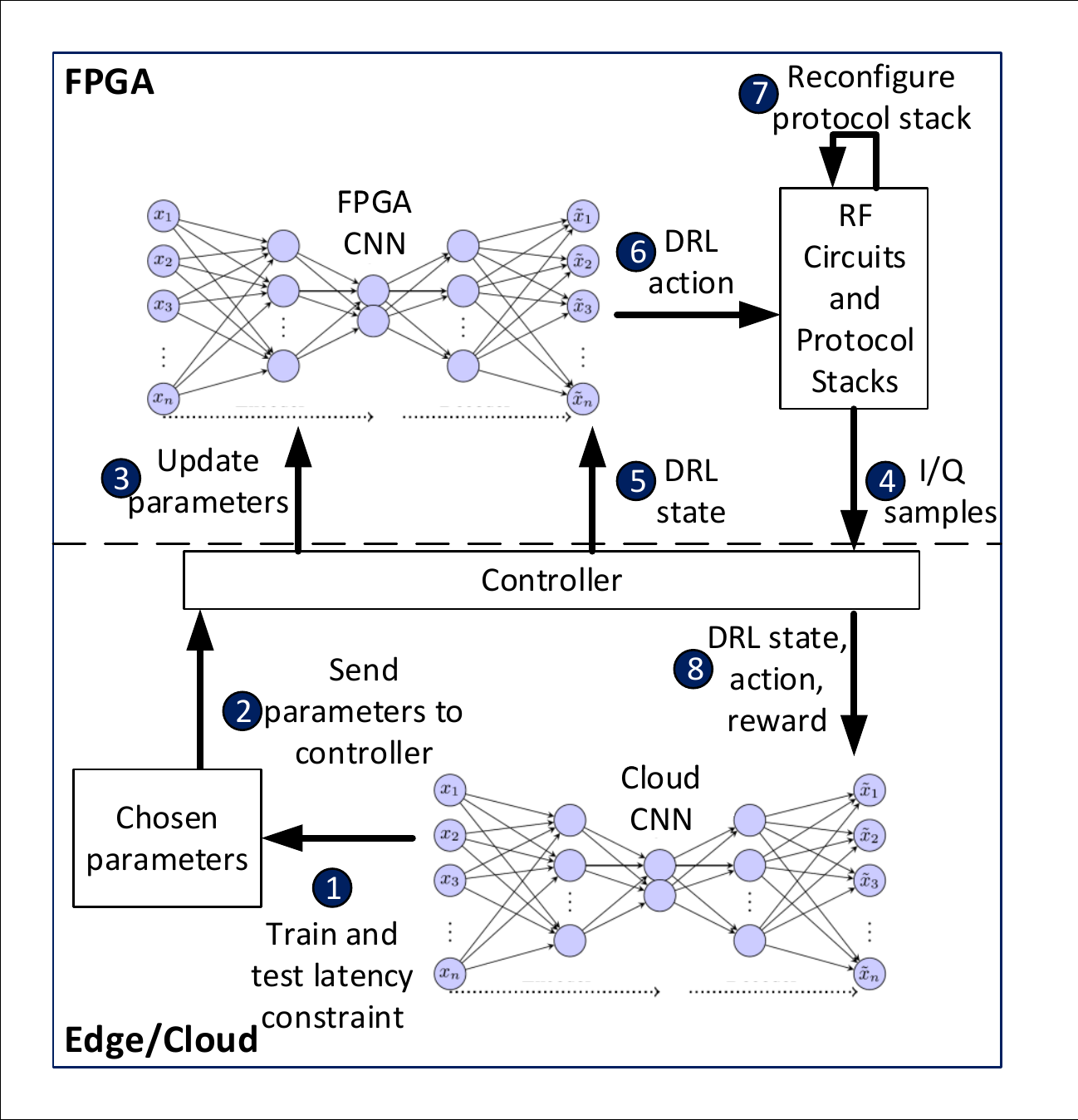}
	\caption{The TL-based approach for embedded IoT devices~\cite{zhao_using_2015}.}
	\vspace{-0.5em}
	\label{fig:challenge}
\end{figure}

Embedded IoT devices are significantly restricted in terms of computational resources and latency requirements. Consequently, this hinders the implementation of ML techniques on embedded IoT devices. To address this issue, TL can be an effective solution because the training process can take part in a central node and the target model on an embedded device only needs to fine-tune shortly. In~\cite{restuccia_deepwierl_2020}, the authors propose to utilize TL for embedded IoT devices. Particularly, a DRL framework is developed for embedded IoT wireless devices, including a one-dimensional CNN located on the FPGA, and another CNN located on the edge/cloud. The CNN on the FPGA receives I/Q samples as input and learns to optimally reconfigure the wireless protocol stack in real-time. Due to the strict time constraints, e.g., 20ms, and the limited computational capacity of the FPGA, it is not possible for the CNN to learn the optimal action. Therefore, a TL approach is developed as illustrated in Fig.~\ref{fig:challenge}, which first trains the model on the edge/cloud with different spectrum states until it can achieve a high accuracy. Then, to address the latency issue, the model is tested and its parameters are reduced until it satisfies the latency constraint. After that, the parameters of the edge/cloud's CNN are transferred to bootstrap the CNN on the FPGA, and they are also periodically transferred to the IoT devices to update. Moreover, the CNN on the FPGA is also trained using I/Q samples, and the obtained information, i.e., states, actions, and values, is transferred to the CNN on the edge/cloud to train the CNN there. Thus, the learning and transfer procedure between the two models can be considered a hybrid of TL and federated learning~\cite{lim_federated_2020}. This can be a promising future research direction that can combine the strengths of both learning frameworks.
\subsubsection{General Optimization Problem in Wireless Networks}
In~\cite{shen_transfer_2019}, the authors formulate the general resource allocation problem as a mixed-integer nonlinear programming (MINLP) model. For solving such MINLP optimization problems, the branch-and-bound algorithm~\cite{conforti_integer_2014} is often employed. However, the solving time of this algorithm grows exponentially as the problem size increases. Therefore, to speed up the branch-and-bound algorithm, the authors propose an imitation learning model to help the branch-and-bound algorithm to prune unnecessary nodes (i.e., pruning policy~\cite{conforti_integer_2014}), thereby improving the branch-and-bound algorithm solving time. However, the pruning policy obtained from the imitation learning model is only effective for the specific MINLP problem which the imitation learning model is trained with. For new problems, the imitation learning model needs to be trained again, which is very time-consuming. Thus, the authors introduce two TL approaches, namely TL via fine-tuning and TL via self-imitation. In the first approach, the previous imitation learning model is transferred to a new target task, and then it is fine-tuned with the target task data, i.e., the new MINLP problem. Although this approach can reduce training time, it still needs labeled data for fine-tuning. In this case, the labels of training data are the optimal solutions, which can only be obtained through the time-consuming process of solving the MINLP. Therefore, the authors propose the second TL approach that does not need labeled data. Particularly, this approach transfers the previous policy from the source task to the target task. Then, the target model will utilize that policy as an initial point and start exploring from there. During the exploration, the objective values of the branch are calculated and set to be the labels, thereby creating labeled data. Simulation results show that the TL via self-imitation approach can help to speed up the training process by 9 to 15 times. It is worth noting that the authors formulate a general optimization problem in this work, and thus the proposed approach can be applied for not only resource allocation problems but also other optimization problems in wireless networks.
\subsubsection{Network Selection} Various traffic types, e.g., virtual reality or high-definition video, and diverse conditions of available wireless networks at a different time may degrade the quality of service (QoS) for the requesting mobile users. For that, a smart network selection leveraging TL can be utilized to maintain high QoS in wireless networks. In~\cite{du_context_2018}, the authors propose a reinforcement TL-based framework using three-level context-aware information to select an indoor network, i.e., LTE, WLAN, or VLC. At the first level, the information of asymmetric downlink-uplink network features and traffic specification is taken into account in the utility function. Then, a context-aware learning algorithm is developed based on the information of traffic type, location, and time at the second level. At the last level, the periodic update rule of load statistical distributions as the knowledge transfer is employed to incorporate the learning algorithm for the online network selection. In particular, the source domain can first train some observations including suitable networks for all traffic types, preferable networks for users, and time-location dependable network statistical distribution. Using these observations, a TL-based Q-learning algorithm can be executed to train the new observations in the target domain. Via the simulations, the proposed reinforcement TL method can achieve a higher average reward with faster convergence compared with that of conventional reinforcement learning.

\subsubsection{Improving TL with other ML Techniques} In~\cite{yang_deep_2020}, a DTL approach is proposed for downlink CSI prediction in different propagation environments. In particular, a fully-connected NN is proposed to predict downlink CSI based on uplink CSI data. To address the lack of labeled data in a new propagation environment, a transductive TL approach is proposed, in which the fully-connected NN is trained with different datasets to learn the generalized weights that are suitable for many tasks. Afterward, this pre-trained model is transferred to initialize the model of the target task, and then the target model is fine-tuned with target data. However, this approach may lead to overfitting if the target data is not sufficient. Therefore, the authors propose another method, namely meta-learning. This method trains a model with several datasets and then transfers the model to the target task for fine-tuning. The main difference is that the training of this approach is done separately for each dataset. Then, the obtained weights from each dataset are aggregated, thereby taking into account the loss function of every task (dataset). As a result, the aggregated weights can be used to adapt to new tasks more efficiently than those of TL. Simulation results show that the meta-learning approach can reduce the Normalized Root Mean Square (NMSE) of the prediction by up to 15\% and 50\% compared to TL and the approach without TL, respectively. Obviously, this meta-learning solution could be very potential to extend to many other problems in wireless networks, such as caching, security and localization. 
\subsubsection{Decentralized TL and Blockchain} Most of the aforementioned TL approaches are centralized, i.e., the transfer of knowledge is governed by a single entity such as a network operator. To fully realize the potential of TL, especially in the era of IoT and 5G/6G networks with a huge number of connected devices, decentralized TL is an aspiring approach. Particularly, the valuable knowledge can be obtained from external sources such as neighboring networks (not controlled by the same operators) or network users (e.g., crowdsourcing). However, for such decentralized approaches, security and privacy become a primary concern, e.g., attackers, eavesdropper, and malicious users. In recent years, blockchain, as a cutting-edge technology for enhancing decentralized network's security and privacy, has been successfully applied for various wireless applications~\cite{dai_blockchain_2019,nguyen_proof_2019}. Thus, utilizing blockchain for decentralized TL is a promising research direction. For example, the authors in~\cite{ul_decentralized_2020} propose a blockchain-based framework for DTL. Using blockchain and smart contracts~\cite{dai_blockchain_2019,nguyen_proof_2019}, the proposed framework allows users to securely share their models, data, and resource. Moreover, the smart contracts also facilitate automatic payment processes, which paves the way for crowdsourcing-enabled TL. Experimental results show that the proposed framework can facilitate ML model sharing over the blockchain network with minimal delay, i.e., less than one minute, while ensuring the users' security and privacy.



\section{Conclusion}\label{sec:Summary}
TL is an effective solution to address various challenges that conventional ML techniques are facing in numerous wireless network applications. In this article, we have presented a comprehensive survey on TL applications in wireless networks. We have first provided an in-depth tutorial of TL including formal definitions, classification, and fundamental background of various TL techniques. Then, we have discussed the applications of TL for various issues of wireless networks, including spectrum management, localization, signal recognition, security, human activity recognition, and caching. For these TL applications, we have provided detailed discussions and analyses on how TL can be leveraged to overcome the challenges that the underlying ML techniques are facing. Finally, we have presented the current challenges, open issues, and introduced some potential research directions of TL for future wireless networks.



\bibliographystyle{ieeetran}

\bibliography{Bibtex/intro,Bibtex/tutorial,Bibtex/spectrum,Bibtex/localization,Bibtex/signal_recognition,Bibtex/security,Bibtex/human,Bibtex/caching,Bibtex/challenge}

\end{document}